\documentclass{article}

\usepackage[preprint,nonatbib]{neurips_data_2022}

\usepackage[utf8]{inputenc} 
\usepackage[T1]{fontenc}    
\usepackage{hyperref}       
\usepackage{url}            
\usepackage{booktabs}       
\usepackage{amsfonts}       
\usepackage{nicefrac}       
\usepackage{microtype}      
\usepackage{xcolor}         
\usepackage{graphicx} 
\usepackage{float} 
\usepackage{subfigure} 
\usepackage{makecell}
\usepackage{multirow}
\usepackage{multicol}
\usepackage[figuresright]{rotating}
\usepackage{enumitem}

\usepackage{amsmath,amsfonts,bm}

\def\eqref#1{equation~\ref{#1}}

\def\1{\bm{1}}

\def\vx{{\bm{x}}}

\DeclareMathAlphabet{\mathsfit}{\encodingdefault}{\sfdefault}{m}{sl}
\SetMathAlphabet{\mathsfit}{bold}{\encodingdefault}{\sfdefault}{bx}{n}

\def\sR{{\mathbb{R}}}

\DeclareMathOperator{\DFT}{\mathfrak{F}}
\DeclareMathOperator{\IDFT}{\mathfrak{F}^{-1}}

\usepackage{stackengine}
\usepackage{scalerel}

\newcommand*{\email}[1]{%
    \normalsize\href{mailto:#1}{#1}\par
    }
    
\usepackage{fontawesome}

\def\ie{$i.e.$}
\def\eg{$e.g.$}
\def\etc{$etc$}
\newcommand{\wrt}{\textit{w.r.t.~}}
\long\def\comment#1{}
\usepackage{xcolor}

\def\blue#1{#1}

\title{BackdoorBench: A Comprehensive Benchmark of Backdoor Learning}

\author{
Baoyuan Wu\textsuperscript{1}\thanks{Corresponds to Baoyuan Wu (\email{wubaoyuan@cuhk.edu.cn}).} 
\ \ \ \ Hongrui Chen\textsuperscript{1} 
\ \ \ \ Mingda Zhang\textsuperscript{1} 
\ \ \ \ Zihao Zhu\textsuperscript{1}
\\
\ \ \ \ \textbf{Shaokui Wei}\textsuperscript{1} 
\ \ \ \ \textbf{Danni Yuan}\textsuperscript{1} 
\ \ \ \ \textbf{Chao Shen}\textsuperscript{2}
\\
\textsuperscript{1}School of Data Science, Shenzhen Research Institute of Big Data, \\
The Chinese University of Hong Kong, Shenzhen\\
\textsuperscript{2}School of Cyber Science and Engineering, 
Xi’an Jiaotong University 
}

\comment{
\author{%
  Baoyuan Wu\thanks{Correspondence to Baoyuan Wu.}, 
  Mingda Zhang, Zihao Zhu, Shaokui Wei, Danni Yuan, Chao Shen, Hongyuan Zha\\
  SDS, SRIBD, CUHKSZ, Shenzhen, China\\
  \texttt{wubaoyuan@cuhk.edu.cn} \\
   \And
   Hongrui Chen \\
}
}

\begin{document}

\maketitle

\begin{abstract} 
  Backdoor learning is an emerging and vital topic for studying deep neural networks' vulnerability (DNNs). 
  Many pioneering backdoor attack and defense methods are being proposed, successively or concurrently, in the status of a rapid arms race. 
  However, we find that the evaluations of new methods are often unthorough to verify their claims and accurate performance, mainly due to the rapid development, diverse settings, and the difficulties of implementation and reproducibility.  
  Without thorough evaluations and comparisons, it is not easy to track the current progress and design the future development roadmap of the literature. 
  To alleviate this dilemma, we build a comprehensive benchmark of backdoor learning called \textit{BackdoorBench}. 
  It consists of an extensible modular-based codebase (currently including implementations of 8 state-of-the-art (SOTA) attacks and 9 SOTA defense algorithms) and a standardized protocol of complete backdoor learning. 
  We also provide comprehensive evaluations of every pair of 8 attacks against 9 defenses, with 5 poisoning ratios, based on 5 models and 4 datasets, thus 8,000 pairs of evaluations in total. 
  We present abundant analysis from different perspectives about these 8,000 evaluations, studying the effects of different factors in backdoor learning. 
  All codes and evaluations of BackdoorBench are publicly available at \url{https://backdoorbench.github.io}.
\end{abstract}

\section{Introduction}
With the widespread application of deep neural networks (DNNs) in many mission-critical scenarios, the security issues of DNNs have attracted more attentions. One of the typical security issue is backdoor learning, which could insert an imperceptible backdoor into the model through maliciously manipulating the training data or controlling the training process.  
It brings in severe threat to the widely adopted paradigm that people often download a unverified dataset/checkpoint to train/fine-tune their models, or even outsource the training process to the third-party training platform.

Although backdoor learning is a young topic in the machine learning community, its development speed is remarkable and has shown the state of a rapid arms race. 
When a new backdoor attack or defense method is developed based on an assumption or observation, it will be quickly defeated or evaded by more advanced adaptive defense or attack methods which break previous assumptions or observations. 
However, we find that the evaluations of new methods are often insufficient, with comparisons with limited previous methods, based on limited models and datasets. 
The possible reasons include the rapid development of new methods, diverse settings (\eg, different threat models), as well as the difficulties of implementing or reproducing previous methods. 
Without thorough evaluations and fair comparisons, it is difficult to verify the real performance of a new method, as well as the correctness or generalization of the assumption or observation it is built upon. 
Consequently, we cannot well measure the actual progress of backdoor learning by simply tracking new methods. 
This dilemma may not only postpone the development of more advanced methods, but also preclude the exploration of the intrinsic reason/property of backdoor learning.

To alleviate this dilemma, we build a comprehensive benchmark of backdoor learning, called \textbf{BackdoorBench}. It is built on an extensible modular based codebase, consisting of the attack module, the defense module, as well as the evaluation and analysis module. Until now, we have implemented 8 stat-of-the-art (SOTA) backdoor attack methods and 9 SOTA defense methods, and provided 5 analysis tools (\eg, t-SNE, Shapley value, Grad-CAM, frequency saliency map and neuron activation). More methods and tools are continuously updated. 
Based on the codebase, to ensure fair and reproducible evaluations, we also provide a standardized protocol of the complete procedure of backdoor learning, covering every step of the data preparation, backdoor attack, backdoor defense, as well as the saving, evaluation and analysis of immediate/final outputs. 
Moreover, we conduct comprehensive evaluations of every pair of attack and defense method (\ie, 8 attacks against 9 defenses), with 5 poisoning ratios, based on 5 DNN models and 4 databases, thus up to 8,000 pairs of evaluations in total. 
These evaluations allow us to analyze some characteristics of backdoor learning. In this work, we present the analysis from four perspectives, to study the effects of attack/defense methods, poisoning ratios, datasets and model architectures, respectively. 
We hope that BackdoorBench could provide useful tools to facilitate not only the design of new attack/defense methods, but also the exploration of intrinsic properties and reasons of backdoor learning, such that to promote the development of backdoor learning.

Our main contributions are three-fold. \textbf{1) Codebase}: We build an extensible modular based codebase, including the implementations of 8 backdoor attack methods and 9 backdoor defense methods.
\textbf{2) 8,000 comprehensive evaluations}: We provide evaluations of all pairs of 8 attacks against 9 defense methods, with 5 poisoning ratios, based on 4 datasets and 5 models, up to 8,000 pairs of evaluations in total.
\textbf{3) Thorough analysis and new findings}: We present thorough analysis of above evaluations from different perspectives to study the effects of different factors in backdoor learning, with the help of 5 analysis tools, and show some interesting findings to inspire future research directions.

\section{Related work}

\textbf{Backdoor attacks} 
According to the threat model, existing backdoor attack methods can be partitioned into two general categories, including \textbf{data poisoning} and \textbf{training controllable}. \textbf{1)} \textbf{Data poisoning attack} means that the attacker can only manipulate the training data. Existing methods of this category focuses on designing different kinds of triggers to improve the imperceptibility and attack effect, including visible (\eg, BadNets  \cite{gu2019badnets}) vs invisible (\ie, Blended \cite{chen2017targeted}, Refool \cite{liu2020reflection}, Invisible backdoor \cite{li2020invisible}) triggers, local (\eg, label consistent attack \cite{clean-label-2018,zhao2020clean}) vs global (\eg, SIG \cite{SIG}) triggers, additive (\eg, Blended \cite{chen2017targeted}) vs non-additive triggers (\eg, smooth low frequency (LF) trigger \cite{zeng2021rethinking}, FaceHack \cite{sarkar2022facehack}), sample agnostic (\eg, BadNets \cite{gu2019badnets}) vs sample specific (\eg, SSBA \cite{ssba}, sleeper agent \cite{souri2021sleeper}) triggers, \etc. The definitions of these triggers can be found in the bottom notes of Table \ref{tab: 8 attacks}. 
\textbf{2)} \textbf{Training controllable attack} means that the attacker can control both the training process and training data simultaneously. Consequently, the attacker can learn the trigger and the model weights jointly, such as LIRA \cite{doan2021lira}, blind backdoor \cite{bagdasaryan2021blind}, WB \cite{doan2021backdoor}, Input-aware \cite{nguyen2020input}, WaNet \cite{nguyen2021wanet}, \etc. 

\textbf{Backdoor defenses} According to the defense stage in the training procedure, existing defense methods can be partitioned into three categories, including \textbf{pre-training}, \textbf{in-training} and \textbf{post-training}. \textbf{1)} \textbf{Pre-training defense} means that the defender aims to remove or break the poisoned samples before training. 
For example, input anomaly detection and input pre-processing were proposed in \cite{liu2017neural} to block the backdoor activation by poisoned samples. 
Februus \cite{doan2020februus} firstly identified the location of trigger using Grad-CAM \cite{grad-cam}, and then used a GAN-based inpainting method \cite{iizuka2017globally} to reconstruct that region to break the trigger.
NEO \cite{udeshi2022model} proposed to use the dominant color in the image to generate a patch to cover the identified trigger. 
Confoc \cite{villarreal2020confoc} proposed to change the style of the input image \cite{gatys2016image} to break the trigger. 
\textbf{2)} \textbf{In-training defense} means that the defender aims to inhibit the backdoor injection during the training. 
For example, anti-backdoor learning (ABL) \cite{li2021anti} utilized the fact that poisoned samples are fitted faster than clean samples, such that they can be distinguished by the loss values in early learning epochs, then the identified poisoned samples are unlearned to mitigate the backdoor effect. DBD \cite{huang2022backdoor} observed that poisoned samples will gather together in the feature space of the backdoored model. To prevent such gathering, DBD utilized the self-supervised learning \cite{simclr} to learn the model backbone, then identified the poisoned samples according to the loss values when learning the classifier. 
\textbf{3)} \textbf{Post-training defense} means that the defender aims to remove or mitigate the backdoor effect from a backdoored model, and most existing defense methods belong to this category. They are often motivated by a property or observation of the backdoored model using some existing backdoor attacks. For example, the fine-pruning (FP) defense \cite{FP} and the neural attention distillation (NAD) \cite{nad-iclr-2020} observed that poisoned and clean samples have different activation paths in the backdoored model. Thus, they aimed to mitigate the backdoor effect by pruning the neurons highly related to the backdoor. 
The channel Lipschitzness based pruning (CLP) method \cite{zheng2022data} found that the backdoor related channels often have a higher Lipschitz constant compared to other channels, such that the channels with high Lipschitz constant could be pruned to remove the backdoor. 
The activation clustering (AC) method \cite{AC} observed that samples of the target class will form two clusters in the feature space of a backdoored model, and the smaller cluster corresponds to poisoned samples. 
The spectral signatures (Spectral) method \cite{tran2018spectral} observed that the feature representation distributions of poisoned and clean samples in the same class class are spectrally separable.
The neural cleanse (NC) method \cite{wang2019neural} assumed that the trigger provides a ``shortcut" between the samples from different source classes and the target class.
The adversarial neuron pruning (ANP) defense \cite{wu2021adversarial} found that the neurons related to the injected backdoor are more sensitive to adversarial neuron perturbation (\ie, perturbing the neuron weight to achieve adversarial attack) than other neurons in a backdoored model. 
We refer the readers to some backdoor surveys \cite{gao2020backdoor,liu2020survey} for more backdoor attack and defense methods.

\textbf{Related benchmarks}
Several libraries or benchmarks have been proposed for evaluating the adversarial robustness of DNNs, such as CleverHans \cite{papernot2016cleverhans}, Foolbox  \cite{rauber2017foolbox,rauber2017foolboxnative}, AdvBox \cite{goodman2020advbox}, RobustBench \cite{croce2020robustbench}, RobustART \cite{tang2021robustart}, ARES \cite{dong2020benchmarking}, Adversarial Robustness Toolbox (ART) \cite{ART-IBM}, \etc. 
However, these benchmarks mainly focused on adversarial examples \cite{fgsm,kurakin2018adversarial}, which occur in the testing stage. 
In contrast, there are only a few libraries or benchmarks for backdoor learning (\eg,  TrojAI \cite{karra2020trojai} and TrojanZoo \cite{trojanzoo2022}).
Specifically, the most similar benchmark is TrojanZoo, which implemented 8 backdoor attack methods and 14 backdoor defense methods. However, there are significant differences between TrojanZoo and our BackdoorBench in two main aspects. \textbf{1) Codebase}: although both benchmarks adopt the modular design to ensure easy extensibility, TrojanZoo adopts the object-oriented programming (OOP) style, where each module is defined as one class. In contrast, BackdoorBench adopts the procedural oriented programming (POP) style, where each module is defined as one function, and each specific algorithm is implemented by several functions in a streamline. 
\textbf{2) Analysis and findings}. TrojanZoo has provided very abundant and diverse analysis of backdoor learning, mainly including the attack effects of trigger size, trigger transparency, data complexity, backdoor transferability to downstream tasks, and the defense effects of the tradeoff between robustness and utility, the tradeoff between detection accuracy and recovery capability, the impact of trigger definition. In contrast, BackdoorBench provides several new analysis from different perspectives, mainly including the effects of poisoning ratios and number of classes, the quick learning of backdoor, trigger generalization, memorization and forgetting of poisoned samples, as well as several analysis tools. 
In summary, we believe that BackdoorBench could provide new contributions to the backdoor learning community, and the competition among different benchmarks is beneficial to the development of this topic.

\begin{table*}[ht]
\centering
\caption{Categorizations of 8 backdoor attack algorithms in BackdoorBench, according to \textit{threat models} and \textit{different kinds of trigger characteristics}.}
\label{tab: 8 attacks}
\scalebox{0.83}{
\begin{tabular}{p{.15\textwidth}|p{.055\textwidth} p{.055\textwidth} | p{.055\textwidth} p{.055\textwidth} | p{.055\textwidth} p{.058\textwidth} | p{.055\textwidth} p{.07\textwidth} | p{.06\textwidth} p{.03\textwidth}  }
\hline
 Attack  & \multicolumn{2}{c|}{Threat model} & \multicolumn{8}{c}{Trigger characteristics} 
 \\
 \cline{4-11}
 algorithm & D-P & T-C & V & In-V & Local & Global  & Add & N-Add & Ag & Sp
\\
\hline \hline
BadNets \cite{gu2019badnets} & \checkmark &   & \checkmark &   & \checkmark &  & \checkmark &  & \checkmark & 
\\
Blended \cite{chen2017targeted} & \checkmark &   & & \checkmark   & & \checkmark  & \checkmark &  & \checkmark & 
\\
LC \cite{clean-label-2018} & \checkmark &   & \checkmark &   & \checkmark &  & \checkmark &  & \checkmark & 
\\
SIG \cite{SIG} & \checkmark &   & & \checkmark   & & \checkmark  &  & \checkmark  & \checkmark & 
\\
LF \cite{zeng2021rethinking} & \checkmark &   & & \checkmark   & & \checkmark  &  & \checkmark  & &  \checkmark 
\\
SSBA \cite{ssba} & \checkmark &   & & \checkmark   & & \checkmark  &  & \checkmark  & &  \checkmark
\\
\scalebox{0.85}{Input-aware} \cite{nguyen2020input} & &  \checkmark   & \checkmark &   & \checkmark &  & \checkmark &  & &  \checkmark 
\\
WaNet \cite{nguyen2021wanet} & &  \checkmark &  & \checkmark &  & \checkmark & & \checkmark & & \checkmark
\\
\hline
\multicolumn{11}{l}{a) \textbf{Threat model}: D-P $\rightarrow$ data poisoning, \ie, the attacker can only manipulate the training data; } 
\\
\multicolumn{11}{l}{ \hspace{3em} T-C $\rightarrow$ training controllable, \ie, the attacker can control the training process and data;} 
\\
\multicolumn{11}{l}{b) \textbf{Trigger characteristics}:  }
\\
\multicolumn{11}{l}{  \hspace{3em} \textbf{b.1) Trigger visibility}: 
V $\rightarrow$ visible; In-V $\rightarrow$ invisible; } 
\\
\multicolumn{11}{l}{  \hspace{3em} \textbf{b.2) Trigger coverage}: Local $\rightarrow$ the trigger is a local patch; Global $\rightarrow$ the trigger covers the whole sample;} 
\\
\multicolumn{11}{l}{ \hspace{3em} \textbf{b.3) Trigger fusion mode}: Add $\rightarrow$ additive, \ie, the fusion between the clean sample and the trigger is additive;} 
\\
\multicolumn{11}{l}{ \hspace{4.9em} N-Add $\rightarrow$ non-additive, \ie, the fusion between the clean sample and the trigger is non-additive;} 
\\
\multicolumn{11}{l}{ \hspace{3em} \textbf{b.4) Trigger fusion mode}: Ag $\rightarrow$ agnostic, \ie, the triggers in all poisoned samples are same;} 
\\
\multicolumn{11}{l}{ \hspace{4.9em} Sp $\rightarrow$ specific, \ie, different poisoned samples have different triggers.} 
\end{tabular}
}
\end{table*}

\begin{table*}[ht]
\centering
\caption{Categorizations of 9 backdoor defense algorithms in BackdoorBench, according to four perspectives, including \textit{input}, \textit{output}, \textit{defense stage} and \textit{defense strategy}.}
\label{tab: 8 defense}
\scalebox{0.73}{
\begin{tabular}{p{.12\textwidth}|| p{.04\textwidth} p{.04\textwidth} p{.04\textwidth} | p{.042\textwidth} p{.042\textwidth} | p{.042\textwidth} p{.052\textwidth} | p{.075\textwidth} | p{.5\textwidth} }
\hline
 Defense  & \multicolumn{3}{c|}{Input} & \multicolumn{2}{c|}{Output} & \multicolumn{2}{c|}{\scalebox{0.8}{Defense stage}} & \scalebox{0.8}{Defense } & \multirow{2}{*}{Motivation/Assumption/Observation}
\\
algorithm & \scalebox{0.95}{B-M} & \scalebox{0.9}{S-CD} & \scalebox{0.95}{P-D} & \scalebox{0.95}{S-M} & \scalebox{0.95}{C-D} & \scalebox{0.95}{In-T} & \scalebox{0.86}{Post-T} & \scalebox{0.8}{strategy} & 
\\
\hline \hline
FT & \checkmark & \checkmark &  & \checkmark & &  & \checkmark & 5 &  Fine-tuning on clean data could mitigate the backdoor effect
\\
\hline
FP \cite{FP} & \checkmark & \checkmark &  & \checkmark &  & & \checkmark & 2 + 5 & Poisoned and clean samples have different activation paths 
\\
\hline
NAD \cite{nad-iclr-2020} & \checkmark & \checkmark &  & \checkmark &  & & \checkmark & 5 & Fine-tuning on clean data could mitigate the backdoor effect
\\
\hline
NC \cite{wang2019neural} & \checkmark & \checkmark &  & \checkmark &  & & \checkmark & 1 + 4 + 5 & Trigger can be reversed through searching a shortcut to the target class
\\
\hline
ANP \cite{wu2021adversarial} & \checkmark & \checkmark &  & \checkmark &  & & \checkmark & 2 + 5 & 
The backdoor related neurons are sensitive to adversarial neuron perturbation
\\
\hline
AC \cite{AC} &  & & \checkmark & \checkmark & \checkmark & & \checkmark & 3 + 5 & Samples labeled the target class will form 2 clusters in the feature space of a backdoored model
\\
\hline
\scalebox{0.85}{Spectral} \cite{tran2018spectral} & & & \checkmark  & \checkmark & \checkmark & & \checkmark & 3 + 5 & The feature representations of poisoned and clean samples have different spectral signatures
\\
\hline
ABL \cite{li2021anti} & & & \checkmark  & \checkmark & \checkmark & \checkmark & & 3 + 5 & Poisoned samples are learned more quickly than clean samples during the training
\\
\hline
DBD \cite{huang2022backdoor} & & & \checkmark & \checkmark & \checkmark & \checkmark & & 3 + 6 & Poisoned samples will gather together in the feature space due to the standard supervised learning
\\
\hline
\multicolumn{10}{l}{a) \textbf{Input}: B-M $\rightarrow$ a backdoored model; S-CD $\rightarrow$ a subset of clean samples; P-D $\rightarrow$ a poisoned dataset;} 
\\
\multicolumn{10}{l}{b) \textbf{Output}: S-M $\rightarrow$ secure model; C-D $\rightarrow$ clean data, \ie, the subset of clean samples in the input poisoned data;}
\\
\multicolumn{10}{l}{c) \textbf{Defense stage}: In-T $\rightarrow$ in-training, \ie, defense happens during the training process; }
\\
\multicolumn{10}{l}{\hspace{3.81em} Post-T $\rightarrow$ post-training, \ie, defense happens after the backdoor has been inserted through training;}
\\
\multicolumn{10}{l}{d) \textbf{Defense strategy}: \textbf{1 $\rightarrow$ backdoor detection}, \ie, determining a model to be backdoored or clean; }
\\
\multicolumn{10}{l}{\hspace{3.81em}  \textbf{2 $\rightarrow$ backdoor identification}, \ie, identifying the neurons in a backdoored model related to the backdoor;}
\\
\multicolumn{10}{l}{\hspace{3.81em}  \textbf{3 $\rightarrow$ poison detection}, \ie, detecting poisoned samples;}
\\
\multicolumn{10}{l}{\hspace{3.6em} \textbf{ 4 $\rightarrow$ trigger identification}, \ie, identifying the trigger location in a poisoned sample;}
\\
\multicolumn{10}{l}{\hspace{3.81em}  \textbf{5 $\rightarrow$ backdoor mitigation}, \ie, mitigating the backdoor effect of a backdoored model;}
\\
\multicolumn{10}{l}{\hspace{3.81em}  \textbf{6 $\rightarrow$ backdoor inhibition}, \ie, inhibiting the backdoor insertion into the model during the training.}
\end{tabular}
}
\end{table*}

\vspace{-0em}
\section{Our benchmark}

\subsection{Implemented algorithms}
\vspace{-0.2em}

We have implemented 8 backdoor attack and 9 backdoor defense algorithms as the first batch of algorithms in our benchmark. 
We hold two criteria for choosing methods. \textbf{First}, it should be classic (\eg, BadNets) or advanced method (\ie, published in recent top-tier conferences/journals in machine learning or security community). The classic method serves as the baseline, while the advanced method represents the state-of-the-art, and their difference could measure the progress of this field. 
\textbf{Second}, the method should be easily implemented and reproducible. We find that some existing methods involve several steps, and some steps depend on a third-party algorithm or a heuristic strategy. Consequently, these methods involve too many hyper-parameters and are full of uncertainty, causing the difficulty on implementation and reproduction. Such methods are not included in BackdoorBench. 

As shown in Table \ref{tab: 8 attacks}, the eight implemented backdoor attack methods cover two mainstream threat models, and with diverse triggers. Among them, BadNets\cite{gu2019badnets}, Blended\cite{chen2017targeted} and LC\cite{clean-label-2018} (label consistent attack) are three classic attack methods, while the remaining 5 are recently published methods. The general idea of each method will be presented in the \textbf{Appendix}. 

The basic characteristics of 9 implemented backdoor defense methods are summarized in Table \ref{tab: 8 defense}, covering different inputs and outputs, different happening stages, different defense strategies. The motivation/assumption/observation behind each defense method is also briefly described in the last column. More detailed descriptions will be presented in the \textbf{Appendix}.

\begin{figure}[!t] 
\centering 
\includegraphics[width=1\textwidth]{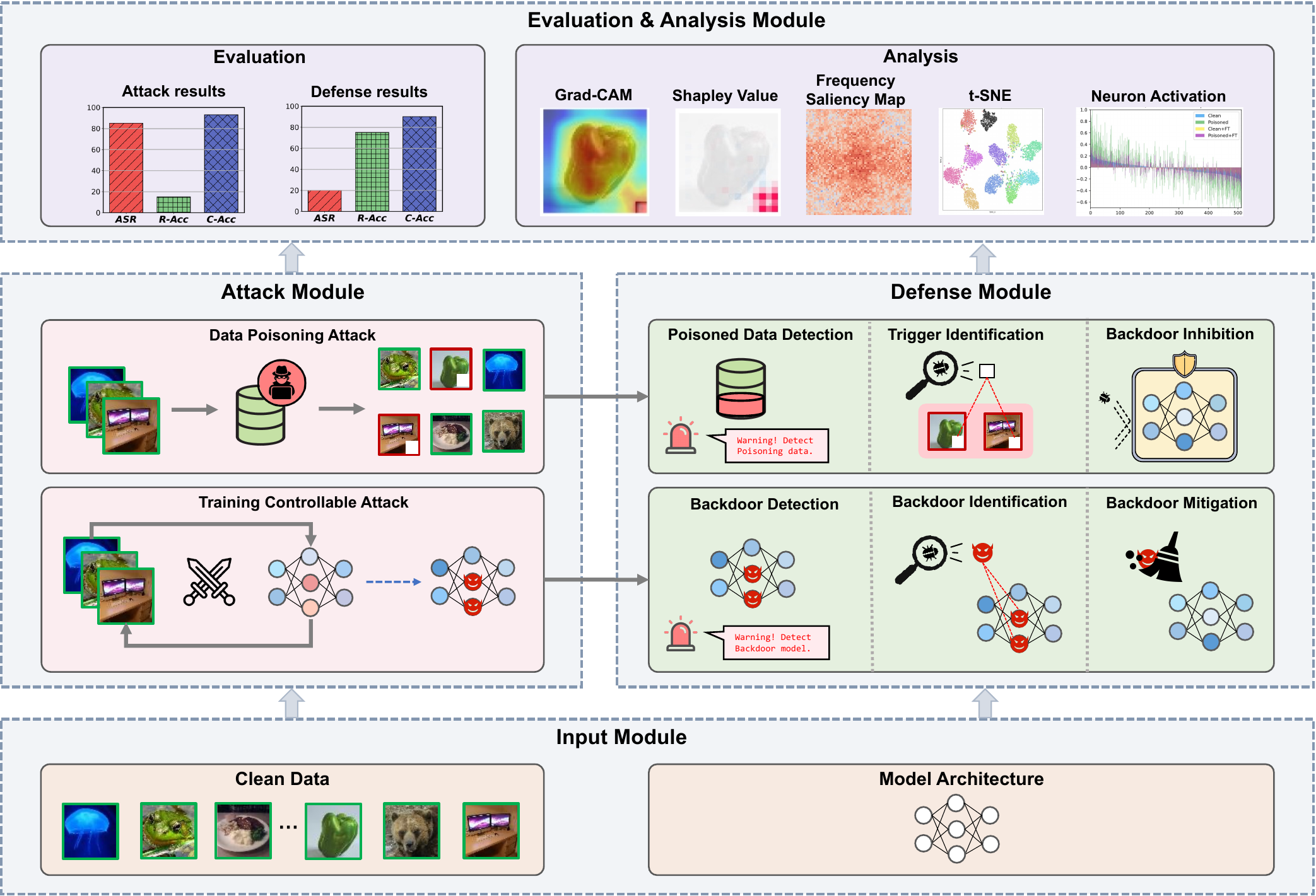} 
\vspace{-0.3em}
\caption{The general structure of the modular based codebase of BackdoorBench.} 
\label{fig:framework} 
\vspace{-1em}
\end{figure}

\vspace{-0.2em}
\subsection{Codebase}
\label{sec: codebase}
\vspace{-0.2em}


We have built an extensible modular-based codebase as the basis of BackdoorBench. As shown in Fig. \ref{fig:framework}, it consists of four modules, including \textit{input module} (providing clean data and model architectures), \textit{attack module}, \textit{defense module} and \textit{evaluation and analysis module}.


\textbf{Attack module}
In the attack module, we provide two sub-modules to implement attacks of two threat models, \ie, \textit{data poisoning} and \textit{training controllable} (see Table \ref{tab: 8 attacks}), respectively. 
For the first sub-module, it provides some functions of manipulating the provided set of clean samples, including trigger generation, poisoned sample generation (\ie, inserting the trigger into the clean sample), and label changing. It outputs a poisoned dataset with both poisoned and clean samples.
For the second sub-module, given a set of clean samples and a model architecture, it provides two functions of learning the trigger and model parameters, and outputs a backdoored model and the learned trigger. 

\textbf{Defense module}
According to the outputs produced by the attack module, there are also two sub-modules to implement backdoor defenses. 
If given a poisoned dataset, the first sub-module provides three functions of \textit{poisoned sample detection} (\ie, determining whether a sample is poisoned or clean), \textit{trigger identification} (\ie, identifying the location in the poisoned sample), \textit{backdoor inhibition} (\ie, training a secure model through inhibiting the backdoor injection). 
If given a backdoored model, as well as a small subset of clean samples (which is widely required in many defense methods), the second sub-module provides three functions of \textit{backdoor detection} (\ie, determining whether a model has a backdoor or not), \textit{badckdoor identification} (\ie, identifying the neurons in the backdoored model that are related to the backdoor effect), \textit{backdoor mitigation} (\ie, mitigating the backdoor effect from the backdoored model).

\textbf{Evaluation and analysis module}
\textbf{1)} We provide \textbf{three evaluation metrics}, including \textit{clean accuracy (C-Acc)} (\ie, the prediction accuracy of clean samples), \textit{attack success rate (ASR)} (\ie, the prediction accuracy of poisoned samples to the target class), \textit{robust accuracy (R-Acc)} (\ie, the prediction accuracy of poisoned samples to the original class). Note that the new metric R-Acc satisfies that ASR + R-Acc $\leq 1$, and lower ASR and higher R-Acc indicate better defense performance. 
\textbf{2)} Moreover, we provide \textbf{five analysis tools} to facilitate the analysis and understanding of backdoor learning. \textit{t-SNE} provides a global visualization of feature representations of a set of samples in a model, and it can help us to observe whether the backdoor is formed or not.
\textit{Gradient-weighted class activation mapping (Grad-CAM)} \cite{grad-cam} and \textit{Shapley value map} \cite{NIPS2017_7062} are two individual analysis tools to visualize the contributions of different pixels of one image in a model, and they can show that whether the trigger activates the backdoor or not. 
We also propose the \textit{frequency saliency map} to visualize the contribution of each individual frequency spectrum to the prediction, providing a novel perspective of backdoor from the frequency space. The definition will be presented in \textbf{Appendix}.
\textit{Neuron activation} calculates the average activation of each neuron in a layer for a batch of samples. It can be used to analyze the activation path of poisoned and clean samples, as well as the activation changes \wrt the model weights' changes due to attack or defense, providing deeper insight behind the backdoor. 


\textbf{Protocol}
We present a standardized protocol to call above functional modules to conduct fair and reproducible backdoor learning evaluations, covering every stage from data pre-processing, backdoor attack, backdoor defense, result evaluation and analysis, \etc. 
We also provide three flexible calling modes, including \textit{pure attack mode} (only calling an attack method), \textit{pure defense mode} (only calling a defense method), as well as \textit{a joint attack and defense mode} (calling an attack against a defense).


\comment{
\textbf{Flexible calling modes}:
We also provide flexible calling modes, including pure attack mode, pure defense mode, as well as a joint attack and defense mode. 

introduce the module-based structure, and a script:
outputs (\eg, backdoored model, recovered model, poisoned samples, metric values) = backdoor-evaluation(dataset, poisoning ratio, attack algorithm, defense algorithm, analysis tool)

the attack and defense module can be separately or jointly called 
}

\vspace{-0em}
\section{Evaluations and analysis}
\vspace{-0.3em}

\subsection{Experimental setup}
\vspace{-0.3em}

\textbf{Datasets and models}
We evaluate our benchmark on 4 commonly used datasets (CIFAR-10 \cite{krizhevsky2009learning}, CIFAR-100 \cite{krizhevsky2009learning}, GTSRB \cite{gtsrb}, Tiny ImageNet \cite{Le2015TinyIV}) and 5 backbone models (PreAct-ResNet18\footnote{\href{https://github.com/VinAIResearch/Warping-based_Backdoor_Attack-release/blob/main/classifier_models/preact_resnet.py}{https://github.com/VinAIResearch/Warping-based\_Backdoor\_Attack-release/blob/main/classifier\_models/preact\_resnet.py}} \cite{he2016identity}, VGG-19 \footnote{\href{https://pytorch.org/vision/0.12/\_modules/torchvision/models/vgg.html\#vgg19}{https://pytorch.org/vision/0.12/\_modules/torchvision/models/vgg.html\#vgg19}} \cite{simonyan2015very} (without the batchnorm layer), EfficientNet-B3 \footnote{\href{https://pytorch.org/vision/main/\_modules/torchvision/models/efficientnet.html\#efficientnet\_b3}{https://pytorch.org/vision/main/\_modules/torchvision/models/efficientnet.html\#efficientnet\_b3}} \cite{tan2019efficientnet},  MobileNetV3-Large \footnote{\href{https://github.com/pytorch/vision/blob/main/torchvision/models/mobilenetv3.py}{https://github.com/pytorch/vision/blob/main/torchvision/models/mobilenetv3.py}}\cite{howard2019searching}, DenseNet-161 \footnote{\href{https://pytorch.org/vision/main/\_modules/torchvision/models/densenet.html\#densenet161}{https://pytorch.org/vision/main/\_modules/torchvision/models/densenet.html\#densenet161}}  \cite{huang2017densely}). To fairly measure the performance effects of the attack and defense method for each model, we only used the basic version of training for each model without adding any other training tricks (\eg, augmentation). The details of datasets and clean accuracy \footnote{Note that to fairly measure the effects of the attack and defense method, we train all victim models from scratch without further training tricks, which explains the low clean accuracy of some models.} of normal training are summarized in Table \ref{table:datasets}.

\begin{table*}[ht]
\renewcommand\arraystretch{1.5}
\center
\caption{Dataset details and clean accuracy of normal training. }
\label{table:datasets}
\resizebox{\linewidth}{!}{
\begin{tabular}{l|c|c|c|ccccc}
\hline
\multirow{2}{*}{Datasets} & \multirow{2}{*}{Classes} & \multirow{2}{*}{\makecell{Training/\\Testing Size}} & \multirow{2}{*}{Image Size} & \multicolumn{5}{c}{Clean Accuracy}\\
\cline{5-9}&&&& PreAct-ResNet18  \cite{he2016identity}& VGG-19 \cite{simonyan2015very}  & EfficientNet-B3 \cite{tan2019efficientnet} & MobileNetV3-Large \cite{howard2019searching} & DenseNet-161\cite{huang2017densely}\\ 
\hline
CIFAR-10 \cite{krizhevsky2009learning}                 & 10  & 50,000/10,000 & $32\times 32$ &93.90\% & 91.38\%&64.69\%&84.44\%& 86.82\%\\
CIFAR-100 \cite{krizhevsky2009learning}                & 100 & 50,000/10,000 & $64\times 64$ &70.51\%& 60.21\%& 48.92\%&50.73\%&57.57\%\\
GTSRB   \cite{gtsrb}                  & 43  & 39,209/12,630 & $32\times 32$ & 98.46\%& 95.84\%& 87.39\%&93.99\%&92.49\%\\
Tiny ImageNet  \cite{Le2015TinyIV}           & 200 & 100,000/10,000& $64\times 64$ &57.28\%& 46.13\%&41.08\%&38.78\%&51.73\%\\ 
\hline
\end{tabular}
}
\end{table*}

\textbf{Attacks and defenses} We evaluate each pair of 8 attacks against 9 defenses in each setting, as well as one attack without defense. 
Thus, there are $8 \times (9+1) = 80$ pairs of evaluations. We consider 5 poisoning ratios, \ie, $0.1\%, 0.5\%, 1\%, 5\%, 10\%$ for each pair, based on all 4 datasets and 5 models, leading to $8,000$ pairs of evaluations in total. 
The performance of every model is measured by the metrics, \ie,  C-Acc, ASR and R-Acc (see Section \ref{sec: codebase}). 
The implementation details of all algorithms, and the results of the DBD defense \cite{huang2022backdoor} 
will be presented in the  \textbf{Appendix}. 

\vspace{-0em}
\subsection{Results overview}
\vspace{-0em}

\begin{figure}[t]
    \centering
    \includegraphics[width=0.95\textwidth]{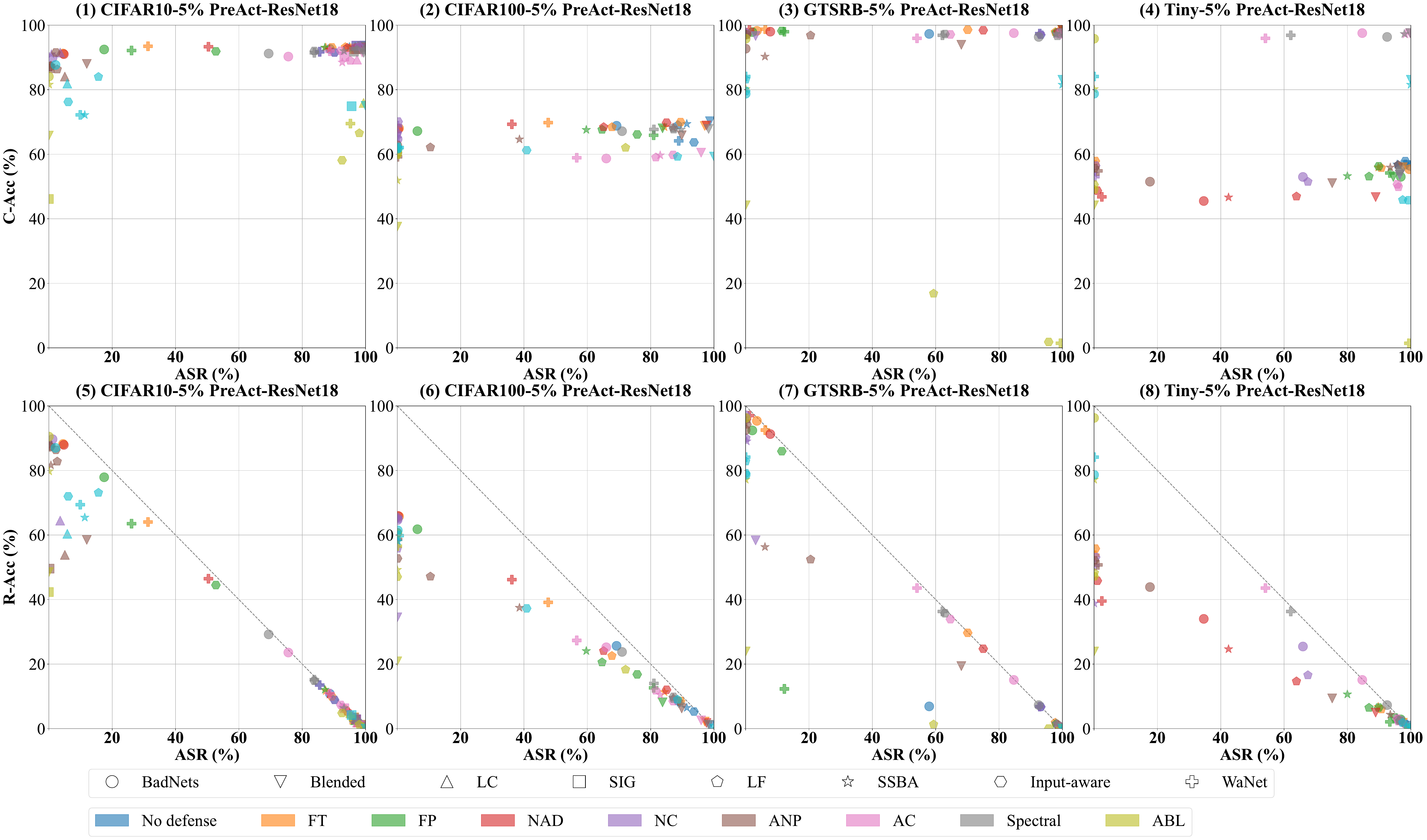}
    \vspace{-0em}
    \caption{Performance distribution of different attack-defense pairs. 
    Each color pattern represents one attack-defense pair, with attacks distinguished by patterns, while defenses by colors.}
    \label{acc-asr-scatter}
    \vspace{-0em}
\end{figure}

We first show the performance distribution of various attack-defense pairs under one model structure (\ie, PreAct-ResNet18) and one poisoning ratio (\ie, $5\%$) in Figure ~\ref{acc-asr-scatter}. 
In the top row, the performance is measured by clean accuracy (C-Acc) and attack success rate (ASR). From the attacker's perspective, the perfect performance should be high C-Acc and high ASR simultaneously, \ie, located  at the top-right corner. From the defender's perspective, the performance should be high C-Acc and low ASR simultaneously, \ie, located  at the top-left corner. It is observed that most color patterns locate  at similar horizontal levels, reflecting that most defense methods could mitigate the backdoor effect while not harming the clean accuracy significantly. 
In the bottom row, the performance is measured by robust accuracy (R-Acc) and ASR. As demonstrated in Section \ref{sec: codebase}, ASR + R-Acc $\leq 1$. From the defender's perspective, it is desired that the reduced ASR value equals to the increased R-Acc, \ie, the prediction of the poisoned sample is recovered to the correct class after the defense. It is interesting to see that most color patterns are close to the anti-diagonal line (\ie, ASR + R-Acc $= 1$) on CIFAR-10 (the first column) and GTSRB (the third column), while most patterns are from that line on CIFAR-100 (the second column) and Tiny ImageNet (the last column). We believe it is highly related to the number of classes of the dataset. Given a large number of classes, it is more difficult to recover the correct prediction after the defense. 
These figures could provide a big picture of the performance of most attacks against defense methods. 
Due to the space limit, the results of other settings will be presented in the  \textbf{Appendix}. 

\vspace{-0.2em}
\subsection{Effect of poisoning ratio}
\label{sec: subsec effect of poisoning ratio}

\begin{figure}[!ht]
    \centering
    \includegraphics[width=1\textwidth]{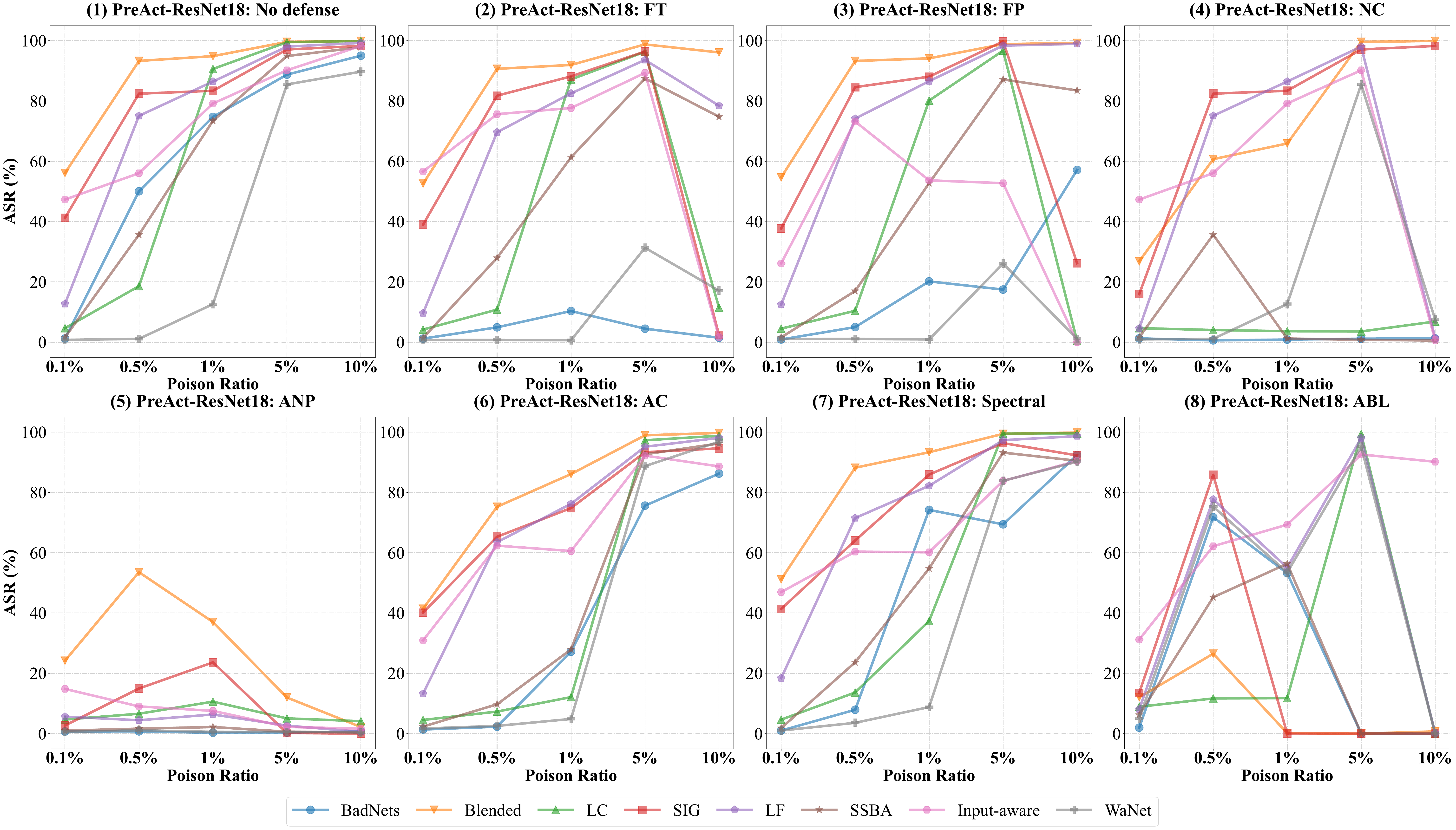}
    \vspace{-0.4em}
    \caption{The effects of different poisoning ratios on backdoor learning.}
    \label{2-attack-asr-preact}
\end{figure}

Here we study the effect of the poisoning ratio on the backdoor performance. 
Figure ~\ref{2-attack-asr-preact} visualizes the results on CIFAR-10 and PreAct-ResNet18, \wrt each poisoning ratio for all attack-defense pairs, and each sub-figure corresponds to each defense. 
In sub-figures (1,6,7), ASR curves increase in most cases, being consistent with our initial impression that higher poisoning ratios lead to stronger attack performance. 
However, in other sub-figures, there are surprisingly sharp drops in ASR curves. To understand such \textit{abnormal} phenomenon, we conduct deep analysis for these defenses, as follows.

\begin{figure}
     \centering
     \includegraphics[width=1\textwidth]{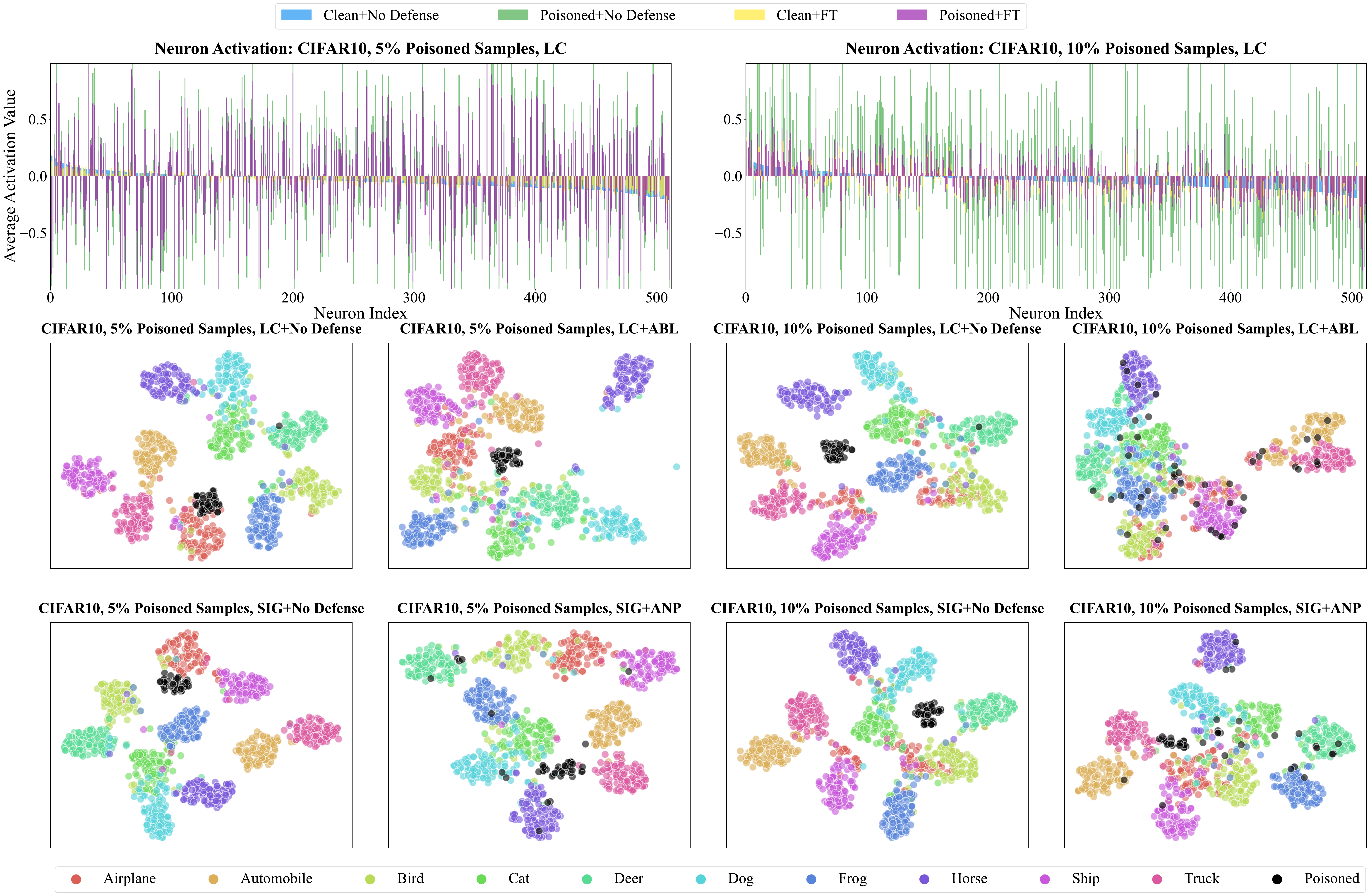}
     \vspace{-0.4em}
     \caption{The changes of neuron activation values due to the FT defense (\textbf{Top row}), and the changes of t-SNE visualization of feature representations due to the ABL defense (\textbf{Middle row}) and the ANP defense (\textbf{Bottom row}), respectively. }
    \label{fig:abnormal}
\end{figure}

\vspace{-1em}
\paragraph{Analysis of FT/FP/NAD/NC} 
The curves for FT, FP \cite{FP}, NAD \cite{nad-iclr-2020} (its plots will be presented in \textbf{Appendix}) and NC\cite{wang2019neural} are similar since they all use fine-tuning on a small subset of clean data (\ie, $5\%$ training data), thus we present a deep analysis for FT as an example.
As shown Figure \ref{fig:abnormal}, we compare the performance of $5\%$ and $10\%$. We first analyze the changes in the average neuron activation (see Section \ref{sec: codebase}) before and after the defense. As shown in the top row, the changes between \textit{Poisoned+No Defense} (green) and \textit{Poisoned+FT} (purple) in the case of $5\%$ are much smaller than those in the case of $10\%$. It tells that the backdoor is significantly affected by FT. 
We believe the reason is that when the poisoning ratio is not very high (\eg, $5\%$), the model fits clean samples very well, while the fitting gets worse if the poisoning ratio keeps increasing after a threshold ratio. We find that the clean accuracy on the $5\%$ clean data used for fine-tuning by the backdoored model before the defense is $99\%$ in the case of $5\%$ poisoning ratio, while $92\%$ in the case of $10\%$ poisoning ratio. It explains why their changes in neuron activation values are different.

\vspace{-1em}
\paragraph{Analysis of ABL} 
The ABL \cite{li2021anti} method uses the loss gap between the poisoned and clean samples in the early training period to isolate some poisoned samples. We find that the loss gap in the case of high poisoning ratio is larger than that in the case of low poisoning ratio. Take the LC \cite{clean-label-2018} attack on CIFAR-10 as example. In the case of 5$\%$ poisoning ratio, the isolated 500 samples by ABL are 0 poisoned and 500 clean samples, such that the backdoor effect cannot be mitigated in later backdoor unlearning in ABL. In contrast, the isolated 500 samples are all poisoned in the case of 10$\%$ poisoning ratio. The t-SNE visualizations shown in the second row of Figure \ref{fig:abnormal} also verify this point.

\vspace{-1em}
\paragraph{Analysis of ANP} 
The ANP \cite{wu2021adversarial} prunes the neurons that are sensitive to the adversarial neuron perturbation, by setting a threshold. As suggested in \cite{wu2021adversarial}, this threshold is fixed as 0.2 in our evaluations. We find that when the poisoning ratio is high, more neurons will be pruned, thus the ASR may decrease. For example, given the SIG \cite{SIG} attack, the pruned neurons by ANP are 328 and 466 for $5\%$ and $10\%$ poisoning ratios, respectively. 
As shown in the last row of Figure \ref{fig:abnormal}, poisoned samples still gather together for $5\%$, while separated for $10\%$.

\textbf{In summary}, the above analysis demonstrates an interesting point that attack with higher poisoning ratios doesn't mean better attack performance, and it may be more easily defended by some defense methods. The reason is that higher poisoning ratios will highlight the difference between poisoned and clean samples, which will be utilized by adaptive defenses. This point inspires two interesting questions that deserve further exploration in the future: \textit{how to achieve the desired attack performance using fewer poisoned samples, and how to defend weak attacks with low poisoning ratios}. 
Moreover, considering the randomness due to weight initialization and some methods' mechanisms, we repeat the above evaluations several times. Although some fluctuations occur, the trend of ASR curves is similar to that in Figure \ref{2-attack-asr-preact}. More details and analysis are presented in  \textbf{Appendix}.

\vspace{-0em}
\subsection{Effect of model architectures}  
\vspace{-0em}

\begin{figure}[!ht]
    \centering
    \includegraphics[width=1\textwidth]{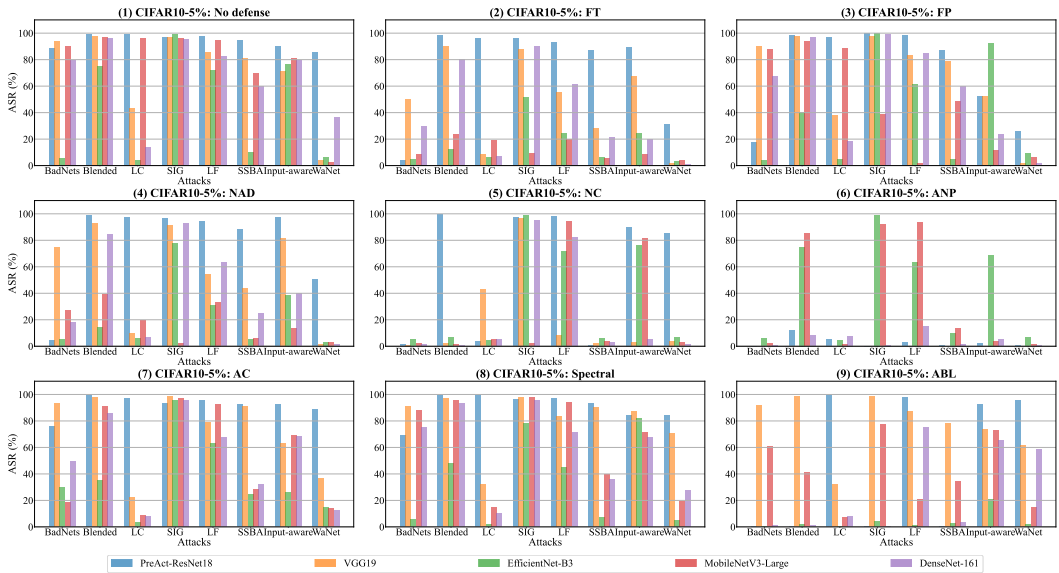}
    \vspace{-0.7em}
    \caption{The effects of different model architectures using  different defense and attack methods.
    }
    \label{3-asr-attack}
\end{figure}

As shown in Figure ~\ref{3-asr-attack}, we analyze the influence caused by model architectures. From the top-left sub-figure, it is worth noting that, under the same training scheme, not all backdoor attacks can successfully plant a backdoor in EfficientNet-B3, such as BadNets, LC, SSBA, and WaNet. In contrast, PreAct-ResNet18 is easy to be planted a backdoor. Besides, we find that most defense methods fail to remove the backdoors embedded in the PreAct-ResNet18 and VGG-19, except ANP. However, ANP is less effective on EfficientNet-B3 attacked by SIG. From the second sub-figure in the first row, we notice that FT is an optimal defense method for MobileNetV3-Large, which could effectively decrease the ASR. In most cases, NC and ANP can remove the backdoors embedded in DenseNet-161. 
The above analysis demonstrates that one attack or defense method may have totally different performance on different model architectures. It inspires us \textit{to further study the effect of model architecture in backdoor learning and to design more robust architectures in the future.}

\subsection{Contents in Appendix}

Due to the space limit, we have put several important contents in the \textbf{Appendix}. Here we present a brief outline of the Appendix to facilitate readers to find the corresponding content, as follows: 
\begin{itemize}[leftmargin=10pt]
    \item Section \ref{appendix: sec additional information of algorithm} Additional information of backdoor attack and defense algorithms:
    \begin{itemize}
        \item Section \ref{appendix: subsec backdoor attack algorithms}: Descriptions of backdoor attack algorithm;
        \item Section \ref{appendix: subsec descriptions of  backdoor defense algorithms}: Descriptions of  backdoor defense algorithms;
        \item Section \ref{appendix: subsec implementation details and computational complexities}: Implementation details and computational complexities.
    \end{itemize}
    \item Section \ref{appendix: sec additional results}: Additional evaluations and analysis:
    \begin{itemize}
        \item Section \ref{appendix: subsec full results of CIFAR-10}: Full results on CIFAR-10;
        \item Section \ref{appendix: subsec result overview}: Results overview; 
        \item Section \ref{appendix: subsec effect of dataset}: Effect of dataset; 
        \item Section \ref{appendix: sec Effect of poisoning ratio}: Effect of poisoning ratio;
        \item Section \ref{appendix: subsec sensitivity test}: Sensitivity to hyper-parameters; 
        \item Section \ref{appendix: sec quick learning}: Analysis of quick learning of backdoor; 
        \item Section \ref{appendix: subsec backdoor forgetting}: Analysis of backdoor forgetting; 
        \item Section \ref{appendix: subsec trigger generalization}: Analysis of trigger generalization of backdoor attacks; 
        \item Section \ref{appendix: subsec vit results}: Evaluation on vision transformer; 
        \item Section \ref{appendix: subsec imagenet results}: Evaluation on ImageNet;
        \item Section \ref{appendix: subsec visualization}: Visualization. 
    \end{itemize}
    \item Section \ref{appendix: sec nlp}: BackdoorBench in Natural Language Processing; 
    \item Section \ref{appendix: sec reproducibility}: Reproducibility; 
    \item Section \ref{appendix: sec license}: License. 
\end{itemize}

\section{Conclusions, limitations and societal impacts}  
\vspace{-0em}

\textbf{Conclusions} We have built a comprehensive and latest benchmark for backdoor learning, including an extensible modular-based codebase with implementations of 8 advanced backdoor attacks and 9 advanced backdoor defense algorithms, as well as 8,000 attack-defense pairs of evaluations and thorough analysis. We hope that this new benchmark could contribute to the backdoor community in several aspects: providing a clear picture of the current progress of backdoor learning, facilitating researchers to quickly compare with existing methods when developing new methods, and inspiring new research problems from the thorough analysis of the comprehensive evaluations.

\textbf{Limitations} 
Until now, BackdoorBench has mainly provided algorithms and evaluations in the computer vision domain and supervised learning. In the future, we plan to expand BackdoorBench to more domains and learning paradigms, \eg, natural language processing (NLP), Speech, and reinforcement learning.

\textbf{Societal impacts} 
Our benchmark could facilitate the development of new backdoor learning algorithms. Meanwhile, like most other technologies, the implementations of backdoor learning algorithms may be used by users for good or malicious purposes. The feasible approach to alleviate or avoid adverse impacts could be exploring the intrinsic property of the technology, regulations, and laws.

\section{Acknowledgement}

This work is supported by the National Natural Science Foundation of China under grant No.62076213, Shenzhen Science and Technology Program under grant No.RCYX20210609103057050, and the university development fund of the Chinese University of Hong Kong, Shenzhen under grant No.01001810. 
Chao Shen is supported by the National Key Research and Development Program of China (2020AAA0107702), National Natural Science Foundation of China (U21B2018, 62161160337, 62132011), Shaanxi Province Key Industry Innovation Program (2021ZDLGY01-02).

{\small
\bibliographystyle{plain}
\bibliography{backdoor}
}

\newpage 
\section*{Checklist}

\begin{enumerate}

\item For all authors...
\begin{enumerate}
  \item Do the main claims made in the abstract and introduction accurately reflect the paper's contributions and scope?
    \answerYes{}
  \item Did you describe the limitations of your work?
    \answerYes{}
  \item Did you discuss any potential negative societal impacts of your work?
    \answerYes{}
  \item Have you read the ethics review guidelines and ensured that your paper conforms to them?
    \answerYes{}
\end{enumerate}

\item If you are including theoretical results...
\begin{enumerate}
  \item Did you state the full set of assumptions of all theoretical results?
    \answerNA{}
	\item Did you include complete proofs of all theoretical results?
    \answerNA{}
\end{enumerate}

\item If you ran experiments (e.g. for benchmarks)...
\begin{enumerate}
  \item Did you include the code, data, and instructions needed to reproduce the main experimental results (either in the supplemental material or as a URL)?
    \answerYes{} See the Github repository of BackdoorBench ( \url{https://github.com/SCLBD/BackdoorBench}
  \item Did you specify all the training details (e.g., data splits, hyperparameters, how they were chosen)?
    \answerYes{} See Tables 1 and 2 in Supplementary Material.
	\item Did you report error bars (e.g., with respect to the random seed after running experiments multiple times)?
    \answerNA{}
	\item Did you include the total amount of compute and the type of resources used (e.g., type of GPUs, internal cluster, or cloud provider)?
    \answerNo{}
\end{enumerate}

\item If you are using existing assets (e.g., code, data, models) or curating/releasing new assets...
\begin{enumerate}
  \item If your work uses existing assets, did you cite the creators?
    \answerYes{} The implementations of some existing algorithms are modified based on their original source codes, and we clearly describe the original link and our modifications in each code file in the Github repository of BackdoorBench (see \url{https://github.com/SCLBD/BackdoorBench}).
  \item Did you mention the license of the assets?
    \answerYes{}
  \item Did you include any new assets either in the supplemental material or as a URL?
    \answerYes{} See the Github repository of BackdoorBench ( \url{https://github.com/SCLBD/BackdoorBench}.
  \item Did you discuss whether and how consent was obtained from people whose data you're using/curating?
    \answerNA{}
  \item Did you discuss whether the data you are using/curating contains personally identifiable information or offensive content?
    \answerNA{}
\end{enumerate}

\item If you used crowdsourcing or conducted research with human subjects...
\begin{enumerate}
  \item Did you include the full text of instructions given to participants and screenshots, if applicable?
    \answerNA{}
  \item Did you describe any potential participant risks, with links to Institutional Review Board (IRB) approvals, if applicable?
    \answerNA{}
  \item Did you include the estimated hourly wage paid to participants and the total amount spent on participant compensation?
    \answerNA{}
\end{enumerate}

\end{enumerate}


\newpage 

\appendix

\section{Additional information of backdoor attack and defense algorithms}
\label{appendix: sec additional information of algorithm}

\subsection{Descriptions of backdoor attack algorithms}
\label{appendix: subsec backdoor attack algorithms}

In addition to the basic information in Table 1 of the main manuscript, here we describe the general idea of eight implemented backdoor attack algorithms in BackdoorBench, as follows. 
\begin{itemize}[leftmargin=10pt]
    \item \textbf{BadNets} \cite{gu2019badnets}: It was the first work in backdoor learning, which simply inserted a small patch with fixed pattern and location to replace the original pixels in the clean image to obtain a poisoned image. 
    \item \textbf{Blended} \cite{chen2017targeted}: It extended BadNets by encouraging the invisibility of the trigger through alpha blending. 
    \item \textbf{Label consistent (LC)} \cite{clean-label-2018}: It generated a poisoned image using adversarial attack, by enforcing it to be close to the clean target image in the original RGB space, and close to the clean source image patched with a trigger in the feature space of a pre-trained clean model. Since the poisoned image is labeled as the target class, the mapping from the trigger to the target class could be learned. 
    \item \textbf{SIG} \cite{SIG}: It adopted a sinusoidal signal as the trigger to perturb the clean images of the target class, while not changing their labels, such that achieving the label consistent backdoor attack. 
    \item \textbf{Low frequency attack (LF)} \cite{zeng2021rethinking} : It was built upon an analysis that the triggers in many backdoor attacks bring in high-frequency artifacts, which are easily detectable. Inspired by this analysis, LF developed a smooth trigger by filtering high-frequency artifacts from a universal adversarial perturbation. 
    \item \textbf{Sample-specific backdoor attack (SSBA)} \cite{ssba}: It utilized an auto-encoder to fuse a trigger (\eg, a string) into clean samples to obtain poisoned samples. The residual between the poisoned and the clean sample varied for different clean images, \ie, sample-specific. 
    \item \textbf{Input-aware dynamic backdoor attack (Input-aware)} \cite{nguyen2020input}: It was a training-controllable attack by simultaneously learning the model parameters and a trigger generator. When testing, the learned trigger generator generated one unique trigger for each clean testing sample.  
    \item \textbf{Warping-based poisoned networks (WaNet)} \cite{nguyen2021wanet}: It was also a training-controllable attack. A fixed warping function is adopted to slightly distort the clean sample to construct the poisoned sample. The attacker further controlled the training process to ensure that only the adopted fixed warping function can activate the backdoor. 
\end{itemize}

\subsection{Descriptions of  backdoor defense algorithms}
\label{appendix: subsec descriptions of  backdoor defense algorithms}

In addition to the basic information in Table 2 of the main manuscript, here we describe the general idea of nine implemented backdoor defense algorithms in BackdoorBench, as follows.

\begin{itemize}[leftmargin=10pt]
    \item \textbf{Fine-tuning (FT)}: It is assumed that fine-tuning the backdoored model on a subset of clean samples could mitigate the backdoor effect. Note that FT is a widely used approach for transferring pre-trained models to new tasks, but it has been used as a basic component in several backdoor defense methods, such as Fine-pruning (FP) \cite{FP}, Neural Attention Distillation (NAD) \cite{nad-iclr-2020}. 
    \item \textbf{Fine-pruning (FP)} \cite{FP}: It is built upon the assumption that \textit{poisoned and benign samples have different activation paths in the backdoored model}. Inspired, FP proposed to firstly prune some inactivated neurons of clean samples, then fine-tune the pruned model based on the subset of benign samples to recover the model performance. 
    \item \textbf{Neural Attention Distillation (NAD)} \cite{nad-iclr-2020}: Its assumption is same with FT. Instead of directly use the fine-tuned model as the mitigated model, NAD adopts the first fine-tuned model as a teacher, and fine-tunes the backdoored model again by encouraging the consistency of the attention representation between the new fine-tuned model and the teacher model. 
    \item \textbf{Neural cleanse (NC)} \cite{wang2019neural}: It is built upon the assumption that \textit{the trigger provides a ``shortcut" between the samples from different source classes and the target class}. Based on this assumption, the possible trigger is searched through optimization. If a small-size trigger (\eg, a small patch in the image) is found, then the model is detected as backdoored model, which is then mitigated through pruning based on the searched trigger. 
    \item \textbf{Adversarial Neuron Pruning (ANP)} \cite{wu2021adversarial}: It is built upon an observation that \textit{the neurons related to the injected backdoor are more sensitive to adversarial neuron perturbation (\ie, perturbing the neuron weight to achieve adversarial attack) than other neurons in a backdoored model}. Inspired by this, ANP proposed to prune these sensitive neurons for backdoor mitigation. 
    \item \textbf{Activation Clustering (AC)} \cite{AC}: It is built upon an observation that \textit{the sample activations (\ie, the feature presentations) of the target class will form two clusters, and the smaller cluster corresponds to poisoned samples, while those of other classes form one cluster}. Then, the model is trained from scratch based on the dataset without poisoned samples.
    \item \textbf{Spectral Signatures (SS)} \cite{tran2018spectral}: Its assumption is that \textit{the feature representation distributions of benign and poisoned samples in one class are spectrally separable}, which is a concept of robust statistics. Consequently, the poisoned samples can be identified through analyzing the spectrum of the covariance matrix of the feature representations. Then, the model is retrained from scratch by removing the poisoned samples from the training set.
    \item \textbf{Anti-Backdoor Learning (ABL)} \cite{li2021anti}: It is built upon an observation that \textit{the loss values of poisoned samples drops much faster than those of benign samples in early epochs during the training process}. Inspired, ABL proposed to firstly isolate poisoned samples from benign samples according to their difference on loss dropping speed, then mitigate the backdoor effect by maximizing the loss of the isolated poisoned samples. 
    \item \textbf{Decoupling-based Backdoor Defense (DBD)} \cite{huang2022backdoor}: It is built upon an observation that \textit{poisoned samples from different samples will gather together in the feature space of a backdoored model}. DBD proposed to prevent the gathering by learning the model backbone through self-supervised learning without labels, rather than the standard supervised learning. Then, since the poisoned samples are separated, their loss values are larger than benign samples when learning the classifier, such that samples with large loss values can be identified as poisoned samples. Finally, the labels of poisoned samples are abandoned, and a semi-supervised fine-tuning of both the backbone and classifier is conducted to improve the model performance. 
\end{itemize}

\begin{table}[t]{\footnotesize 
        \centering
        \caption{Hyper-parameter settings of all implemented attack methods.} 
        \label{tab:attack_param}
        \renewcommand{\arraystretch}{1.2}
        \setlength{\tabcolsep}{2pt}
        \scalebox{0.84}{
        \begin{tabular}{c|l|l|l}
        \hline
             {Attack}                    & {Hyper-parameter}                              & {Setting}            & Theoretical complexity   \\
            \hline
            \hline
            \multirow{12}{*}{General Settings} & attack target            &  all-to-one with class 0          &   train sample size($ N_{\mathcal{T}} $), \\
                                      & optimizer                 & SGD         &   batch size ($ B $),\\
                                      & momentum                  & 0.9                     &  forward process for net C ($ f_{C}  $), \\
                                      & weight decay              & 0.0005                    &  backward process for net C ($ b_{C} $), \\
                                      & batch size             & 128                   &  epochs ($ E $) , \\
                                      & epochs on CIFAR10 and CIFAR-100 & 100&  number of classes ( $ N_{cls} $)\\
                                      & epochs on GTSRB & 50&  \\
                                      &lr schedule (except for &CosineAnnealingLR&  \\
                                      & training-controllable attack) on &&\\
                                      
                                      & CIFAR10, CIFAR-100 and GTSRB&&  \\
                                      & epochs on TinyImageNet-200 & 200&  \\
                                      & lr schedule  on TinyImageNet-200&ReduceLROnPlateau&  \\
                                      & random seed & 0 & \\
                                                  \hline
            
            \multirow{2}{*}{BadNets \cite{gu2019badnets}}      &  pattern \& location       & $3\times3$, pure white,  at &  $ O(N_{\mathcal{T}} / B  E  (f + b)) $ \\
             																								&&downright corner       &  \\
             																								&&(no margin left)             &\\
             																								
                        \hline

            \multirow{2}{*}{Blended \cite{chen2017targeted}}      & pattern        & hello kitty               &  $ O(N_{\mathcal{T}} / B  E  (f + b))$  \\
             & alpha      & 0.2                &  \\
            
            \hline
            
            \multirow{4}{*}{Label Consistent \cite{turner2019labelconsistent}}      &  adversarial attack &  PGD &   $ O(N_{\mathcal{T}} / B  E  (f + b) + N_{\mathcal{T}} / B  T_{step}  (f+ b) )$ \\
                         													& step ($ T_{step} $)      & 100                &  \\
                         													& $ \alpha$      & 1.5                &  \\
        													                 & $ \epsilon$      & 8              &  \\
                        \hline

            \multirow{2}{*}{SIG \cite{SIG}}     & $ \delta $        & 40          &  $ O(N_{\mathcal{T}} / B  E  (f + b)) $ \\ 
                                      & $ f $             & 6                       &  \\
                                      
            \hline

            \multirow{6}{*}{Low Frequency \cite{zeng2021rethinking}}     &maximum number of termination& 50  &     $ O(T_{ter}  N_{sample}  (2f+ T_{df}f + T_{df}  N_{cls} b) +  $      \\
            																	& iteration   ($ T_{ter}$)     &&  $N_{\mathcal{T}} / B  E  (f + b))$\\
            																	
														            &fooling rate&0.2&  \\
														              &overshoot        & 0.02                    &  \\
														              &maximum number of iterations       & 200                &  \\
														              &for deepfool  ($ T_{df} $)&&\\
														              
														              														              &sample number for UAP ($ N_{sample} $)& 100&  \\

            \hline

            \multirow{1}{*}{SSBA \cite{ssba}}      & encoded bit          & 1       &  $ O(T_{step} (f_{auto} + b_{auto}) + N_{\mathcal{T}} / B  E  (f + b))  $\\ 
            &autoencoder train step ($T_{step}$)& 140000& \\
            \hline

            \multirow{10}{*}{Input-aware \cite{nguyen2020input}}     & Generator lr  (both for M, G)              & 0.01            &  $O( E_{mask} N_{\mathcal{T}} / B (f_{mask} + b_{mask}) +  $\\
            														& schedule (for M, G  and C) & MultiStepLR & $(E - E_{mask} ) N_{\mathcal{T}} / B  (2f_{mask} +$\\
            														& schedule milestones for G             &200, 300, 400, 500            & $  2f_{generator} + 2b_{generator} + f + b) )$ \\
            														& schedule milestones for C             &100, 200, 300, 400           &  \\
            														& schedule milestones for M             & 10, 20         &  \\
            														& schedule gamma (for M and G)             & 0.1         &  \\
           																& $ \lambda_{div} $& 1&  \\
           														            & $ \lambda_{norm} $& 400&  \\
           														            &mask density& 0.032&  \\
           														            &  cross\_ratio&  1 &  \\
           														            &  mask train epochs ($ E_{mask} $)&  25 &  \\
            \hline

            \multirow{3}{*}{WaNet \cite{nguyen2021wanet}}     & cross\_ratio      & 2                     &  $ O(N_{\mathcal{T}} / B  E  (f + b)) $ \\
            							&lr schedule&MultiStepLR&  \\
            							& schedule milestones &100, 200, 300, 400           &  \\
                                      & grid\_rescale                   & 1        &  \\
                                      \hline
        \end{tabular}}
        }
\end{table}

\begin{table}[t]{\footnotesize 
        \centering
        \caption{Hyper-parameter settings of all implemented defense methods. } 
        \label{tab:defense_param}
        \renewcommand{\arraystretch}{1.2}
        \setlength{\tabcolsep}{2pt}
        \scalebox{0.68}{
        \begin{tabular}{c|l|l|l}
        \hline
             {Defense}                    & {Hyper-parameter}                              & {Setting}  &{Theoretical complexity }           \\
            \hline
            \hline
            \multirow{5}{*}{General Settings} 
                                      & optimizer                 & SGD     &   train sample size ($ N_{\mathcal{T}} $), batch size ($ B $),    \\
                                      & momentum                  & 0.9   &   forward process for net C ($ f_{C}  $),                   \\
                                      & weight decay              & 0.0005     &backward process for net C ($ b_{C} $),                 \\
                                      & batch size             & 256             &  epochs ($ E $) , number of classes ( $ N_{cls} $)       \\
                                      & epochs on CIFAR10, CIFAR-100 and GTSRB & 100 & the number of pruning neurons ($N_{neu}$) \\
                                      &lr schedule (except for special learning defense)  &CosineAnnealingLR &\\
                                      &on CIFAR10, CIFAR-100 and GTSRB& &\\
                                      & epochs on TinyImageNet-200 & 200 &\\
                                      & lr schedule  on TinyImageNet-200&ReduceLROnPlateau&\\
                                      &the number of pruning neurons  & the number of neurons in the last layer & \\
                                      & random seed & 0 & \\
                                                  \hline
            FT     & ratio of validation data ($p_{v}$)          & 5$\%$ & $O(N_{\mathcal{T}}p_{v}/ B  E  (f + b))$             \\
            \hline
            \multirow{2}{*}{FP \cite{FP}}   & ratio of validation data ($p_{v}$)         & 5$\%$   & $O(N_{neu}N_{\mathcal{T}}p_{v} f / B  ) $          \\
            & the tolerance of accuracy reduction\footnotemark[1]
      & 10$\%$  & $+ O(N_{\mathcal{T}}p_{v} / B  E  (f + b))$             \\
            \hline
            \multirow{2}{*}{NAD \footnotemark[3] \cite{nad-iclr-2020}}   
             & ratio of validation data for teacher model ($p_{v}$)       & 5$\%$  & $O(N_{\mathcal{T}}p_{v} / B  E_{ft}  (f + b))$             \\
            & $\beta_1$ for the loss & 500 &$+O(N_{\mathcal{T}}p_{v} / B  E  (2f + b))$\\
    & $\beta_2$ for the loss & 1000\\
    & $\beta_3$ for the loss & 1000\\
    & the power for attention & 2.0 \\
    & the epoch for teacher model to fine-tune ($E_{ft}$) & 10\\
            \hline
            \multirow{2}{*}{NC \cite{wang2019neural}}   &the norm used for the reversed trigger & L1 & $O(N_{\mathcal{T}}p_{v}  E_{r} N_{cls} /B (f + b) )$\\
    &cleaning ratio ($p_{v}$) & 0.05 & $+O(N_{\mathcal{T}}p_{v} / B  E  (f + b))$\\
    &unlearning ratio & 0.2\\
    &the epoch of learning trigger ($E_{r}$) & 80\\
            \hline
            
            \multirow{4}{*}{ANP \cite{wu2021adversarial}}  & the tolerance of accuracy reduction\footnotemark[1]     & 10$\%$ & \multirow{4}{*}{$O(e_{i}(f + b))$}   \\
        & number of validation data ($p_{v}$)         & 5$\%$              \\
        &  the number of iteration during pertubation ($e_{i}$) & 2000 \\
        & $\epsilon$ & 0.4\\
        & $\alpha$ & 0.2 \\
            \hline
            AC \cite{AC}     & number of reduced dimensions         & 10     & $O(N_{\mathcal{T}} / B N_{fe}^{3})+O(N_{\mathcal{T}} / B  E  (f + b))$\footnotemark[2]         \\
            \hline
            Spectral \cite{tran2018spectral}   & The percentile of backdoor data & 85$\%$  & $O(N_{fe}^{3}) + O(N_{\mathcal{T}} / B  E_{d}  (f + b))$\footnotemark[2]            \\
            
            \hline
            
            \multirow{6}{*}{ABL \cite{li2021anti}}     &  tuning epochs ($E_{tu}$) for CIFAR10, CIFAR-100 and GTSRB & 20 &$O(N_{\mathcal{T}} / B  E_{tu}  (f + b)) $\\
             					&finetuning epochs ($E_{ft}$) for CIFAR10, CIFAR-100 and GTSRB & 60   & $+O(N_{\mathcal{T}}(1-p_{i}) / B  E_{ft}  (f + b))$        \\
             						&unlearning epochs ($E_{u}$) for CIFAR10, CIFAR-100 and GTSRB & 20 &  $+O(N_{\mathcal{T}}p_{i} / B  E_{u}  (f + b))$       \\
             						&  tuning epochs for Tiny & 40\\
             					&finetuning epochs for Tiny & 120           \\
             						&unlearning epochs for Tiny & 4          \\
             						&lr for unlearning & 0.0005\\
             						&the value of flooding & 0.5 \\
             						&the isolation ratio of training data ($p_{i}$) & 0.01\\
                        \hline
            
            \multirow{4}{*}{DBD \cite{huang2022backdoor}}  & the epoch for self-supervised learning ($E_{se}$)& 100&$O(N_{\mathcal{T}}/B_{se} E_{se}(f + b))$ \\
& the epoch for warmup ($E_{wa}$) & 10 &$+O(N_{\mathcal{T}}/B_{semi}E_{se}(f + b))$ \\

&  the epsilon for the dataset ($\epsilon$)  & 0.5 &$+O(N_{\mathcal{T}}\epsilon/B_{semi}E(f + b))$ \\
& during the semi-supervised learning & &\\
& The batch size of self learning ($B_{self}$) & 512\\
& The batch size of self learning ($B_{semi}$) & 128\\
            \hline
        \end{tabular}}
        }
\end{table}
\footnotetext[1]{For FP and ANP, we define a hyper-parameter \textit{the tolerance of clean accuracy reduction} as the maximum relative drop of clean accuracy. It is used to determine the number of pruned neurons.}
\footnotetext[2]{For AC and Spectral, $N_{fe}$ is the dimensions of the representation.} 
\footnotetext[3]{For NAD, we use the code and recommended hyper-parameters at \href{https://github.com/bboylyg/NAD/tree/d61e4d74ee697f125336bfc42a03c707679071a6}{https://github.com/bboylyg/NAD/tree/d61e4d74ee697f125336bfc42a03c707679071a6}.} 

\subsection{Implementation details and computational complexities}
\label{appendix: subsec implementation details and computational complexities}

\textbf{Running environments} Our evaluations are conducted on GPU servers with 2 Intel(R) Xeon(R) Platinum 8170 CPU @ 2.10GHz, RTX3090 GPU (32GB) and 320 GB RAM (2666MHz).
The versions of all involved softwares/packages are clearly described in the README file of the Github repository (see \url{https://github.com/SCLBD/BackdoorBench}). Here we didn't repeat the descriptions. 

\textbf{Hyper-parameter settings} The hyper-parameter settings adopted in our evaluations about backdoor attack and defense algorithms are described in Table \ref{tab:attack_param} and Table \ref{tab:defense_param}, respectively. With these hyper-parameter settings, the reported results of 8,000 pairs of evaluations could be reproduced. 
\blue{
Moreover, we would like to explain our rules to adopt above settings, as follows:
\begin{itemize}[leftmargin=10pt]
    \item \textbf{We don't perform a separate hyper-parameter search for each method}, mainly due to the following two reasons:
    \begin{itemize}
        \item As shown in Tables \ref{tab:attack_param} and \ref{tab:defense_param}, most methods have several hyper-parameters. For most hyper-parameters of a method, there is neither a good rule to determine the values, nor a suitable range of the values suggested in its original manuscript. And, the suitable value or range of each hyper-parameter may vary across different datasets, different model architectures, different against attack/defense methods. Consequently, the hyper-parameter search space for each method could be very large, requiring unimaginably high computational resource.
        \item Even assuming sufficient computing resources, then we can search a good value for each hyper-parameter of each method in each evaluation. However, the comparison results and analysis based on sufficient hyper-parameter search may be unfair and make no sense in practice. Because, we still cannot tell a rule or even some experiences to determine the hyper-parameter values in practice. The sensitivity to hyper-parameters should also be an important metric of one method's performance, not just the best ACC/ASR values through the sufficient hyper-parameter search.
    \end{itemize}
    \item \textbf{How do we set the hyper-parameter values in our current 8000 pairs of evaluations.}
    \begin{itemize}
        \item If the original paper has provided the suggested good values of some hyper-parameters, then we adopt those values in our evaluations. For example, the ANP defense method explicitly wrote that "the perturbation budget $\epsilon = 0.4$ and the trade-off coefficient $\alpha=0.2$", so we also adopt these values in our evaluations.
        \item For those hyper-parameters without suggested values/ranges (or even without descriptions) in their original papers, we will search values that lead to comparable results (ACC/ASR) with the reported results in the same setting (\ie,  same dataset, same/similar model architecture, same poisoning ratio), then fix these values in evaluations of other settings (\eg, changing the poisoning ratio).
        \item The consistent values of hyper-parameters of each method across different settings somewhat guarantee the fairness of evaluations. And, since the adopted values may not be the optimal ones for some hyper-parameters, we didn't conduct the fine-grained analysis about the effects of some specific hyper-parameters (\eg, the trigger size/location in attack methods with patch based triggers). Instead, we provided some high-level analysis \wrt the shared hyper-parameters in all methods (\eg, the number of classes, the poisoning ratio, the model architecture). The findings of these high-level analysis will not be significantly affected by the particular hyper-parameters of each individual method.
    \end{itemize}
\end{itemize}
}

\textbf{Computational complexities} 
The computational complexity of each attack and each defense algorithm is also described in Table \ref{tab:attack_param} and Table \ref{tab:defense_param}, respectively.

\section{Additional evaluations and analysis}
\label{appendix: sec additional results}

\subsection{Full results on CIFAR-10}
\label{appendix: subsec full results of CIFAR-10}

The full results on CIFAR-10 with five different poisoning ratios (\ie, 10\%, 5\%, 1\%, 0.5\%, 0.1\%) are presented in Tables \ref{tab:cifar10_0.1} -- \ref{tab:cifar10_0.001}, respectively. 
The remaining results on other datasets and model architectures among 8,000 attack-defense pairs of evaluations are presented in the Leaderborad in the BackdoorBench website (see \url{https://backdoorbench.github.io}).

\subsection{Results overview} 
\label{appendix: subsec result overview}

\begin{figure}[!ht]
    \centering
    \subfigure[Attack-defense pairs with PreAct-ResNet18 and 10\% poisoning ratio on CIFAR-10.]{
    \begin{minipage}[b]{0.84\textwidth}
    \includegraphics[width=0.9\textwidth]{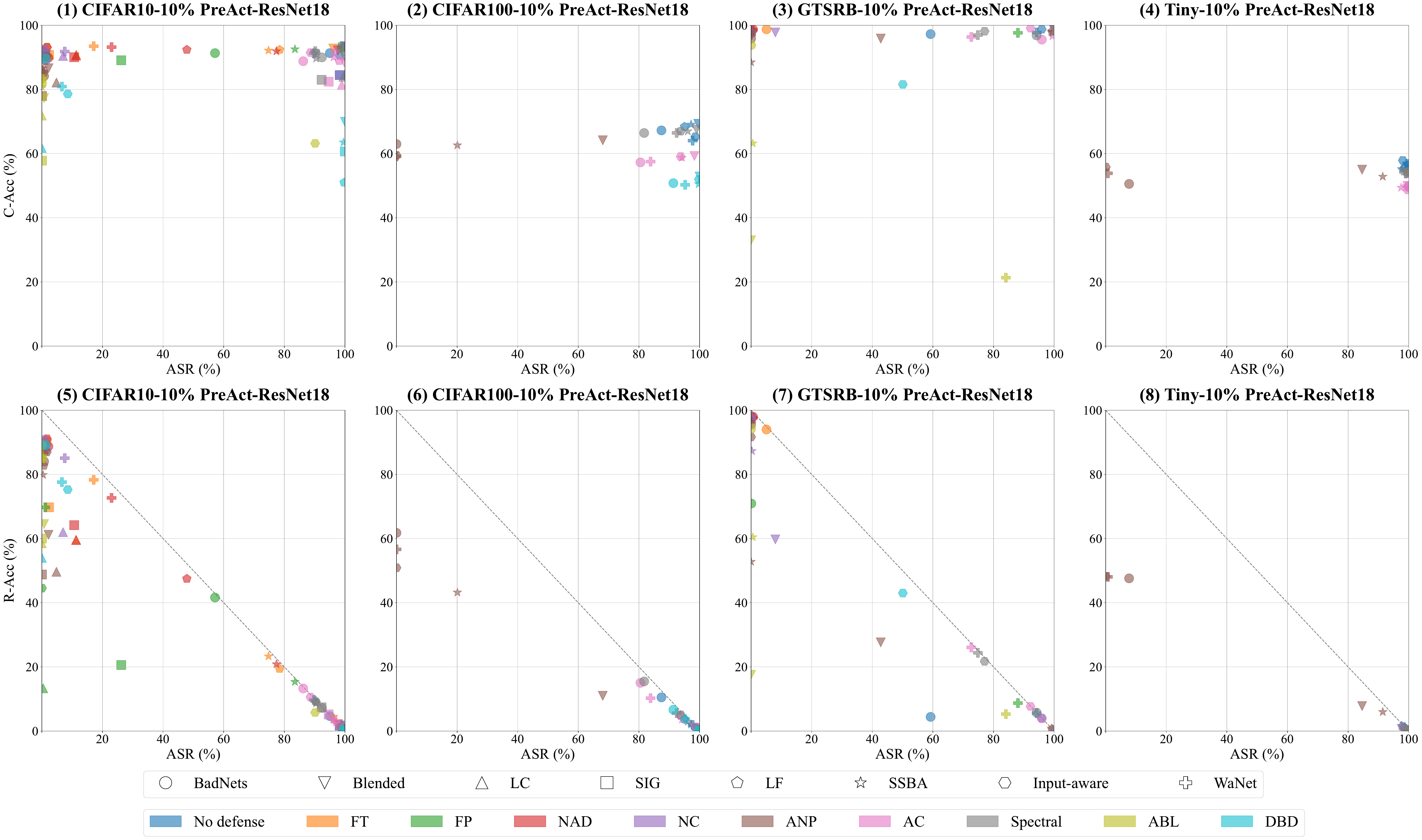}
    \end{minipage}
    }
    \subfigure[Attack-defense pairs with VGG-19 and 10\% poisoning ratio on CIFAR-10.]{
    \begin{minipage}[b]{0.84\textwidth}
    \includegraphics[width=0.9\textwidth]{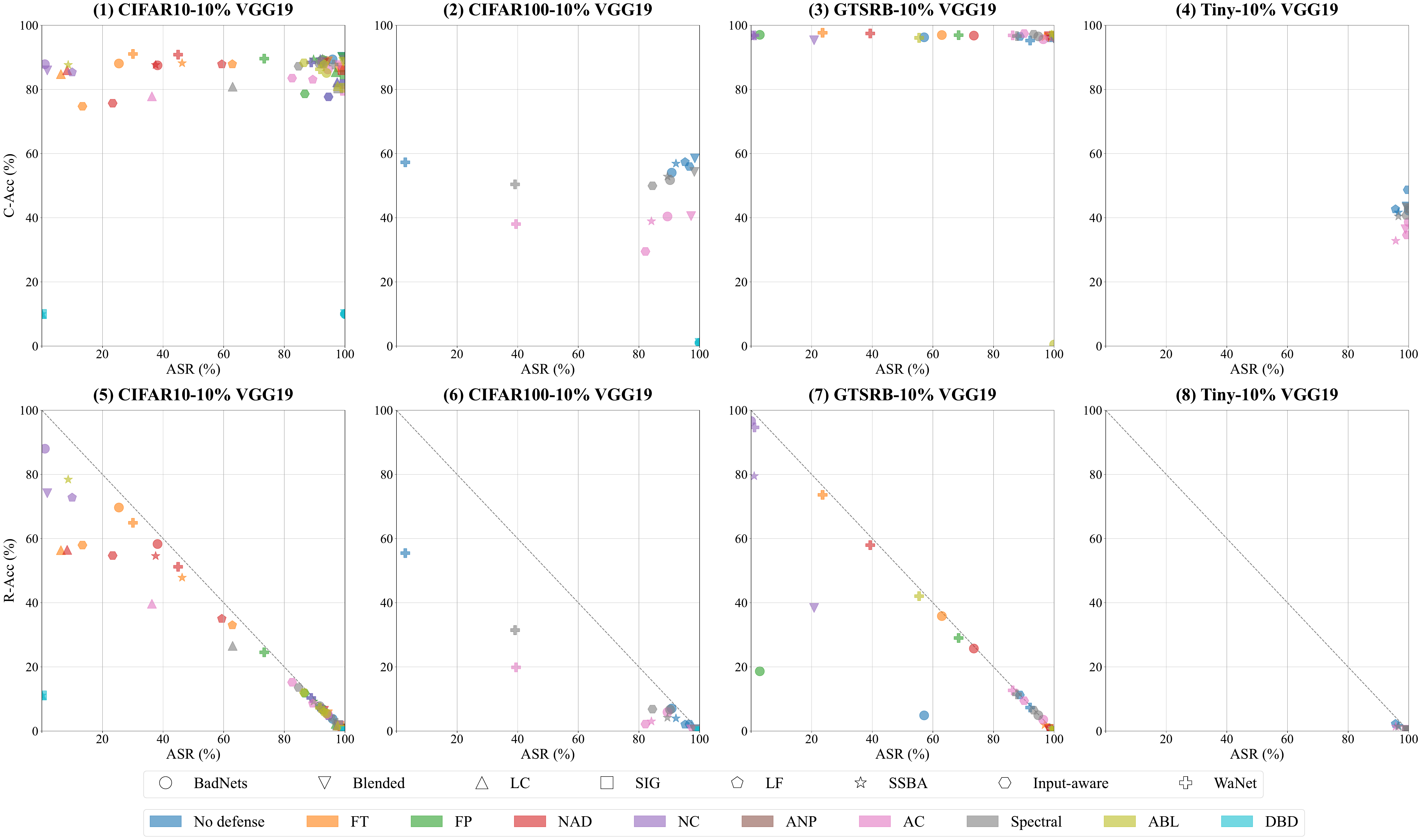}
    \end{minipage}
    }
    \subfigure[Attack-defense pairs with VGG-19 and 5\% poisoning ratio on CIFAR-10.]{
    \begin{minipage}[b]{0.84\textwidth}
    \includegraphics[width=0.9\textwidth]{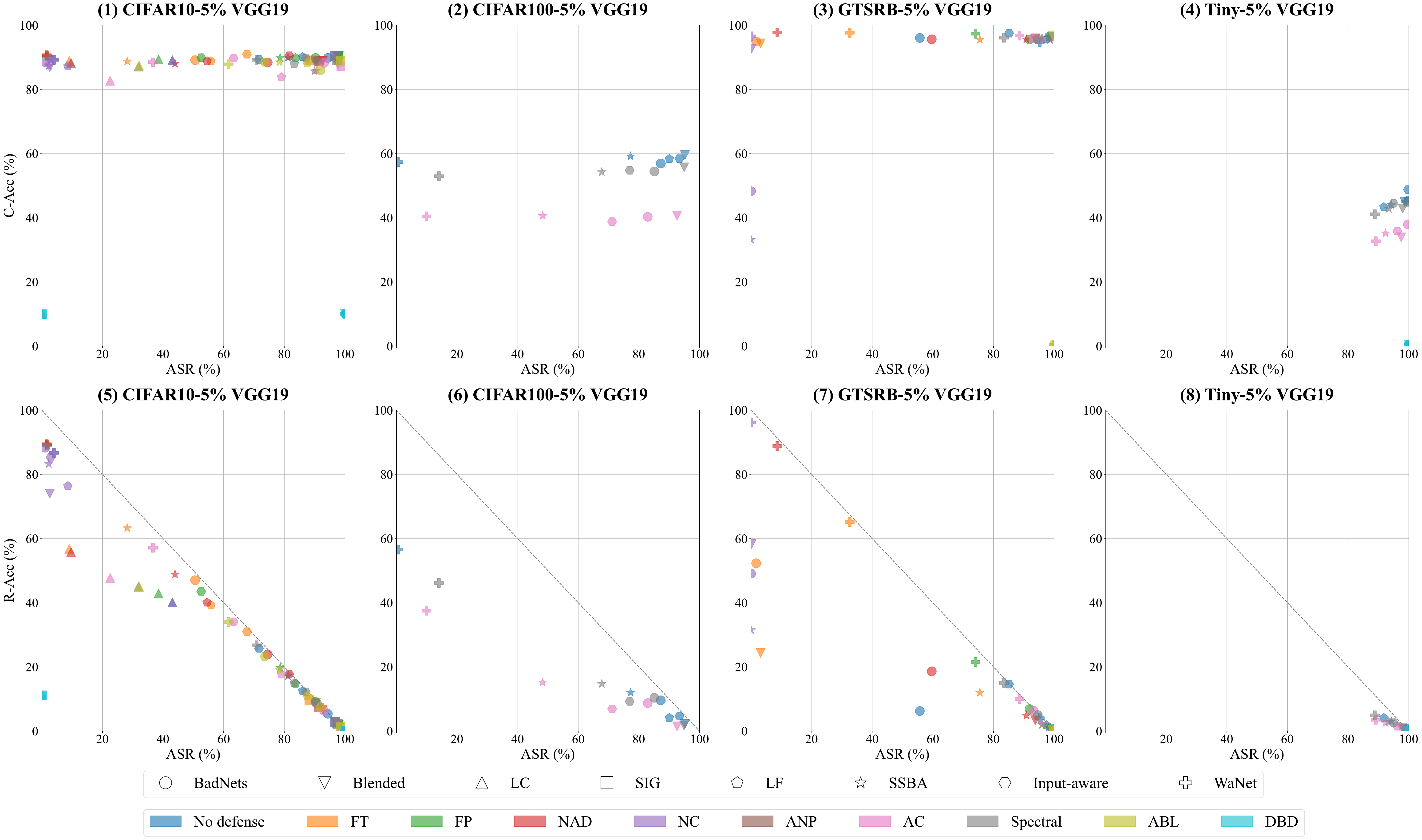}
    \end{minipage}
    }
    \caption{Performance distribution of attack-defense pairs on different model structure and poisoning ratios. A successful attack method should be high C-Acc and ASR, while a successful defense method should be high C-Acc and low ASR. Besides, if the reduced ASR value equals to the increased R-Acc, the color patters would be close to the anti-diagonal line.}
    \label{scatters}
\end{figure}

In Figure ~\ref{scatters}, we present the performance distribution of attack-defense pairs on Preact-ResNet18 and VGG-19 with two poisoning ratios of 5\% and 10\%, respectively. As we mentioned in Section 4.2, if we measured the effectiveness of methods by clean accuracy (C-Acc) and ASR, a perfect attack method should be located at the top-right corner; the perfect defense method should show in the top-left corner. If we measured robust accuracy (R-Acc) and ASR, the reduced ASR value would be desirable to equal the increased R-Acc. This defense method can recover the correct prediction and eliminate the backdoor successfully. Even if we change the model structure from Preact-ResNet18 to VGG-19, the conclusion coincides with our analysis. Besides, with the increase in poisoning ratio, some color patterns are closer to the anti-diagonal line, which means these defense methods can achieve better performance in this situation. 
Please refer to Section 4.2 in the manuscript for the analysis. 

\subsection{Effect of dataset}
\label{appendix: subsec effect of dataset}

\begin{figure}[!ht]
    \centering
    \includegraphics[width=\textwidth]{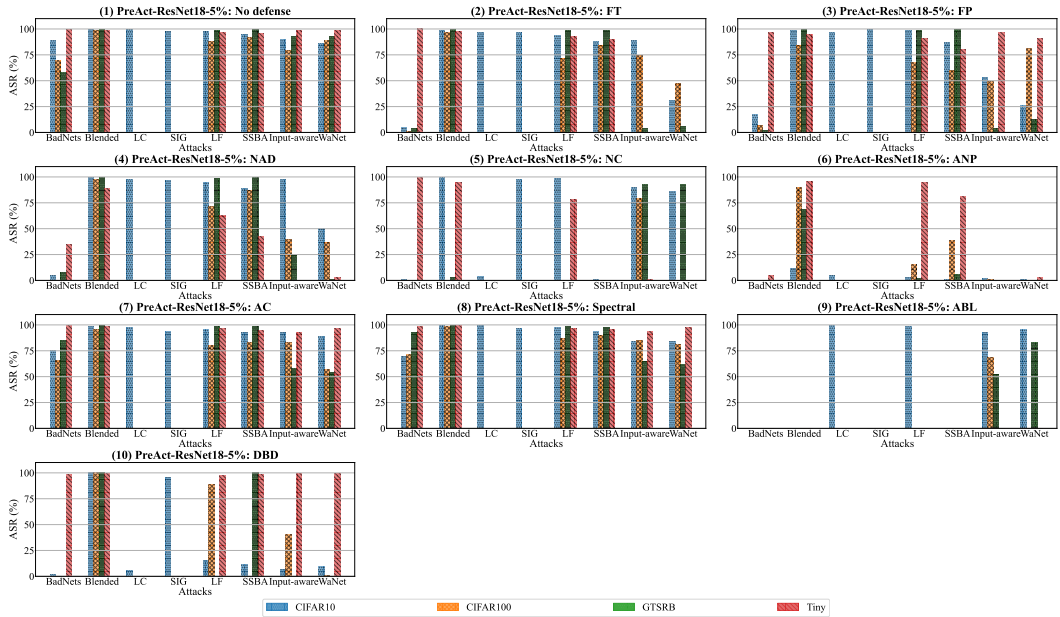}
    \caption{The effects of different datasets on backdoor learning. Note that for the clean-label (\ie, LC \cite{clean-label-2018} and SIG \cite{SIG}) attack, the number of poisoned samples must be less than the target class size, thus it may be not applied to the case of high poisoning ratios.)}
    \label{fig:dataset_influence}
\end{figure}

As shown in Figure ~\ref{fig:dataset_influence}, we make a detailed comparison of the performance of attacks and defenses under different datasets using the PreAct-ResNet18 model and 5\% poison ratio. Where the different colored bars correspond to the four datasets, the height of the bars represents the ASR, and the various subplots correspond to the multiple defenses (and no defenses). Looking down from the undefended perspective, we can see that, by and large, the effect of the attack fluctuates across the different datasets. Blended is the most stable across datasets, while BadNets has the most fluctuating effect across datasets. For BadNets, we find that CIFAR-100 and GTSRB are more complex than CIFAR-10, which leads to the decrease in effectiveness on these two datasets, but the ASR on Tiny ImageNet has rebounded significantly due to the enlargement of the trigger size. From different defense perspectives, we can find that the two methods, AC and Spectral Signature, are relatively unaffected by changes in the dataset compared with each other. In contrast, the rest of the defense methods may all have large fluctuations in their effectiveness in the face of specific attacks. Although fluctuating, ANP has better results on CIFAR-10 for all attack methods, while ABL is also very effective on Tiny ImageNet for all attack methods.

\subsection{\blue{Effect of poisoning ratio}}
\label{appendix: sec Effect of poisoning ratio}

\subsubsection{Effect of poisoning ratio with randomness}

\blue{
As demonstrated in Section 4.3 in the main manuscript, in the following we will further verify the abnormal phenomenon of poisoning ratio's effect shown in Figure 3 in the main manuscript, under the randomness of weight initialization and some methods' mechanisms. 
}


\blue{
\paragraph{Experimental setting}
In the reported 8,000 pairs of evaluations, we set the random seed as 0 to fix all randomness in each evaluations, to ensure all results could be reproduced. 
For each evaluation plotted in Figure 3 in the main manuscript, we re-run the script with five different random seeds, and record the mean and the standard deviation of these five evaluations.}

\begin{figure}[!ht]
    \centering
    \includegraphics[width=\textwidth]{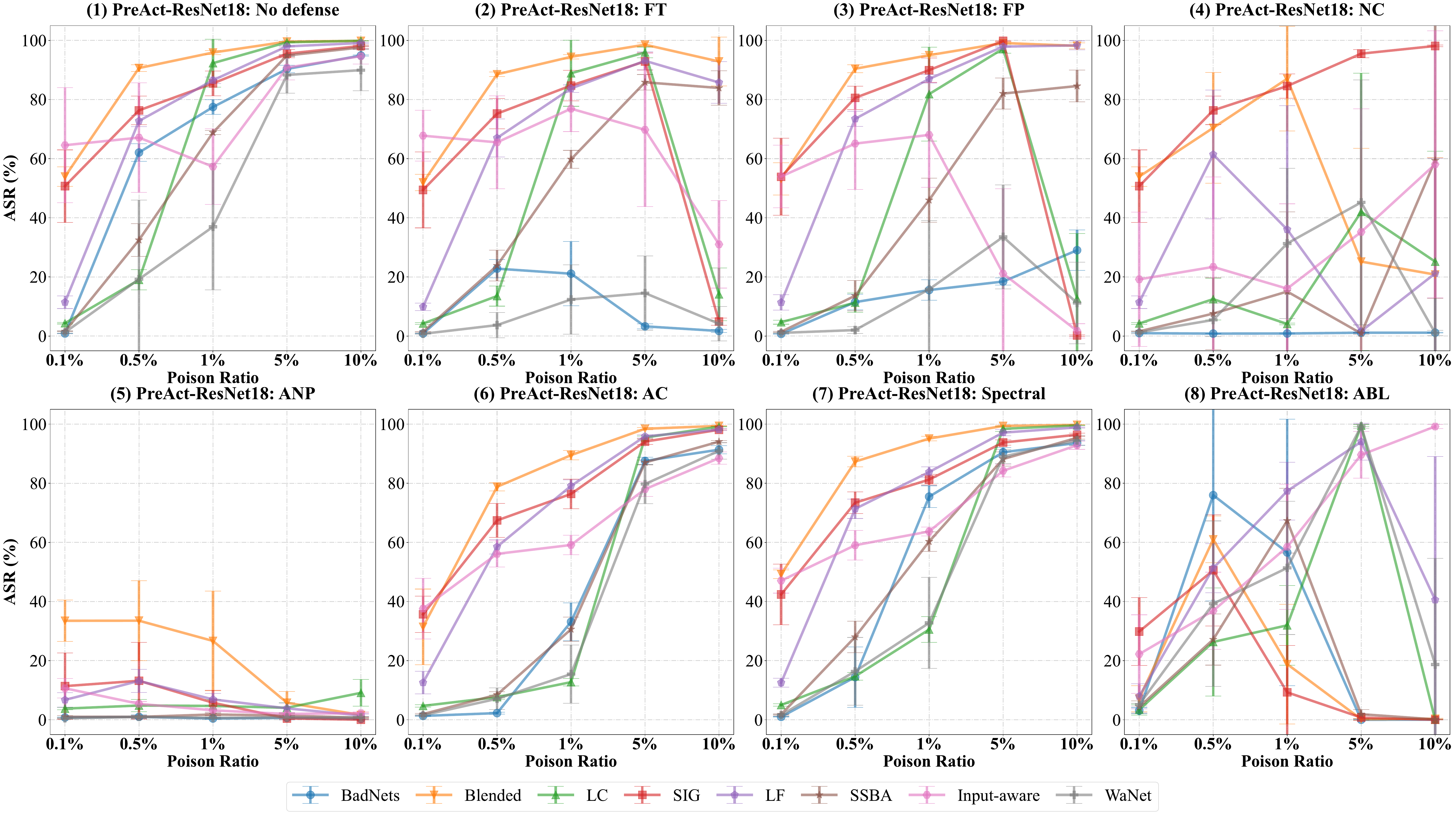}
    \caption{\blue{The effects of different poisoning ratios of backdoor learning with 5 random seeds.}}
    \label{stability-of-poison-ratio}
\end{figure}

\blue{
\paragraph{Analysis}
As shown in Figure \ref{stability-of-poison-ratio}, the trends of ASR curves are almost consistent with those in Figure 3 in the main manuscript, and the standard deviation (\ie, the error bar) is small, indicating that the abnormal phenomenon about the poisoning ratio's effect is not affected by the randomness. 
However, there are still a few special cases. 
For example, the error bars of some attacks under the ABL defense is very large when the poisoning ratio is low. The reason is that ABL identifies the fixed 1\% of all training samples as the poisoning samples according to the training loss. However, we find that the poisoning identification accuracy is very unstable, especially when the poisoning ratio is low, leading to the large fluctuation.   
The standard deviations of evaluations under the NC defense are also large. 
As described in Section \ref{appendix: subsec descriptions of  backdoor defense algorithms}, NC consists of two consecutive steps, \ie, firstly searching a candidate trigger to determine whether it is a backdoored model or not, then mitigating the backdoor effect through pruning.
We observe that the first step is very unstable within 5 random evaluations. If the backdoored model is successfully detected, then the ASR will be reduced significantly, other keeping the high value, causing the high standard deviations of 5 random evaluations. 
}

\subsubsection{Effect of poisoning ratio of other model architectures}
\label{sec: subsec Effect of poisoning ratio of other model architectures}

In this part, we intend to give the more detailed information about the variation of ASR values against poisoning ratio on different model structures, which are VGG-19, DenseNet-161, EfficientNet-B3, and MobileNetV3-Large, respectively. The corresponding results are established in Figure ~\ref{curves}. We have analyzed the effect of poisoning ratio in Section 4.3 in the main manuscript based on the results of Preact-ResNet18 on CIFAR-10. We found that the most ASR curves increase with the increase of poisoning ratio, while there are some curves which increase at first and then collapse dramatically. However, this phenomenon still exists for multiple model structures. It is interesting to notice that if the model structure is changed, the tendency of curves is different. The curve of NAD against BadNets can serve as an example. It keeps increasing in DenseNet-161, while increases at first and then drops down in VGG-19 and MobileNetV3-Large. Thus, it is valuable to further explore the relationship between model architecture and backdoor performance. 
Note that we don't provide the results of ANP and DBD on VGG-19, as we adopt the VGG-19 architecture without the batch normalization (BN) layer (see the demonstration in Section 4.1 of the main manuscript). According to the ANP author's comments at \href{https://github.com/csdongxian/ANP_backdoor/issues/2}{https://github.com/csdongxian/ANP\_backdoor/issues/2}, ANP is not suitable to the model architecture without the BN layer. 
Besides, in our evaluations, the defense performance of DBD is not very stable on the VGG-19 without the BN layer, at its semi-supervised learning phase. Thus, we also don't report the evaluation of DBD on VGG-19. However, we also observe that DBD performs stably on the VGG-19 model with the BN layer. The behind reason will be explored in future.


\begin{figure}[!ht]
    \centering
    \subfigure[The variation of ASR on different poisoning ratios with VGG-19 and CIFAR-10.]{
    \begin{minipage}[b]{1\textwidth}
    \includegraphics[width=\textwidth]{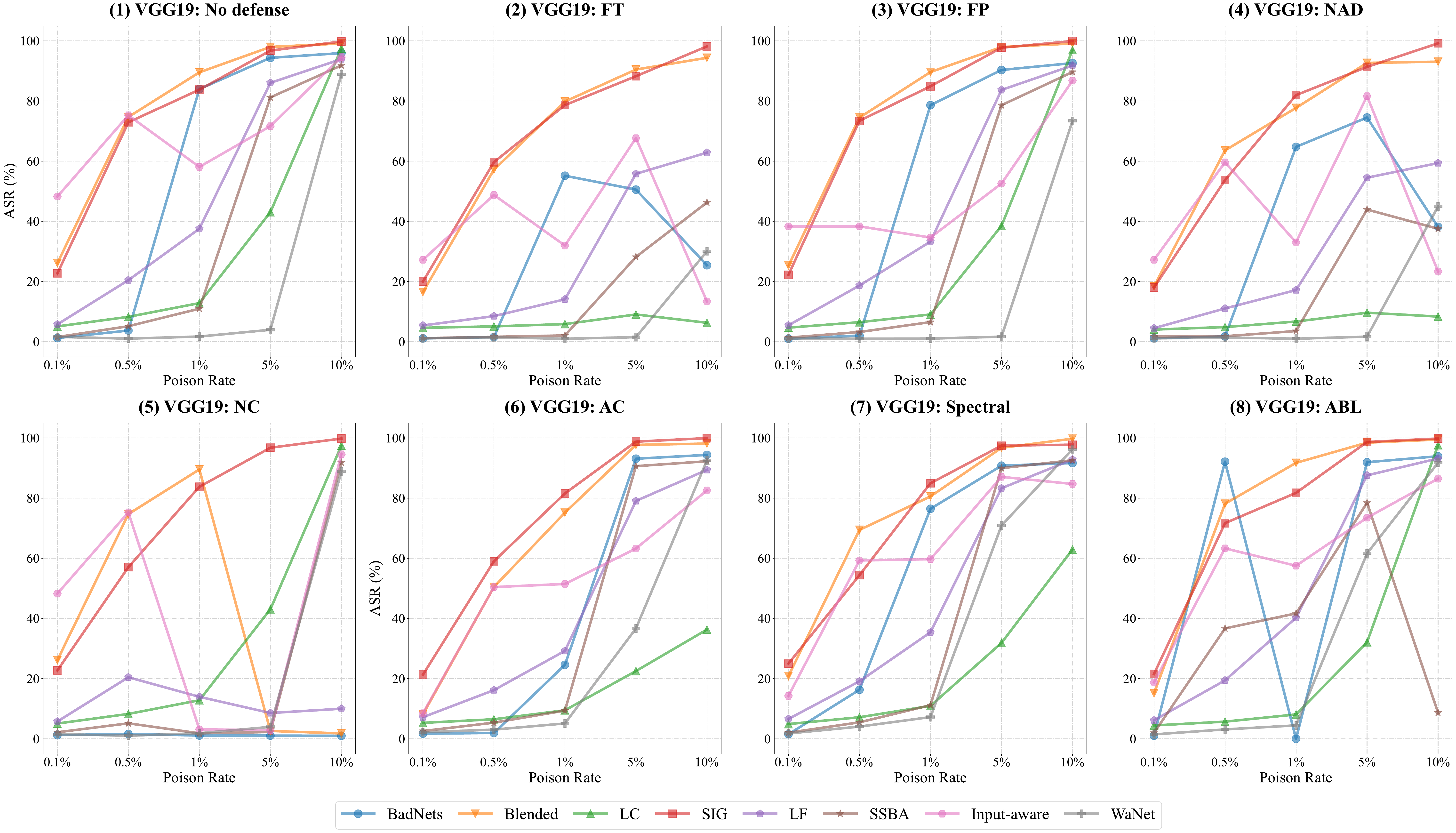}
    \end{minipage}
    }
    \subfigure[The variation of ASR on different poisoning ratios with DenseNet-161 and CIFAR-10.]{
    \begin{minipage}[b]{1\textwidth}
    \includegraphics[width=\textwidth]{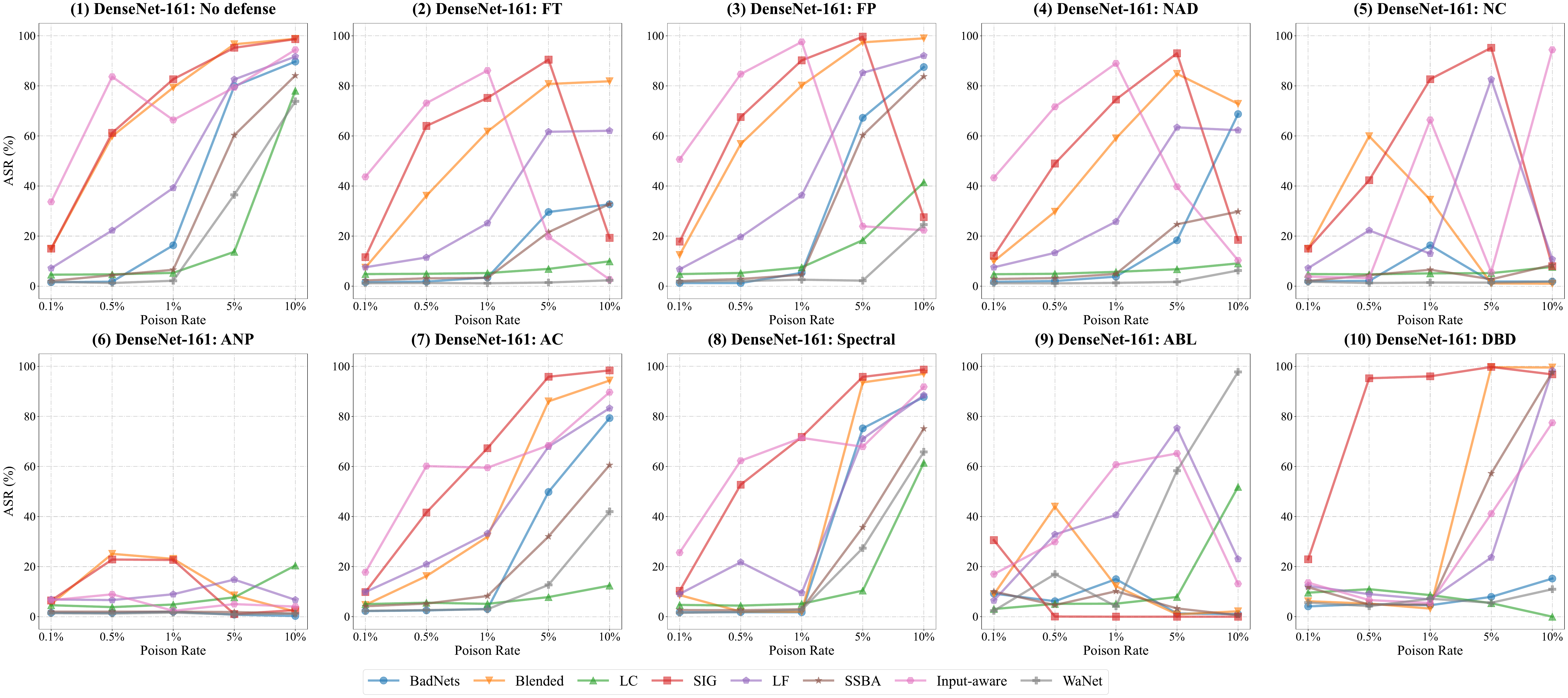}
    \end{minipage}
    }
    \subfigure[The variation of ASR on different poisoning ratios with EfficientNet-B3 and CIFAR-10.]{
    \begin{minipage}[b]{1\textwidth}
    \includegraphics[width=\textwidth]{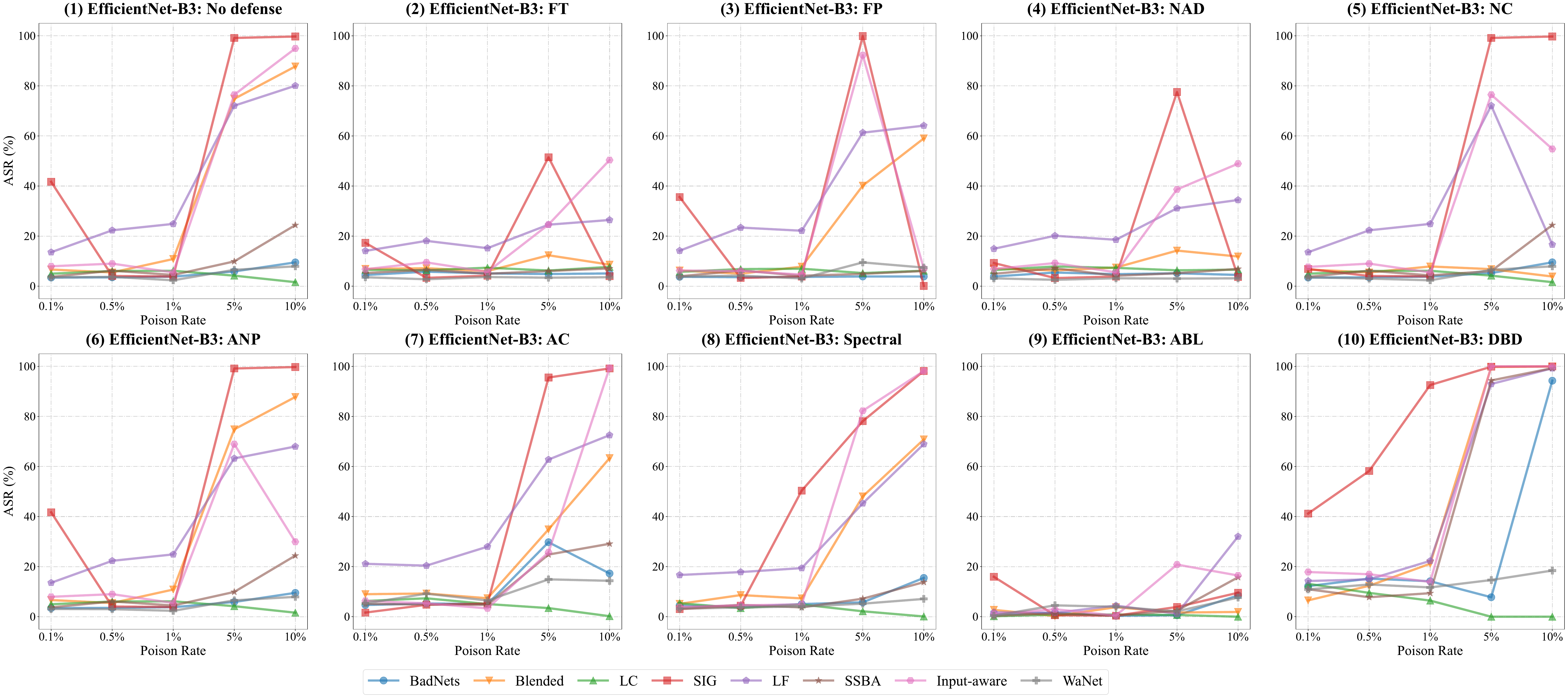}
    \end{minipage}
    }
\end{figure}
\begin{figure}[!ht]
    \centering
    \subfigure[The variation of ASR on different poisoning ratios with MobileNetV3-Large and CIFAR-10.]{
    \begin{minipage}[b]{1\textwidth}
    \includegraphics[width=\textwidth]{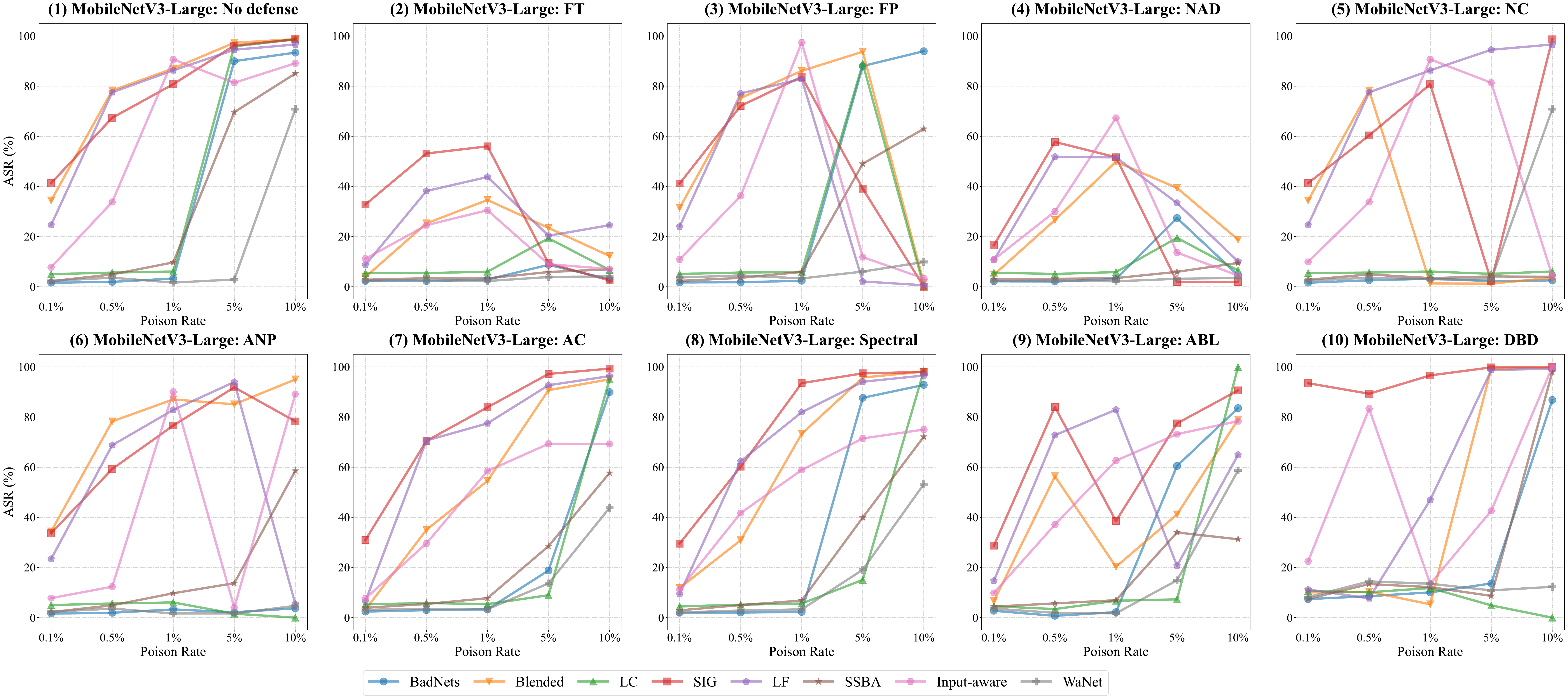}
    \end{minipage}
    }
    \caption{The effect of different ratios on backdoor learning. From (a) to (d), the structure of models are different. In the condition of no defense, the higher poisoning ratio, the higher ASR value. In the defense situation, some ASR curves raise with the increase of poisoning ratio, while some curves go up first and then sharply drop down. It could also be noticed that the performances of same defense method on different model structures are distinctive, \ie, ABL on VGG-19 and DenseNet-161. 
    Note that we don't provide the results of ANP and DBD on VGG-19, and the reason is illustrated in Section \ref{sec: subsec Effect of poisoning ratio of other model architectures}.
    }
    \label{curves}
\end{figure}
\subsection{\blue{Sensitivity to hyper-parameters}}
\label{appendix: subsec sensitivity test}

\blue{
As illustrated in Section \ref{appendix: subsec implementation details and computational complexities}, we adopt a fixed setting for each attack/defense method to ensure reproducibility and fair comparison. However, the sensitivity to hyper-parameters is also a very critical metric of one method's performance and practical usage. 
In the following, we pick three attack methods (\ie, BadNets, SIG, InputAware), two defense methods (\ie, ABL and ANP), two datasets (\ie, CIFAR-10 and GTSRB), and two models (\ie, PreAct-ResNet18 and VGG-19), to present a partial analysis about the sensitivity to hyper-parameters. For each attack/defense method, we study one key hyper-parameter, such as the trigger's patch size for BadNets, the trigger's frequency for SIG, the mask density for Input-aware, the flooding value for ABL and the poisoning threshold for ANP. 
}

\blue{
Results are shown in Tables \ref{tab:sens_cifar10_preact}, \ref{tab:sens_cifar10_vgg19}, \ref{tab:sens_gtsrb_preact}, and \ref{tab:sens_gtsrb_vgg19}. 
For BadNets, a larger square pattern means a more vigorous attack but is also easier to find by defense methods. For SIG, higher frequency means stronger attacks and harder to defend. For Input-aware, the ASR values fluctuate a lot \wrt the mask density in the case of no defense but are relatively stable under defenses.  
For ABL, a higher flooding parameter often leads to worse defense performance for most attacks. For ANP, the situation is complicated. When defending BadNets and SIG, the higher threshold often leads to better defense performance, but the defense performance against Input-aware is rather stable \wrt the threshold. 
When comparing the performance across different model architectures and different datasets, we find that the sensitivities to hyper-parameters of each method are very diverse. Picking a good hyper-parameter for a backdoor learning method in practice is a challenge.
}

\begin{sidewaystable}[!ht]
\Huge
\caption{Sensitivity results on CIFAR10 with PreAct-ResNet18.}
\label{tab:sens_cifar10_preact}
\renewcommand\arraystretch{2}
\resizebox{\textwidth}{!}
{
\centering
\begin{tabular}{@{}ll|lll|lll|lll|lll|lll|lll|lll|lll|lll|lll|lll@{}}
\toprule
           &       & No Defense   & No Defense   & No Defense   & ABL        & ABL      & ABL        & ABL        & ABL      & ABL        & ABL        & ABL      & ABL        & ABL        & ABL      & ABL        & ABL        & ABL      & ABL        & ANP        & ANP      & ANP        & ANP        & ANP      & ANP        & ANP        & ANP      & ANP        & ANP        & ANP      & ANP        & ANP        & ANP      & ANP        \\
           &       & $\backslash$ & $\backslash$ & $\backslash$ & 0.1        & 0.1      & 0.1        & 0.5        & 0.5      & 0.5        & 0.9        & 0.9      & 0.9        & 1.3        & 1.3      & 1.3        & 1.7        & 1.7      & 1.7        & 0.2        & 0.2      & 0.2        & 0.3        & 0.3      & 0.3        & 0.4        & 0.4      & 0.4        & 0.5        & 0.5      & 0.5        & 0.6        & 0.6      & 0.6        \\
           &       & C-Acc (\%)   & ASR (\%)     & R-Acc (\%)   & C-Acc (\%) & ASR (\%) & R-Acc (\%) & C-Acc (\%) & ASR (\%) & R-Acc (\%) & C-Acc (\%) & ASR (\%) & R-Acc (\%) & C-Acc (\%) & ASR (\%) & R-Acc (\%) & C-Acc (\%) & ASR (\%) & R-Acc (\%) & C-Acc (\%) & ASR (\%) & R-Acc (\%) & C-Acc (\%) & ASR (\%) & R-Acc (\%) & C-Acc (\%) & ASR (\%) & R-Acc (\%) & C-Acc (\%) & ASR (\%) & R-Acc (\%) & C-Acc (\%) & ASR (\%) & R-Acc (\%) \\ \midrule
BadNets     & 1     & 88.81        & 64.61        & 33.90        & 83.50      & 0.01     & 90.12      & 82.09      & 0.12     & 87.72      & 83.66      & 0.03     & 89.82      & 10.00      & 100.00   & 0.00       & 8.51       & 99.74    & 0.07       & 76.44      & 38.08    & 58.50      & 82.39      & 9.14     & 78.18      & 81.49      & 2.17     & 80.29      & 69.58      & 1.20     & 70.16      & 57.70      & 2.09     & 58.03      \\
BadNets     & 2     & 90.20        & 90.23        & 9.32         & 83.85      & 0.00     & 90.82      & 83.66      & 0.01     & 90.09      & 86.10      & 0.03     & 90.72      & 83.23      & 0.00     & 89.42      & 10.00      & 100.00   & 0.00       & 82.74      & 17.92    & 72.26      & 83.50      & 4.41     & 80.21      & 80.89      & 0.43     & 80.99      & 78.04      & 0.29     & 81.40      & 67.91      & 0.04     & 74.24      \\
BadNets     & 3     & 91.32        & 95.03        & 4.67         & 85.45      & 0.01     & 90.43      & 82.96      & 0.00     & 88.78      & 82.69      & 0.00     & 90.06      & 84.10      & 0.01     & 89.82      & 84.20      & 0.01     & 90.58      & 86.93      & 8.09     & 83.27      & 87.46      & 2.87     & 85.68      & 85.37      & 1.20     & 84.69      & 80.17      & 0.43     & 81.80      & 70.27      & 0.07     & 74.50      \\
BadNets     & 4     & 92.12        & 96.94        & 2.84         & 84.96      & 0.00     & 90.73      & 84.89      & 0.00     & 90.67      & 83.93      & 0.01     & 90.04      & 82.58      & 0.00     & 90.37      & 84.01      & 0.00     & 90.80      & 88.45      & 4.97     & 86.09      & 88.05      & 1.17     & 88.08      & 83.35      & 0.18     & 86.18      & 75.67      & 0.11     & 79.26      & 69.23      & 0.09     & 72.72      \\
BadNets     & 5     & 93.24        & 97.87        & 2.03         & 80.35      & 0.03     & 87.86      & 79.11      & 0.00     & 86.49      & 81.73      & 0.00     & 89.67      & 81.73      & 0.00     & 90.30      & 80.71      & 0.00     & 89.62      & 90.47      & 6.29     & 83.17      & 88.12      & 1.34     & 84.39      & 83.92      & 0.21     & 82.67      & 82.23      & 0.21     & 82.79      & 84.23      & 0.17     & 86.81      \\ \midrule
SIG        & 1     & 84.75        & 48.31        & 49.69        & 67.84      & 0.00     & 69.82      & 56.99      & 0.00     & 58.98      & 62.08      & 0.00     & 68.39      & 47.52      & 0.00     & 49.50      & 52.05      & 0.00     & 57.00      & 84.97      & 10.97    & 80.12      & 86.49      & 11.62    & 80.44      & 84.60      & 5.80     & 85.11      & 83.24      & 3.53     & 86.01      & 80.79      & 1.41     & 84.93      \\
SIG        & 2     & 84.65        & 83.22        & 16.41        & 74.41      & 0.00     & 65.97      & 68.86      & 0.00     & 69.53      & 67.62      & 0.00     & 72.92      & 71.82      & 0.00     & 72.42      & 64.90      & 0.00     & 67.59      & 84.05      & 19.07    & 65.84      & 84.94      & 17.30    & 70.73      & 83.23      & 13.64    & 70.00      & 80.06      & 6.16     & 74.82      & 76.59      & 4.63     & 70.16      \\
SIG        & 3     & 84.55        & 90.11        & 9.73         & 77.02      & 0.00     & 61.33      & 69.57      & 0.00     & 67.66      & 74.04      & 0.00     & 74.47      & 72.98      & 0.00     & 71.80      & 67.89      & 0.00     & 67.14      & 84.70      & 25.48    & 59.76      & 83.10      & 12.87    & 62.52      & 81.74      & 8.00     & 62.59      & 80.23      & 4.02     & 62.81      & 78.49      & 1.12     & 71.80      \\
SIG        & 4     & 84.47        & 97.16        & 2.82         & 74.14      & 0.00     & 51.79      & 57.33      & 0.00     & 62.82      & 68.04      & 0.00     & 63.59      & 65.23      & 0.00     & 60.92      & 53.67      & 0.00     & 64.31      & 83.81      & 9.59     & 65.92      & 82.31      & 6.98     & 67.88      & 82.87      & 7.59     & 66.82      & 81.78      & 3.21     & 65.49      & 80.45      & 1.01     & 64.67      \\
SIG        & 5     & 84.52        & 97.18        & 2.80         & 73.70      & 0.00     & 65.23      & 56.90      & 0.00     & 61.12      & 59.87      & 0.00     & 55.80      & 41.41      & 0.00     & 49.43      & 54.33      & 0.00     & 52.24      & 82.37      & 1.73     & 59.69      & 81.77      & 1.48     & 56.46      & 81.36      & 1.98     & 58.32      & 80.00      & 1.99     & 62.07      & 77.82      & 0.82     & 57.62      \\ \midrule
Input-aware & 0.032 & 90.94        & 89.22        & 10.11        & 61.30      & 95.94    & 3.12       & 78.60      & 93.34    & 5.36       & 48.16      & 100.00   & 0.00       & 47.99      & 98.61    & 0.97       & 49.40      & 25.06    & 34.81      & 90.59      & 2.40     & 86.34      & 90.27      & 3.93     & 84.11      & 89.64      & 5.17     & 80.08      & 88.43      & 7.22     & 77.42      & 85.79      & 8.61     & 75.84      \\
Input-aware & 0.048 & 90.98        & 91.81        & 7.70         & 80.78      & 90.71    & 7.56       & 34.29      & 99.33    & 0.28       & 16.89      & 99.96    & 0.04       & 42.13      & 94.16    & 3.23       & 52.34      & 11.12    & 47.86      & 91.33      & 3.03     & 84.57      & 91.13      & 2.78     & 84.76      & 90.72      & 3.77     & 84.42      & 89.25      & 5.62     & 78.68      & 86.97      & 7.07     & 74.92      \\
Input-aware & 0.064 & 89.94        & 92.30        & 7.29         & 79.11      & 98.81    & 1.01       & 67.28      & 99.84    & 0.10       & 39.14      & 99.98    & 0.01       & 55.76      & 97.41    & 1.73       & 60.86      & 15.43    & 55.21      & 90.26      & 1.48     & 86.79      & 89.83      & 1.44     & 85.66      & 89.86      & 1.87     & 86.02      & 87.99      & 3.97     & 83.12      & 82.33      & 4.22     & 77.69      \\
Input-aware & 0.08  & 90.69        & 95.97        & 3.91         & 74.14      & 5.90     & 62.78      & 64.92      & 98.67    & 0.97       & 33.29      & 99.97    & 0.01       & 31.56      & 99.38    & 0.29       & 80.89      & 0.00     & 84.03      & 90.26      & 1.00     & 87.07      & 89.64      & 1.51     & 85.48      & 90.34      & 1.98     & 85.22      & 88.98      & 2.68     & 82.12      & 87.66      & 4.08     & 78.22      \\
Input-aware & 0.096 & 90.61        & 95.58        & 4.17         & 87.96      & 0.27     & 82.32      & 34.55      & 99.83    & 0.02       & 10.19      & 100.00   & 0.00       & 15.27      & 99.88    & 0.00       & 82.02      & 0.00     & 87.09      & 89.95      & 1.56     & 86.84      & 90.20      & 1.28     & 86.92      & 89.96      & 2.46     & 85.90      & 89.29      & 2.84     & 85.02      & 87.62      & 2.29     & 83.34      \\ \bottomrule
\end{tabular}
}
\end{sidewaystable}

\begin{sidewaystable}[!ht]
\Huge
\caption{Sensitivity results on CIFAR10 with VGG-19.}
\label{tab:sens_cifar10_vgg19}
\renewcommand\arraystretch{2}
\resizebox{\textwidth}{!}
{

    \centering
\begin{tabular}{ll|lll|lll|lll|lll|lll|lll}
\hline
     &  & No Defense   & No Defense   & No Defense   & ABL        & ABL      & ABL        & ABL        & ABL      & ABL        & ABL        & ABL      & ABL        & ABL        & ABL      & ABL        & ABL        & ABL      & ABL        \\
     &  & $\backslash$ & $\backslash$ & $\backslash$ & 0.1        & 0.1      & 0.1        & 0.5        & 0.5      & 0.5        & 0.9        & 0.9      & 0.9        & 1.3        & 1.3      & 1.3        & 1.7        & 1.7      & 1.7        \\
     &  & C-Acc (\%)   & ASR (\%)     & R-Acc (\%)   & C-Acc (\%) & ASR (\%) & R-Acc (\%) & C-Acc (\%) & ASR (\%) & R-Acc (\%) & C-Acc (\%) & ASR (\%) & R-Acc (\%) & C-Acc (\%) & ASR (\%) & R-Acc (\%) & C-Acc (\%) & ASR (\%) & R-Acc (\%) \\ \hline
BadNets     & 1      & 85.79        & 76.44        & 22.46        & 84.14      & 70.67    & 26.70      & 71.31      & 45.73    & 43.34      & 53.99      & 45.03    & 36.38      & 10.00      & 100.00   & 0.00       & 10.00      & 100.00   & 0.00       \\
BadNets     & 2      & 87.26        & 93.76        & 5.80         & 84.55      & 73.70    & 23.83      & 85.18      & 89.32    & 9.62       & 83.58      & 88.21    & 10.67      & 78.82      & 85.81    & 12.76      & 10.00      & 100.00   & 0.00       \\
BadNets     & 3      & 88.75        & 95.77        & 3.93         & 85.93      & 91.29    & 7.87       & 84.67      & 90.51    & 8.56       & 84.31      & 93.03    & 6.28       & 83.75      & 92.74    & 6.54       & 79.32      & 91.17    & 7.73       \\
BadNets     & 4      & 89.50        & 96.73        & 3.08         & 86.41      & 93.91    & 5.52       & 86.77      & 92.56    & 6.80       & 87.24      & 94.44    & 5.02       & 85.75      & 95.06    & 4.50       & 85.69      & 94.64    & 4.90       \\
BadNets     & 5      & 89.87        & 97.03        & 2.78         & 88.04      & 96.18    & 3.53       & 88.67      & 96.74    & 3.02       & 87.79      & 96.98    & 2.76       & 88.22      & 97.04    & 2.73       & 88.28      & 97.50    & 2.27       \\ \hline
SIG        & 1      & 82.20        & 41.61        & 53.47        & 80.54      & 41.06    & 51.78      & 80.55      & 44.52    & 49.36      & 80.68      & 44.10    & 50.00      & 80.33      & 44.80    & 49.26      & 80.14      & 45.06    & 47.74      \\
SIG        & 2      & 81.92        & 74.87        & 21.84        & 80.16      & 72.33    & 23.38      & 80.16      & 76.11    & 20.68      & 80.39      & 78.21    & 19.44      & 80.02      & 82.79    & 14.37      & 80.05      & 83.04    & 14.23      \\
SIG        & 3      & 81.44        & 88.27        & 8.38         & 80.25      & 87.62    & 10.23      & 80.79      & 89.33    & 9.21       & 80.14      & 87.14    & 10.50      & 80.46      & 85.22    & 11.08      & 80.40      & 90.36    & 7.99       \\
SIG        & 4      & 81.44        & 98.03        & 1.42         & 80.73      & 98.43    & 1.32       & 80.51      & 98.39    & 1.40       & 80.45      & 99.27    & 0.61       & 80.55      & 98.76    & 1.01       & 78.89      & 94.52    & 4.13       \\
SIG        & 5      & 82.16        & 99.50        & 0.30         & 80.04      & 99.34    & 0.42       & 80.89      & 99.19    & 0.63       & 80.17      & 99.17    & 0.72       & 80.58      & 98.56    & 1.08       & 80.01      & 98.41    & 1.24       \\ \hline
Input-aware & 0.032  & 88.48        & 96.77        & 2.89         & 88.27      & 90.49    & 8.19       & 88.18      & 92.08    & 6.90       & 24.65      & 100.00   & 0.00       & 86.56      & 7.87     & 73.64      & 10.00      & 100.00   & 0.00       \\
Input-aware & 0.048  & 88.86        & 94.98        & 4.44         & 88.31      & 81.49    & 16.21      & 88.43      & 79.66    & 17.93      & 88.78      & 72.57    & 24.50      & 10.00      & 100.00   & 0.00       & 10.00      & 100.00   & 0.00       \\
Input-aware & 0.064  & 88.71        & 85.72        & 12.91        & 88.58      & 84.52    & 13.64      & 76.06      & 98.67    & 1.03       & 10.00      & 100.00   & 0.00       & 50.51      & 0.00     & 37.17      & 10.00      & 100.00   & 0.00       \\
Input-aware & 0.08   & 88.54        & 97.00        & 2.61         & 87.81      & 84.16    & 13.40      & 88.22      & 86.89    & 11.50      & 88.04      & 88.20    & 10.16      & 87.66      & 87.62    & 10.60      & 10.00      & 100.00   & 0.00       \\
Input-aware & 0.096  & 89.11        & 96.52        & 3.36         & 88.01      & 81.91    & 15.76      & 88.74      & 83.54    & 14.17      & 88.38      & 84.06    & 14.08      & 87.39      & 6.69     & 73.10      & 10.00      & 100.00   & 0.00       \\ \hline
\end{tabular}

}
\end{sidewaystable}
\begin{sidewaystable}[!ht]
\Huge
\caption{Sensitivity results on GTSRB with PreAct-ResNet18.}
\label{tab:sens_gtsrb_preact}
\renewcommand\arraystretch{2}
\resizebox{\textwidth}{!}
{

    \centering
    \begin{tabular}{ll|lll|lll|lll|lll|lll|lll|lll|lll|lll|lll|lll}
\hline
     &  & No Defense   & No Defense   & No Defense   & ABL        & ABL      & ABL        & ABL        & ABL      & ABL        & ABL        & ABL      & ABL        & ABL        & ABL      & ABL        & ABL        & ABL      & ABL        & ANP        & ANP      & ANP        & ANP        & ANP      & ANP        & ANP        & ANP      & ANP        & ANP        & ANP      & ANP        & ANP   & ANP  & ANP   \\
     &  & $\backslash$ & $\backslash$ & $\backslash$ & 0.1        & 0.1      & 0.1        & 0.5        & 0.5      & 0.5        & 0.9        & 0.9      & 0.9        & 1.3        & 1.3      & 1.3        & 1.7        & 1.7      & 1.7        & 0.2        & 0.2      & 0.2        & 0.3        & 0.3      & 0.3        & 0.4        & 0.4      & 0.4        & 0.5        & 0.5      & 0.5        & 0.6   & 0.6  & 0.6   \\
     &  & C-Acc (\%)   & ASR (\%)     & R-Acc (\%)   & C-Acc (\%) & ASR (\%) & R-Acc (\%) & C-Acc (\%) & ASR (\%) & R-Acc (\%) & C-Acc (\%) & ASR (\%) & R-Acc (\%) & C-Acc (\%) & ASR (\%) & R-Acc (\%) & C-Acc (\%) & ASR (\%) & R-Acc (\%) & C-Acc (\%) & ASR (\%) & R-Acc (\%) & C-Acc (\%) & ASR (\%) & R-Acc (\%) & C-Acc (\%) & ASR (\%) & R-Acc (\%) & C-Acc (\%) & ASR (\%) & R-Acc (\%) & ACC   & ASR  & RA    \\ \hline
BadNets     & 1      & 95.97        & 92.25        & 7.48         & 90.19      & 0.00     & 91.02      & 85.45      & 0.01     & 85.45      & 0.48       & 100.00   & 0.00       & 0.48       & 100.00   & 0.00       & 0.48       & 100.00   & 0.00       & 95.00      & 9.75     & 89.14      & 94.19      & 3.51     & 93.16      & 93.38      & 3.70     & 92.39      & 91.23      & 2.35     & 90.80      & 88.51 & 0.70 & 88.70 \\
BadNets     & 2      & 96.83        & 93.99        & 5.98         & 93.60      & 0.00     & 94.15      & 93.21      & 0.00     & 93.72      & 93.93      & 0.00     & 94.53      & 91.68      & 0.00     & 92.32      & 91.65      & 0.00     & 91.93      & 95.10      & 0.07     & 95.19      & 96.22      & 0.00     & 96.23      & 96.09      & 0.01     & 96.17      & 93.28      & 0.00     & 93.37      & 89.28 & 0.00 & 89.69 \\
BadNets     & 3      & 97.56        & 94.84        & 5.10         & 82.26      & 0.00     & 82.29      & 93.82      & 0.00     & 94.21      & 93.49      & 0.00     & 93.77      & 94.37      & 0.00     & 94.91      & 71.26      & 0.00     & 71.92      & 97.59      & 16.49    & 82.91      & 96.66      & 0.12     & 96.46      & 96.68      & 0.00     & 96.79      & 94.93      & 0.00     & 95.06      & 93.35 & 0.00 & 93.75 \\
BadNets     & 4      & 97.31        & 96.83        & 3.07         & 94.54      & 0.00     & 94.98      & 95.20      & 0.00     & 95.67      & 95.01      & 0.00     & 95.45      & 93.97      & 0.00     & 94.44      & 93.92      & 0.00     & 94.29      & 97.75      & 0.00     & 97.82      & 97.50      & 0.00     & 97.64      & 96.10      & 0.00     & 96.32      & 93.74      & 0.00     & 94.02      & 91.93 & 0.00 & 92.31 \\
BadNets     & 5      & 97.85        & 97.49        & 2.46         & 80.69      & 0.00     & 81.11      & 93.60      & 0.00     & 94.11      & 92.45      & 0.00     & 92.90      & 89.10      & 0.00     & 89.63      & 82.58      & 0.00     & 83.09      & 98.00      & 0.00     & 98.04      & 97.98      & 0.00     & 98.01      & 97.61      & 0.00     & 97.57      & 97.00      & 0.00     & 97.02      & 95.39 & 0.00 & 95.38 \\ \hline
Input-aware & 0.032  & 97.23        & 97.36        & 2.52         & 41.34      & 85.39    & 7.55       & 29.45      & 68.56    & 13.83      & 7.90       & 74.31    & 0.56       & 6.73       & 83.58    & 2.13       & 3.79       & 90.03    & 0.06       & 96.88      & 0.01     & 95.25      & 95.62      & 0.00     & 93.80      & 94.39      & 0.00     & 91.91      & 91.84      & 0.00     & 88.30      & 91.46 & 0.00 & 88.44 \\
Input-aware & 0.048  & 97.48        & 96.67        & 3.31         & 64.47      & 50.61    & 27.53      & 12.82      & 94.71    & 0.51       & 5.48       & 80.84    & 1.35       & 7.05       & 96.89    & 0.31       & 18.07      & 76.04    & 4.65       & 97.27      & 0.00     & 96.91      & 96.70      & 0.00     & 95.93      & 96.72      & 0.00     & 96.09      & 96.49      & 0.00     & 95.69      & 96.34 & 0.02 & 95.12 \\
Input-aware & 0.064  & 97.66        & 88.88        & 10.54        & 52.05      & 66.21    & 16.32      & 9.57       & 79.77    & 2.32       & 4.50       & 62.74    & 1.99       & 27.28      & 48.58    & 13.97      & 3.63       & 91.19    & 1.19       & 96.63      & 0.31     & 94.58      & 96.44      & 0.28     & 94.32      & 95.63      & 0.22     & 92.90      & 95.65      & 0.20     & 93.02      & 94.21 & 0.45 & 91.69 \\
Input-aware & 0.08   & 97.77        & 96.65        & 3.34         & 15.19      & 70.62    & 5.27       & 12.72      & 41.96    & 6.54       & 19.73      & 80.36    & 3.37       & 25.81      & 55.12    & 11.89      & 5.82       & 92.60    & 1.89       & 97.14      & 0.00     & 95.02      & 96.23      & 0.00     & 93.87      & 95.85      & 0.00     & 93.44      & 95.40      & 0.04     & 91.86      & 94.75 & 0.02 & 90.53 \\
Input-aware & 0.096  & 98.26        & 97.24        & 2.74         & 23.83      & 67.91    & 7.46       & 25.49      & 75.73    & 6.00       & 1.63       & 96.47    & 0.25       & 0.86       & 99.55    & 0.00       & 1.69       & 98.63    & 0.00       & 97.69      & 0.64     & 96.25      & 97.54      & 0.48     & 96.04      & 97.03      & 0.16     & 95.53      & 96.59      & 0.32     & 94.71      & 95.12 & 0.29 & 92.84 \\ \hline
\end{tabular}
}
\end{sidewaystable}

\begin{sidewaystable}[!ht]
\Huge
\caption{Sensitivity results on GTSRB with VGG-19.}
\label{tab:sens_gtsrb_vgg19}
\renewcommand\arraystretch{2}
\resizebox{\textwidth}{!}
{

    \centering
    \begin{tabular}{ll|lll|lll|lll|lll|lll|lll}
\hline
           &       & No Defense   & No Defense   & No Defense   & ABL   & ABL    & ABL   & ABL   & ABL    & ABL   & ABL   & ABL    & ABL   & ABL   & ABL    & ABL  & ABL   & ABL    & ABL  \\
           &       & $\backslash$ & $\backslash$ & $\backslash$ & 0.1   & 0.1    & 0.1   & 0.5   & 0.5    & 0.5   & 0.9   & 0.9    & 0.9   & 1.3   & 1.3    & 1.3  & 1.7   & 1.7    & 1.7  \\
           &       & ACC          & ASR          & RA           & ACC   & ASR    & RA    & ACC   & ASR    & RA    & ACC   & ASR    & RA    & ACC   & ASR    & RA   & ACC   & ASR    & RA   \\ \hline
BadNets     & 1     & 96.17        & 89.01        & 9.84         & 96.13 & 91.85  & 7.81  & 0.48  & 100.00 & 0.00  & 0.48  & 100.00 & 0.00  & 0.48  & 100.00 & 0.00 & 0.48  & 100.00 & 0.00 \\
BadNets     & 2     & 95.20        & 93.50        & 5.91         & 0.48  & 100.00 & 0.00  & 0.48  & 100.00 & 0.00  & 0.48  & 100.00 & 0.00  & 0.48  & 100.00 & 0.00 & 57.59 & 96.01  & 2.79 \\
BadNets     & 3     & 96.14        & 94.52        & 5.18         & 0.48  & 100.00 & 0.00  & 70.24 & 42.78  & 41.52 & 0.48  & 100.00 & 0.00  & 0.48  & 100.00 & 0.00 & 0.48  & 100.00 & 0.00 \\
BadNets     & 4     & 94.35        & 96.38        & 3.31         & 96.26 & 95.55  & 4.27  & 96.26 & 93.09  & 6.57  & 95.56 & 93.35  & 6.38  & 92.91 & 93.20  & 6.26 & 90.07 & 91.03  & 8.23 \\
BadNets     & 5     & 95.77        & 96.40        & 3.31         & 96.63 & 96.36  & 3.37  & 96.63 & 96.86  & 2.80  & 96.31 & 96.25  & 3.45  & 95.93 & 96.13  & 3.64 & 96.38 & 96.12  & 3.75 \\ \hline
Input-aware & 0.032 & 97.51        & 88.97        & 10.61        & 96.83 & 53.95  & 43.74 & 96.63 & 58.14  & 40.04 & 95.94 & 59.78  & 38.02 & 0.48  & 100.00 & 0.00 & 0.48  & 100.00 & 0.00 \\
Input-aware & 0.048 & 97.44        & 60.09        & 39.36        & 96.35 & 61.02  & 37.30 & 8.99  & 99.83  & 0.17  & 0.48  & 100.00 & 0.00  & 0.48  & 100.00 & 0.00 & 0.48  & 100.00 & 0.00 \\
Input-aware & 0.064 & 96.55        & 89.63        & 9.99         & 96.67 & 54.98  & 43.28 & 96.71 & 51.89  & 45.87 & 0.48  & 100.00 & 0.00  & 0.48  & 100.00 & 0.00 & 0.48  & 100.00 & 0.00 \\
Input-aware & 0.08  & 96.89        & 86.45        & 13.43        & 96.46 & 51.20  & 46.48 & 96.68 & 44.12  & 53.95 & 0.48  & 100.00 & 0.00  & 0.48  & 100.00 & 0.00 & 0.48  & 100.00 & 0.00 \\
Input-aware & 0.096 & 96.25        & 67.77        & 31.54        & 96.16 & 49.47  & 47.90 & 0.48  & 100.00 & 0.00  & 0.48  & 100.00 & 0.00  & 0.48  & 100.00 & 0.00 & 0.48  & 100.00 & 0.00 \\ \hline
\end{tabular}

}
\end{sidewaystable}


\subsection{\blue{Analysis of quick learning of backdoor}}
\label{appendix: sec quick learning}
\blue{
The quick learning phenomenon of backdoor has been observed in some previous works \cite{li2021anti}, \ie, the backdoor could be quickly learned in a few epochs, for almost all backdoor attacks. However, the behind reason has not been studied. In the following, we provide a detailed analysis from the perspective of gradient. Specifically, for each epoch during the training process, we record the following information: 
      \begin{itemize}
        \item Losses of training samples, clean testing samples, and poisoned testing samples;
        \item Accuracy on training samples, clean testing samples, and poisoned testing samples;
        \item Gradient signal to noise ratios (GSNR) \cite{liu2020understanding} on training samples, clean train samples, and poisoned training samples averaged over model parameters;
        \item Norms of average gradient on total training samples, clean training samples, and poisoned training samples;
        \item Pairwise cosine similarities between average gradients on total training samples, clean training samples, and poisoned training samples.
    \end{itemize}
    }  
    
\begin{figure}[!ht]
    \centering
    \includegraphics[width=\textwidth]{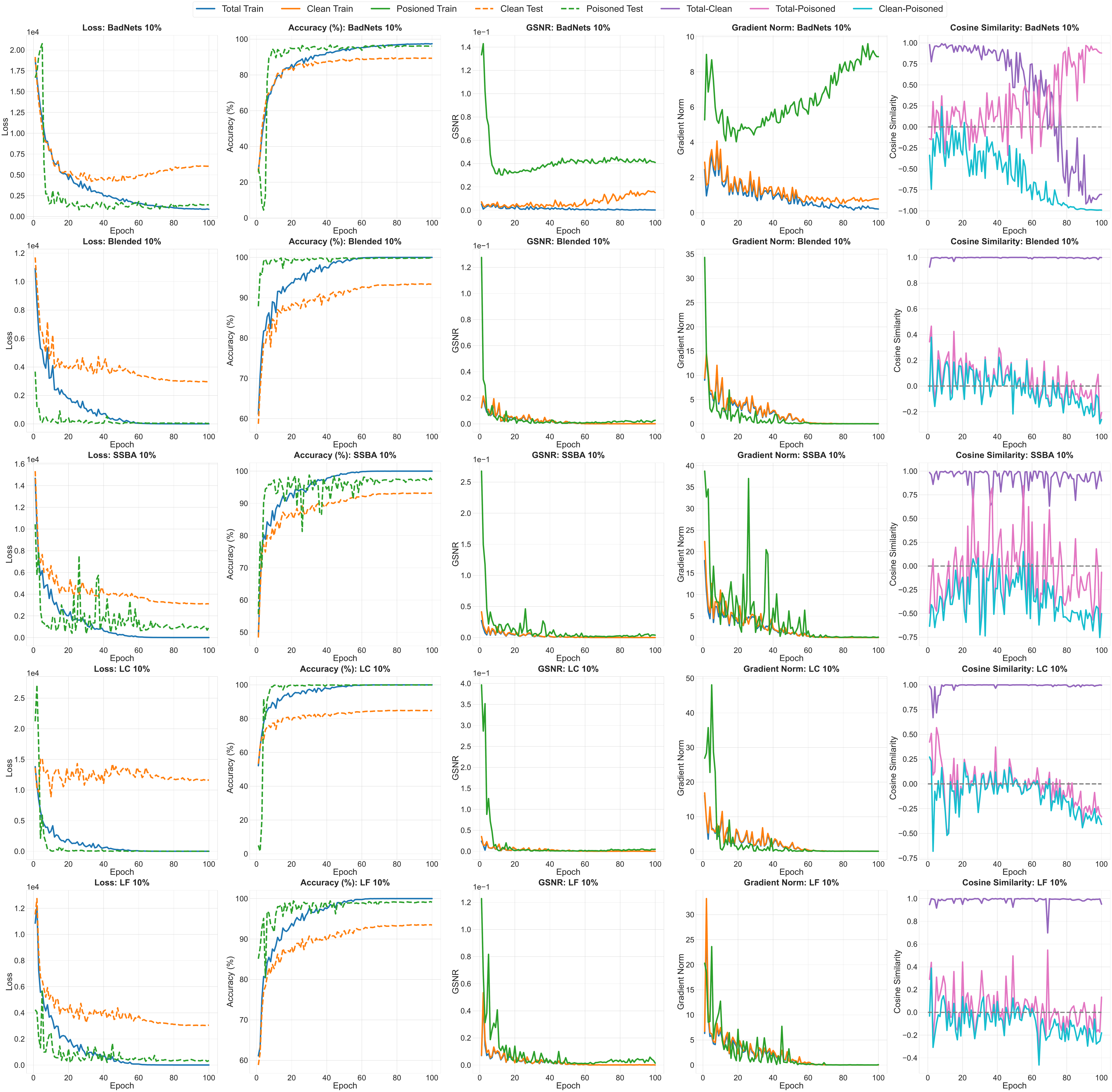}
    \caption{\blue{Analysis of the quick learning phenomenon based on the details of the backdoor learning process. From the first column to the last column, we report the curves of loss, accuracy, GSNR, gradient norm and cosine similarity of gradients in poisoned samples, clean samples and all samples, respectively.}}
    \label{grad_info}
\end{figure}

    \blue{
    As shown in Figure \ref{grad_info}, we report the results of 5 backdoor attacks, including BadNets, Blended, SSBA, LC and LF with poisoning ratio $10\%$, on the CIFAR-10 dataset and the PreAct-ResNet18 model.
    As shown in the first column, the testing loss of poisoned samples drops quickly in the early stages of training and converges to a low value, while the testing loss of clean samples drops at a slower rate and converges to a much larger value. It verifies the quick learning phenomenon of backdoor under these five backdoor attack methods. 
    As shown in the third column, we first observe that the GSNR of poisoned samples is significantly larger than the GSNR of clean samples at the early stages. The high GSNR values of poisoned samples indicate that the backdoor has better generalization performance and is consistent with the higher accuracy (ASR) and lower loss on poisoned testing samples. 
    Secondly, we notice that the norm of the gradient on poisoned samples is much larger than the norm of the gradient on clean samples in early epochs, as shown in the fourth column. Consequently, the cosine similarity between gradients on total training samples and poisoned training samples is significantly larger than the cosine similarity between gradients on clean training samples and poisoned training samples, though the number of poisoned samples is much smaller than the number of clean samples.
    }
    
\subsection{\blue{Analysis of backdoor forgetting}}
\label{appendix: subsec backdoor forgetting}

From the above analysis about the quick learning of backdoor in Section \ref{appendix: sec quick learning}, we get one impression that the backdoored model memorizes the poisoned samples quickly and stably. 
To obtain more insights about the inner mechanism of backdoor learning, we adopt the concept of \textit{forgetting event} \cite{forgetting} to characterize the learning dynamics during the training process. Specifically, one forgetting event is recorded when a correctly predicted training sample at the current epoch is incorrectly predicted at the next epoch. 
Formally, given a training sample $(\mathbf{x},y)$, where $\mathbf{x}$ is input feature, $y$ is ground-truth label. 
If $\mathbf{x}$ is correctly predicted at the epoch $t$, \textit{i.e.}, $f_{\theta^t}(\mathbf{x})=y$, but is misclassified at epoch $t+1$, \textit{i.e.}, $f_{\theta^{t+1}}(\mathbf{x})\neq y$, where $f$ denotes the model and $\theta^t,\theta^{t+1}$ are model parameters at the epoch $t$ and $t+1$, then a forgetting event is recorded for this sample.

\begin{figure}[!ht]
    \centering
    \includegraphics[width=0.9\textwidth]{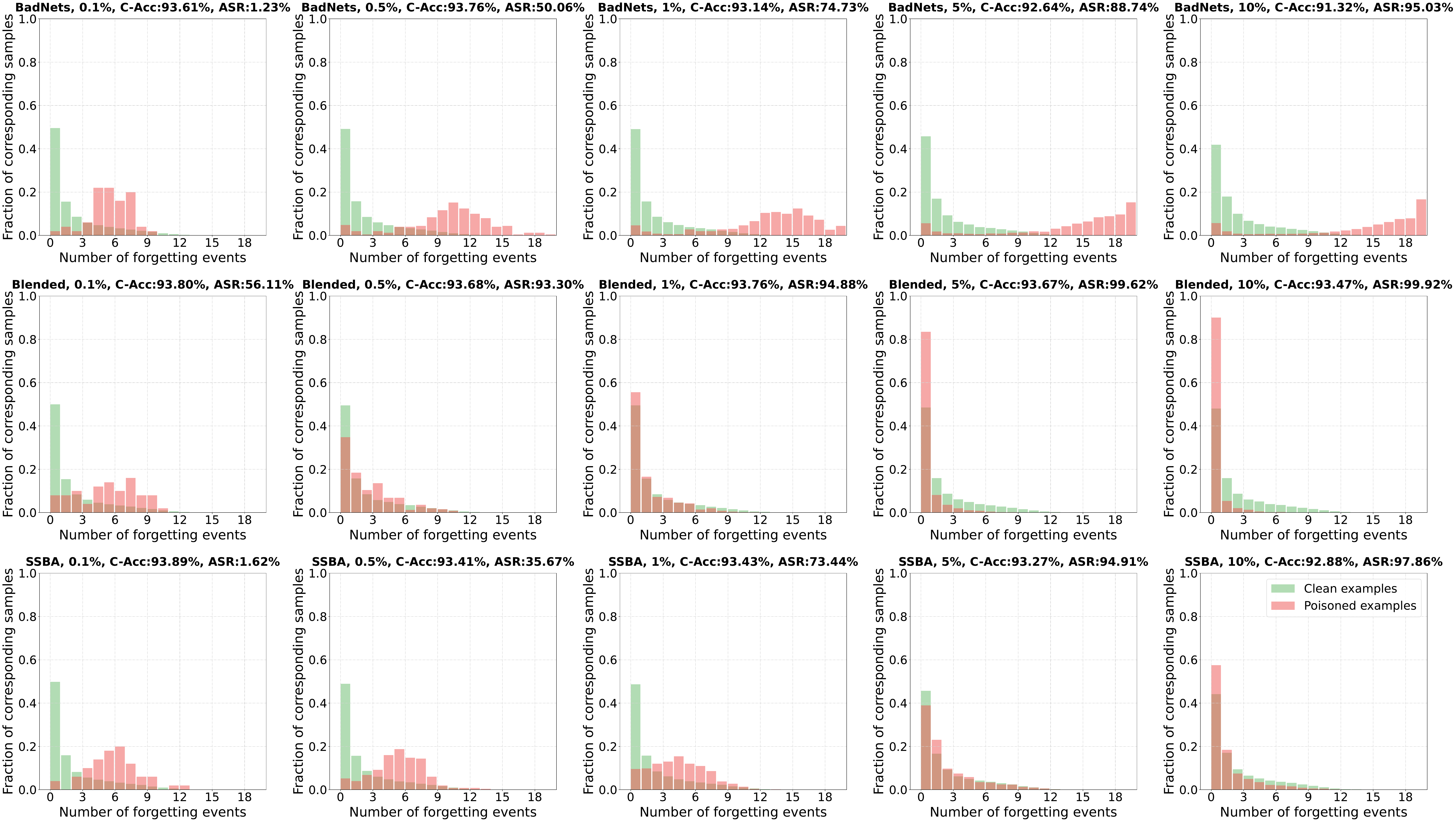}
    \caption{\blue{Distributions of forgetting events of clean training examples and poisoned examples in CIFAR-10 with Preact-Resnet18 backbone.}}
    \label{app:forget}
\end{figure}

Specifically, we count the number of forgetting events for clean and poisoned training samples, respectively, on CIFAR-10 with Preact-ResNet18 backbone. 
The distributions of forgetting events of clean and poisoned samples are shown in Figure \ref{app:forget}. 
The results show that: 
\begin{itemize}
\item  The forgetting events of clean training samples follow an exponential distribution, and are similar among different cases. 
\item For poisoned training samples: \textbf{1)} when the poisoning ratio is low (\eg, 0.1\%, 0.5\%), the forgetting numbers of poisoned samples are often larger than those of clean samples; \textbf{2)} when the poisoning ratio is high (\eg, 5\%, 10\%), the forgetting numbers of poisoned samples are often smaller than those of clean samples.
\end{itemize}
The above observations are compatible with our high-level observation that the backdoor attack with higher poisoning ratios could quickly learn the stable mapping from the poisoned samples to the target class. Moreover, the forgetting event provides a fine-grained tool to analyze the contribution of each individual training sample, which could facilitate the development of more advanced backdoor attack and defense methods.

\blue{\subsection{Analysis of trigger generalization of backdoor attacks}}
\label{appendix: subsec trigger generalization}

\blue{
In all existing backdoor attacks, there is a default assumption that the triggers used in both backdoor training and backdoor testing are exactly same. However, we find that an interesting property in backdoor learning that the backdoored model trained with one trigger could be also activated by other triggers. We name it as \textbf{trigger generalization}. 
}

\begin{figure}[!ht]
    \centering
    \includegraphics[width=\textwidth]{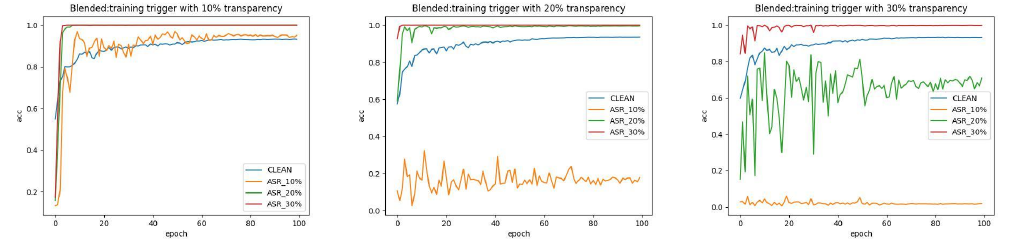}
    \caption{\blue{Analysis of trigger generalization in Blended attack: a) the training trigger with 10\% transparency; b) the training trigger with 20\% transparency; c) the training trigger with 30\% transparency. For each case, we evaluate the attack success rate of testing triggers with the transparency 10\%, 20\%, and 30\%, respectively.}}
    \label{blended generalization}
\end{figure}

\blue{
In this following, we take the Blended attack as an example to study the trigger generalization. Specifically, we set the trigger transparency to different values in training and testing phase, including 10\%, 20\%, and 30\%. 
As shown in the figure~\ref{blended generalization}, we obtain the following observations: (1) If a more obvious trigger (\ie, high transparency 30\%) is applied during training, then the backdoor will not be easily activated by a different trigger with lower transparency; (2) If using a less obvious backdoor triggers (\ie, low transparency 10\%) in the training phase, then the backdoor can be successfully activated by the trigger with higher transparency (\eg, 20\% or 30\%) in the test phase.
}

\blue{
In addition to the above example, we find that the trigger generalization is a common property of the backdoor models under several backdoor attacks. For example, the trigger in SSBA \cite{ssba} is a string, and its backdoored model could not only be activated by its training trigger. However, we find that the model is likely to be activated by many other strings with the same length. Exploring the behind reason of trigger generalization is important for us to better understand the backdoor mechanism. Moreover, we notice that there have been some attempts to utilize the backdoor as the technique to protect the intellectual property (IP) of AI models or datasets, based on the unique mapping from the trigger to the target class. However, due to the trigger generalization, the uniqueness of the training trigger no longer exists, which undermines the legitimacy backdoor learning in IP protection. Thus, the study of trigger generation is also important for the usage of backdoor learning in practice. 
We will provide more analysis about trigger generalization in BackdoorBench in future. 
}

\begin{table*}[!h]
\renewcommand{\arraystretch}{1.8}
\caption{\blue{Evaluation of ViT-b-16 on CIFAR-10 with $10\%$ poisoning ratio.}}
\label{tab:vit}
\resizebox{0.99\linewidth}{!}{
\begin{tabular}{@{}l|lll|lll|lll|lll@{}}
\toprule
            & \multicolumn{3}{c|}{No defense} & \multicolumn{3}{c|}{FT} & \multicolumn{3}{c|}{NC} & \multicolumn{3}{c}{ABL} \\ \midrule
Backdoor Attack &
  \multicolumn{1}{c}{C-Acc (\%)} &
  \multicolumn{1}{c}{ASR (\%)} &
  \multicolumn{1}{c|}{R-Acc (\%)} &
  \multicolumn{1}{c}{C-Acc (\%)} &
  \multicolumn{1}{c}{ASR (\%)} &
  \multicolumn{1}{c|}{R-Acc (\%)} &
  \multicolumn{1}{c}{C-Acc (\%)} &
  \multicolumn{1}{c}{ASR (\%)} &
  \multicolumn{1}{c|}{R-Acc (\%)} &
  \multicolumn{1}{c}{C-Acc (\%)} &
  \multicolumn{1}{c}{ASR (\%)} &
  \multicolumn{1}{c}{R-Acc (\%)} \\ \midrule
BadNets     & 94.58     & 94.11    & 5.66     & 42.00  & 8.81   & 38.60 & 36.09  & 11.12  & 33.60 & 10.00  & 0.00   & 11.11 \\
Blended     & 96.47     & 99.72    & 0.28     & 41.36  & 7.24   & 36.29 & 44.92  & 3.90   & 39.40 & 96.88  & 99.81  & 0.19  \\
LC          & 86.87     & 99.84    & 0.16     & 33.03  & 16.40  & 24.36 & 40.68  & 11.17  & 30.74 & 87.33  & 99.76  & 0.23  \\
SIG         & 87.01     & 92.60    & 7.28     & 44.46  & 0.78   & 25.92 & 87.01  & 92.60  & 7.28  & 87.60  & 86.47  & 13.39 \\
SSBA        & 96.30     & 97.58    & 2.34     & 45.35  & 7.62   & 43.49 & 46.24  & 8.17   & 43.60 & 96.78  & 98.22  & 1.70  \\
Input-aware & 92.10     & 96.28    & 3.51     & 89.56  & 25.14  & 69.91 & 43.13  & 7.79   & 37.81 & 96.70  & 93.21  & 6.58  \\ \hline

\end{tabular}}
\end{table*}

\blue{\subsection{Evaluation on vision transformer}
\label{appendix: subsec vit results}
Until now, the reported 8,000 pairs of evaluations in BackdoorBench are all conducted on the convolutional neural networks. 
In the following, we expand the evaluations to another popular family of models, \ie, vision transformer (ViT), which has shown superior performance on many vision tasks (\ie, image classification, object detection, semantic segmentation). 
In our evaluations, a initial checkpoint of ViT that is pre-trained on ImageNet \cite{deng2009imagenet} is downloaded from \url{https://github.com/pytorch/vision/tree/main/references/classification}. 
We then fine-tune this pre-trained checkpoint on the poisoned CIFAR-10 dataset. 
Note that the input size of ViT is $224 \times 224$, while the size of raw images in CIFAR-10 is $32 \times 32$. Thus, in data poisoning based attack, we firstly insert the trigger into the raw image, then re-scale the poisoned image to the size $224 \times 224$. 
}

\comment{
\blue{\subsection{Evaluation on vision transformer}
Our proposed BackdoorBench can easily extend to vision transformer, a family of attention-based encoder-decoder networks, which have been demonstrated their effectiveness on many computer vision tasks (such as classification, detection, segmentation). 
In our benchmark, we implement ViT proposed in \cite{vit} as a representative of vision transformer.
The pre-trained weights are downloaded from \url{https://github.com/pytorch/vision/tree/main/references/classification}. 
These weights were trained from scratch on the ImageNet-1K. 
Because the ViT expects each image to be of the same resolution ($224 \times 224$), we first add trigger on the clean image and then resize it into $224 \times 224$. Then we fine-tune the ViT on the poisoned dataset. 
}
}

\blue{We evaluate ViT-Base model with $16\times 16$ input patch size (ViT-b-16) on CIFAR-10 with 10\% poison ratio, where the settings of all hyper-parameters are same with those for learning other models, as shown in Tables \ref{tab:attack_param} and \ref{tab:defense_param}. The backdoor evaluation results are summarized in Table \ref{tab:vit}. As a baseline, the accuracy of fine-tuning ViT-b-16 on the clean dataset with same hyper-parameters is $96.56\%$. 
According to Table \ref{tab:vit} and the comparison with the evaluations on other models, we have the following observations.
\textbf{1)} In the case of no defense, the ASR of ViT-b-16 is still very high under all evaluated backdoor attacks, revealing that the ViT model architecture is also vulnerable to backdoor attacks. 
\textbf{2)} The evaluated three defense methods show very poor performance for the attack on ViT-b-16. In terms of FT and NC, although the ASR is reduced significantly, the clean accuracy is also downgraded. In terms of ABL, the ASR doesn't decrease for most attacks, with the only exception for BadNets of which the model after defense is fully degenerated. It implies that the effective defense that have been verified on the CNN architecture may not suitable for the ViT architecture. It inspires us to develop more effective defense methods for the ViT architecture specially. More evaluations and analysis about the ViT architecture will be added in BackdoorBench in future.}

\subsection{\blue{Evaluation on ImageNet}}
\label{appendix: subsec imagenet results}

\blue{Due to the high computational and memory costs, one of the benchmark datasets of image classification, \ie, ImageNet \cite{deng2009imagenet} with 1,000 classes, has rarely been evaluated in existing backdoor learning works. 
We plan to provide comprehensive evaluations of backdoor learning methods on ImageNet, to find whether there are some unique challenges for backdoor learning on large-scale datasets. }

\begin{table*}[!h]
\centering
\caption{BadNets and Blended attack result on ImageNet}
\label{table: results on imagenet}
\begin{tabular}{@{}l|lll@{}}
\toprule
        & C-Acc & ASR   & R-Acc \\ \midrule
BadNets & 69.21 & 75.86 & 0.33  \\
Blended & 69.24 & 98.59 & 0.11  \\ \bottomrule
\end{tabular}

\end{table*}

\blue{Here, we provide some partial evaluations, including BadNets and Blended attack with 0.1\% poison ratio on ImageNet and the PreAct-ResNet18 model, as shown in Table \ref{table: results on imagenet}. 
Note that due to the 1,000 classes, we do not set a higher poisoning ratio to ensure that the number of poisoned samples is not much larger than the number of clean samples of the target class. 
Both Badnets and Blended show good attack performance with high ASR and C-Acc. Compared with the baseline model, \ie, PreAct-ResNet18 trained on the clean ImageNet dataset (please refer to \url{https://paperswithcode.com/sota/image-classification-on-imagenet?tag_filter=3}), there is a slight drop of C-Acc, from 72.33\% to 69.22\%. 
More backdoor attack and defense evaluations on ImageNet will be added to our BackdoorBench in the future.
}

\subsection{Visualization} 
\label{appendix: subsec visualization}

\subsubsection{Individual visualization tools}
\label{appendix: subsubsec individual visualization}

Here we provide three visualization tools to analyze each individual image. 

\noindent
\textbf{Gradient-weighted class activation mapping (Grad-CAM)} \cite{grad-cam} explains the contribution of each pixel to the prediction of one image, based on the gradient of the logit of one class \wrt each pixel. Note that in the codebase of BackdoorBench, we implement a variant of Grad-CAM, called FullGrad \cite{NEURIPS2019_80537a94}.

\noindent
\textbf{Shapley Value} \cite{NIPS2017_7062} is another popular interpretation tool that assigns an importance factor to each pixel for a particular prediction. Inspired by the cooperative game theory, the competition among pixels is also taken into account in the computation of each individual importance factor. 

\noindent
\textbf{Frequency saliency map (FSM)} 
shows the contribution of every Fourier basis to model classification. Consider an image classification task with $S$ classes. Let $\vx$ be a clean image with size $H\times W \times C$ and $\tilde{\vx}=\DFT(\vx)$ be the corresponding frequency spectrum with $\DFT$ being the channel-wise Discrete Fourier Transform (DFT) operator. 
Let $\mathbb{C}$ be the set of complex numbers. Denote the classifier by $f:\mathbb{R}^{H\times W\times C}\to \mathbb{R}^S$. We define $F:\mathbb{C}^{H\times W\times C}\rightarrow\sR^{S}$ as the corresponding classifier in the frequency domain, which means
\begin{equation}
\label{DFT_f}
F(\tilde{\vx}) = f(\vx) = f\big( \IDFT(\tilde{\vx}) \big), 
\end{equation}
where $\IDFT$ is the channel-wise Inverse Discrete Fourier Transform (IDFT) operator, which means $\IDFT(\tilde{\vx})=\sum_{h=0}^{H-1}\sum_{w=0}^{W-1}\tilde{\vx}(h,w)e^{-2\pi i(\frac{uh}{H}+\frac{vw}{W})}$.

Inspired by the saliency map in the RGB space, we intend to establish the connection between model prediction and image’s frequency spectrum by the norm of gradient. According to the chain's rule, we can estimate the gradient \wrt the frequency spectrum as follows
\begin{equation}
    \begin{split}
    \label{fre_equation}
    \frac{\partial F_s(\tilde{\vx})}{\partial \tilde{\vx}(u,v,c)} &=
    \sum_{c=1}^{C}\sum_{h=0}^{H-1}\sum_{w=0}^{W-1}\sum_{c'=0}^{C}\frac{\partial f_s(\vx)}{\partial \vx(h,w,c')} \cdot \frac{\partial \vx(h,w,c')}{\partial \tilde{\vx}(u,v,c)} \\
    &= 
    \sum_{c=1}^{C}\sum_{h=0}^{H-1}\sum_{w=0}^{W-1}\frac{\partial f_s(\vx)}{\partial \vx(h,w,c)} e^{2\pi i(\frac{uh}{H}+\frac{vw}{W})}, 
    \end{split}
\end{equation}
where $f_s$ means the logit output of the model $f$ \wrt the $s$-th class.

\subsubsection{Visualization results}
\label{appendix: subsubsec visualization results}

\begin{figure}[!ht]
    \centering
    \includegraphics[width=\textwidth]{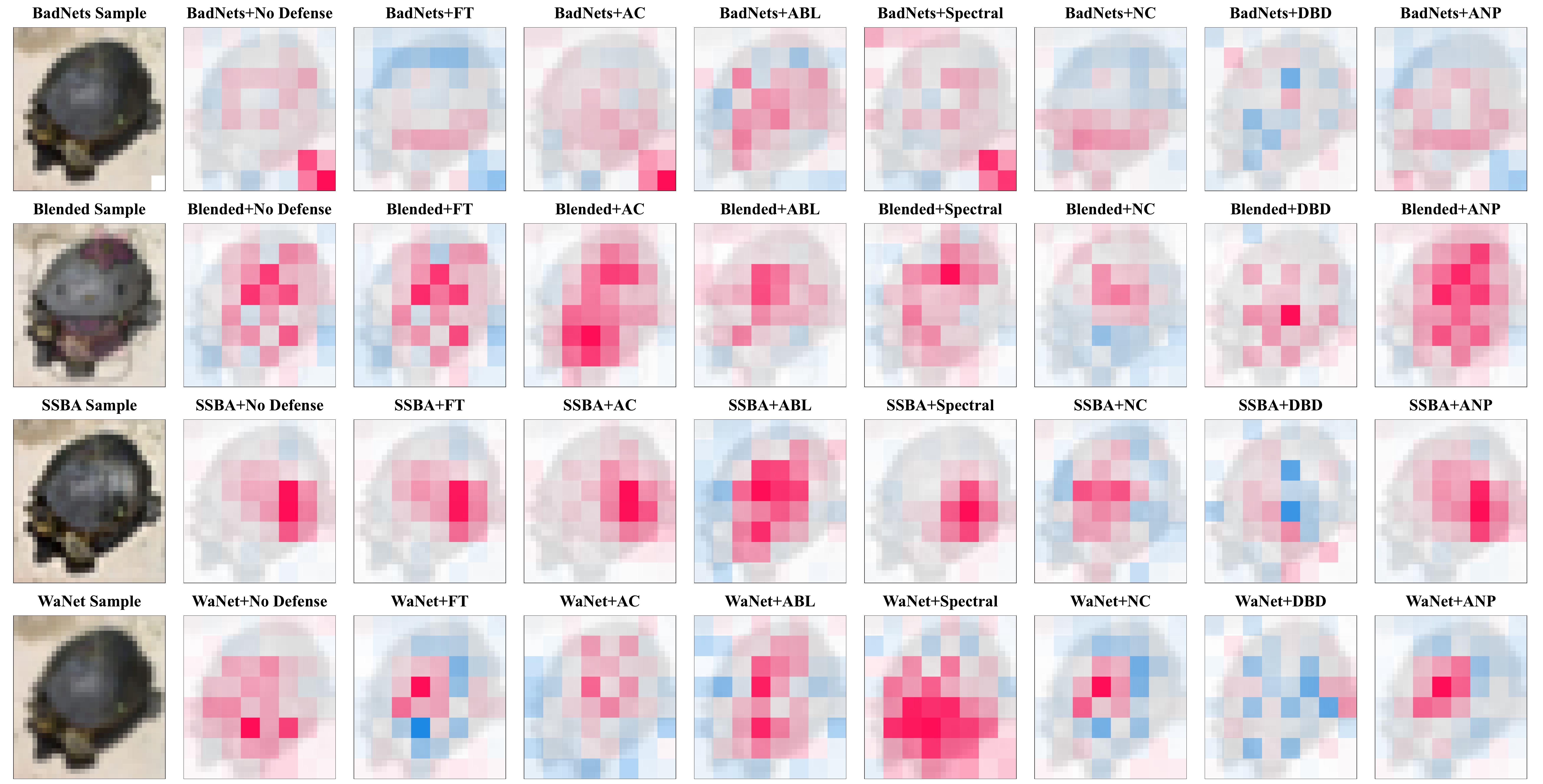}
    \caption{Shapley Value visualization of regions contributed to model decision under different attack methods and defense methods with  PreAct-ResNet18 and $5\%$ poisoning rate  on CIFAR-100.}
    \label{shap_cifar}
\end{figure}

\begin{figure}[!ht]
    \centering
    \includegraphics[width=\textwidth]{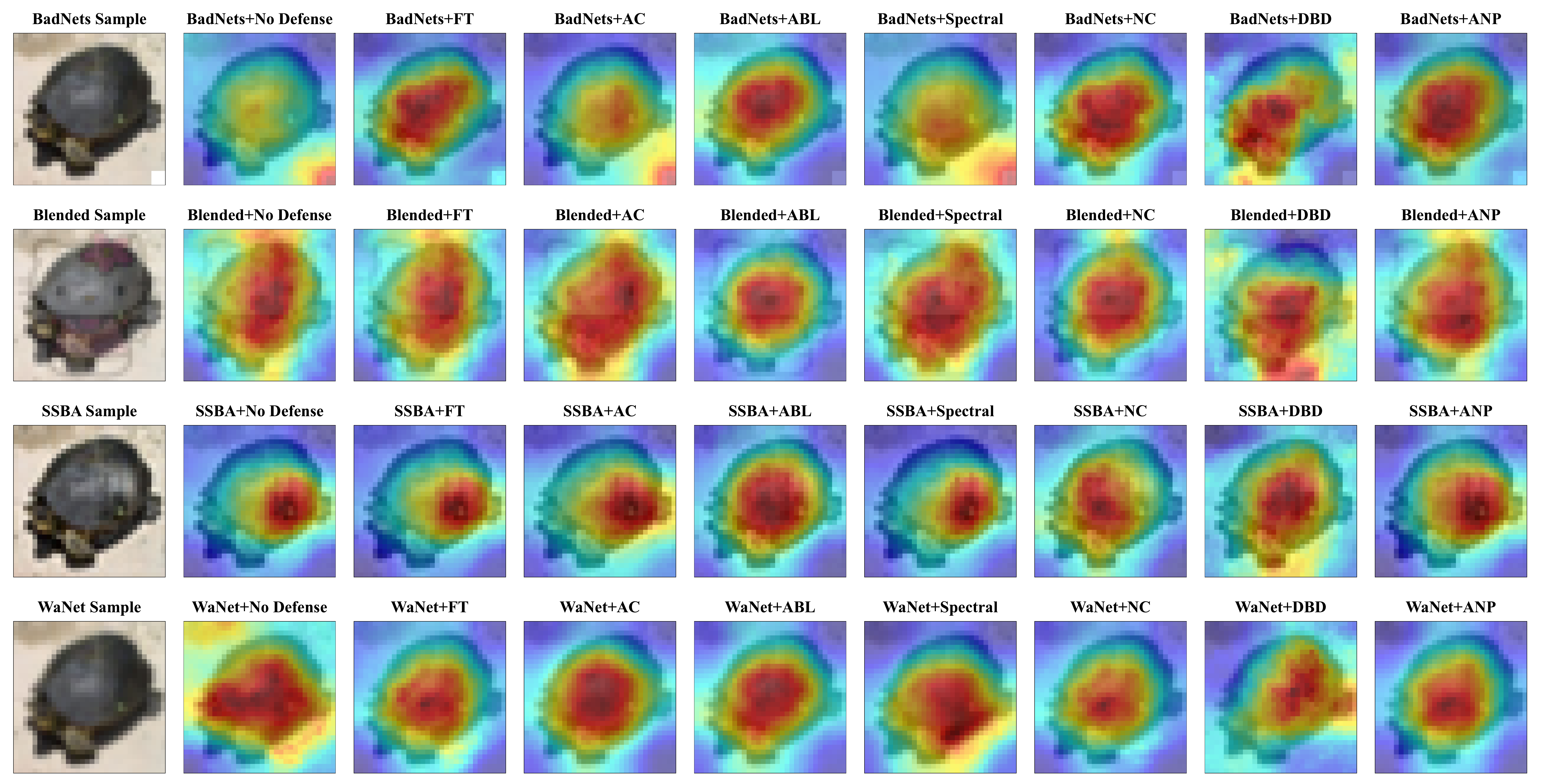}
    \caption{Grad-CAM visualization of regions contributed to model decision under different attack methods and defense methods with PreAct-ResNet18 and $5\%$ poisoning rate  on CIFAR-100.}
    \label{cam_cifar}
\end{figure}

\begin{figure}[!ht]
    \centering
    \includegraphics[width=\textwidth]{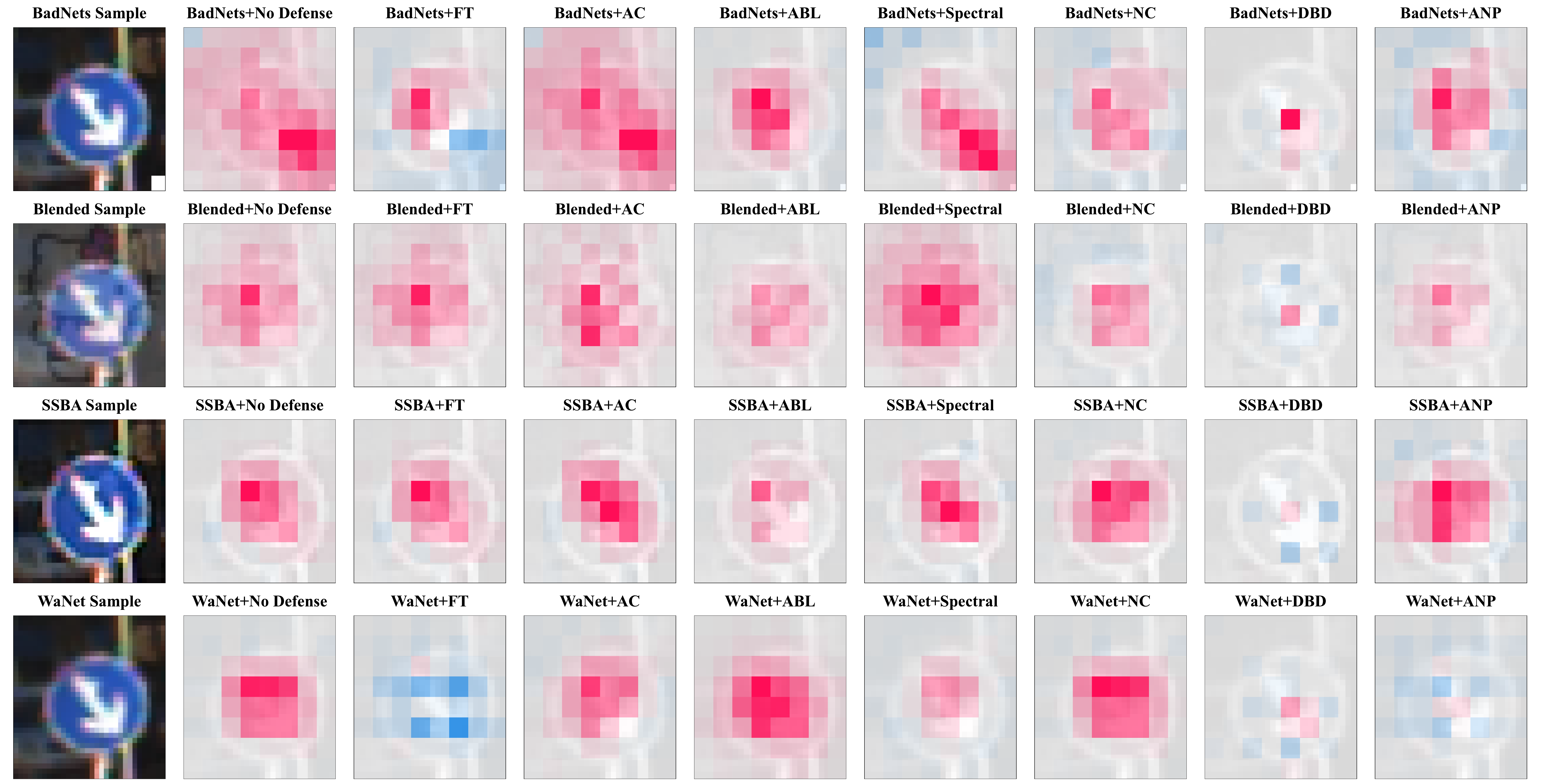}
    \caption{Shapley Value visualization of regions contributed to model decision under different attack methods and defense methods with  PreAct-ResNet18 and $5\%$ poisoning rate on GTSRB.}
    \label{shap_gt}
\end{figure}

\begin{figure}[!ht]
    \centering
    \includegraphics[width=\textwidth]{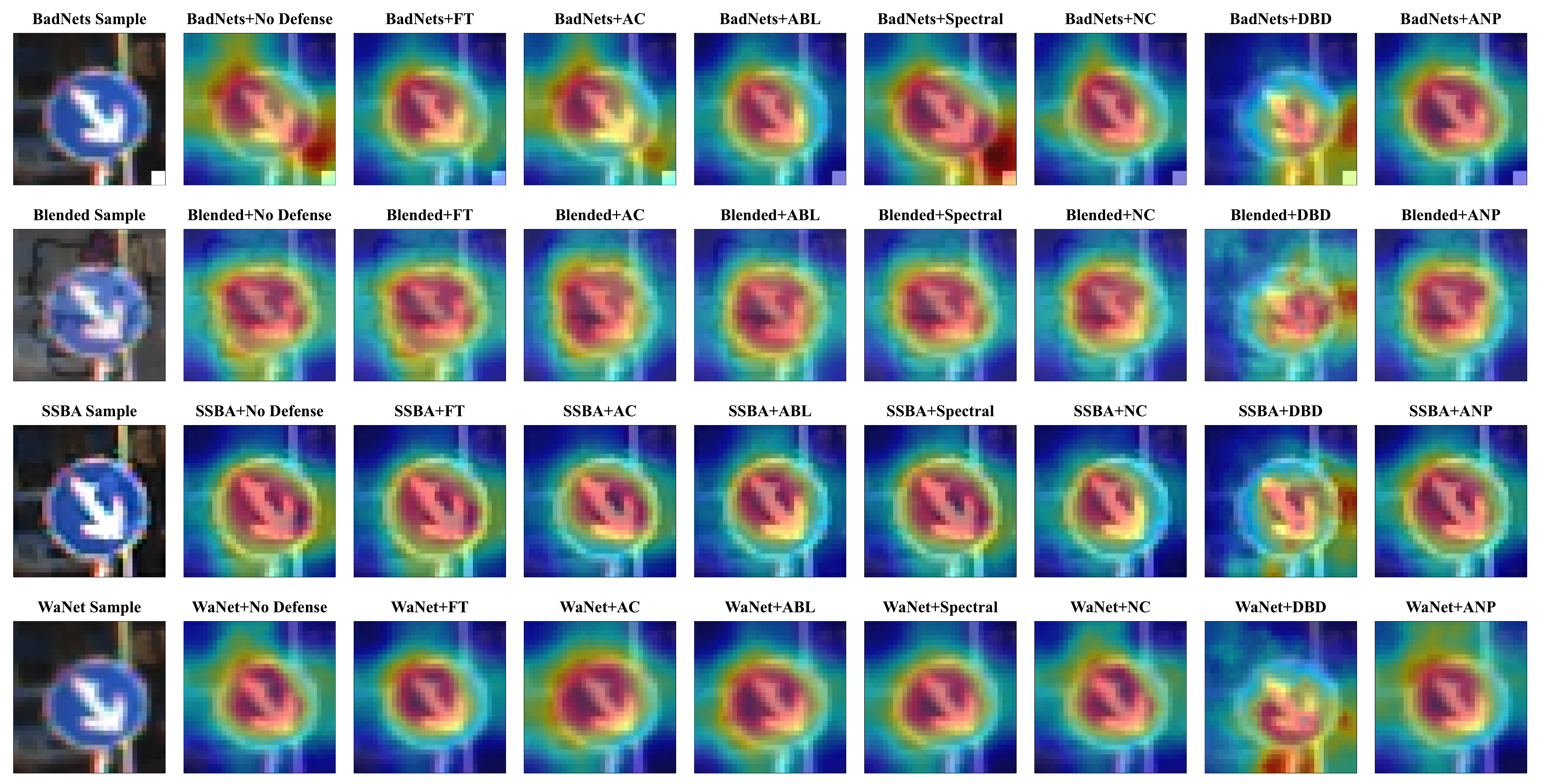}
    \caption{Grad-CAM visualization of regions contributed to model decision under different attack methods and defense methods with PreAct-ResNet18 and $5\%$ poisoning rate  on GTSRB.}
    \label{cam_gt}
\end{figure}

\begin{figure}[!ht]
    \centering
    \includegraphics[width=\textwidth]{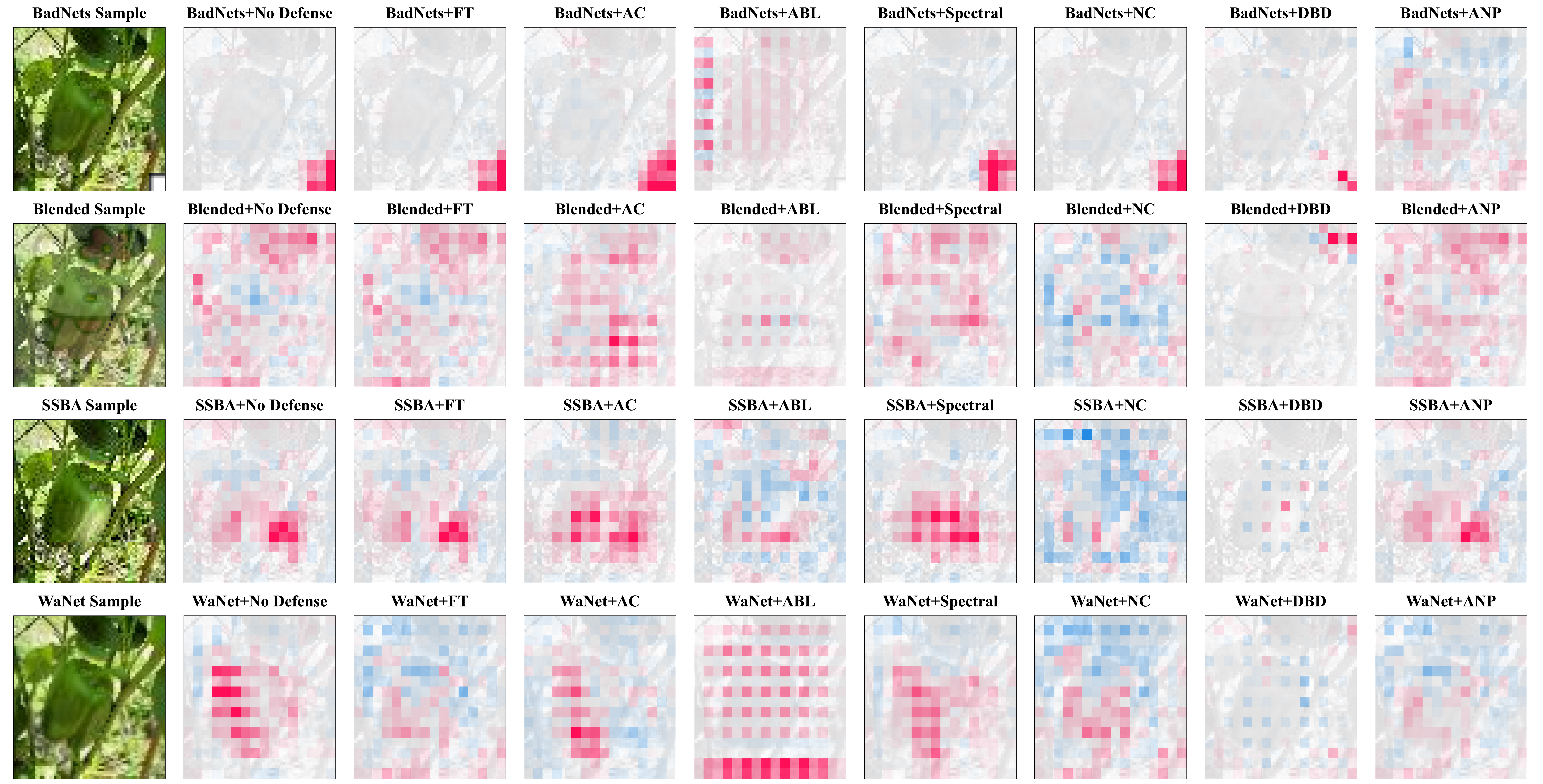}
    \caption{Shapley Value visualization of regions contributed to model decision under different attack methods and defense methods with  PreAct-ResNet18 and $5\%$ poisoning rate  on Tiny ImageNet.}
    \label{shap_tiny}
\end{figure}

\begin{figure}[!ht]
    \centering
    \includegraphics[width=\textwidth]{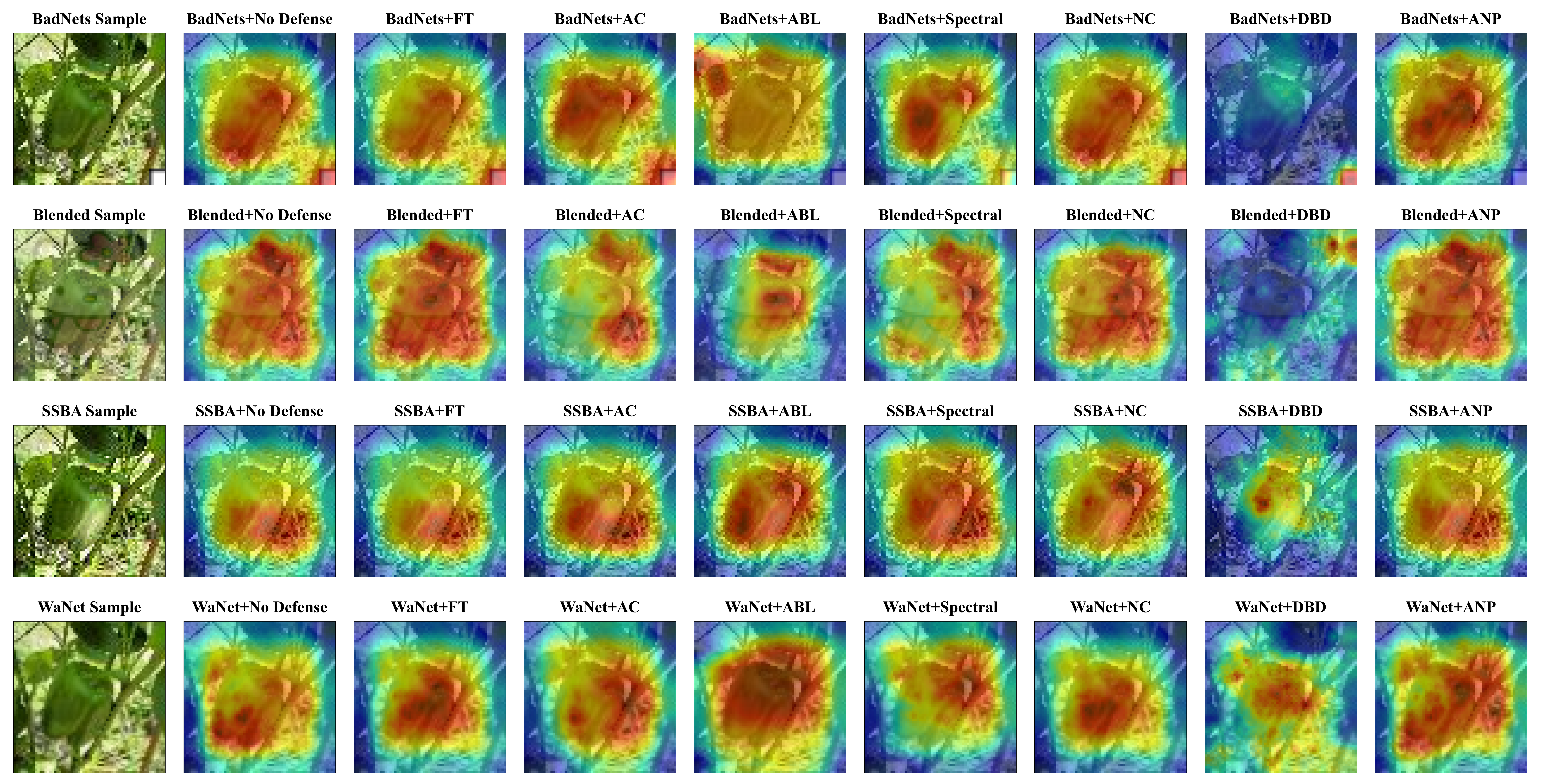}
    \caption{Grad-CAM visualization of regions contributed to model decision under different attack methods and defense methods with PreAct-ResNet18 and $5\%$ poisoning rate  on Tiny ImageNet.}
    \label{cam_tiny}
\end{figure}

In the following, we present some visualization results using the above three tools to understand the inner mechanism of backdoor learning better. Specifically, 
Specifically, we train the PreAct-ResNet18 model under various backdoor attacks and defenses with the poisoning ratio $5\%$, on 3 datasets, including CIFAR-100, GTSRB, and Tiny ImageNet. Then, we randomly select a poisoned sample from the test set and show its visualization. 
The visualization results using Shapley Value and Grad-CAM are shown in Figures \ref{shap_cifar} to \ref{cam_tiny}, respectively.

\begin{figure}[!ht]
    \centering
    \includegraphics[width=\textwidth]{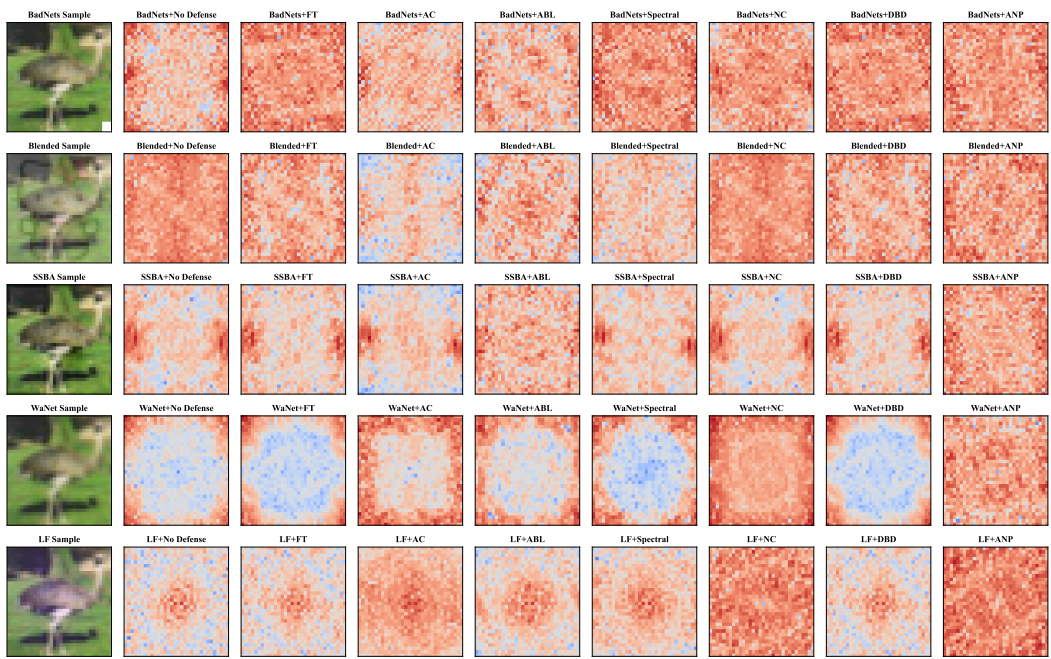}
    \caption{\blue{Frequency visualization of regions which contribute to model classification under different attack methods and defense methods with PreAct-ResNet18 and 10\% poisoning rate on CIFAR-10. Warmer color was allocated to higher value whose range is 0 to 255.}}
    \label{frequency map}
\end{figure}

\blue{
The frequency saliency map (FSM) visualization results are shown in Figure ~\ref{frequency map}, where low-frequency components are shifted into the central regions. In contrast, high-frequency components are distributed in surrounding regions. 
The first column displays the studied poisoned images generated by different attack methods, including BadNet, Blended, SSBA, WaNet, and LF. 
The second column shows the contribution of each frequency basis to the backdoored model's prediction. It tells that most backdoor models pay attention to high-frequency regions, while the model under the LF attack makes the model concentrates more on low-frequency regions. Besides, we can see an apparent difference between SSBA and WaNet in the frequency domain, even though their spatial images look similar. These observations demonstrate the potential usage of FSM in trigger or backdoor detection, which will be explored in our future work. }
\blue{
The remaining columns show the contribution of each frequency basis under various defense methods. For example, the contribution of some high-frequency regions is enormous for the poisoned image in a BadNet model without defense. However, after conducting FT on this backdoor model, the low-frequency regions regain attention from the model. It explains well that FT can effectively remove backdoors embedded by BadNet and WaNet (see Figure 3 in the main manuscript). We plan to explore more backdoor learning properties from the frequency domain perspective. 
}




\blue{
\section{BackdoorBench in Natural Language Processing}
\label{appendix: sec nlp}
Apart from the analysis of backdoor attack and defense methods in computer vision, we also expand our benchmark to the field of Natural Language Processing (see \url{https://github.com/SCLBD/BackdoorBench/tree/main/backdoorbench_nlp}). We implement two stage-of-the-art backdoor attack methods (\ie, LWS \cite{lws} and HiddenKiller \cite{hiddenkiller}) and one defense method (\ie, Onion \cite{onion}) in NLP as a complement to the original BackdoorBench. We closely follow the original implementation of the attack and defense methods and make necessary changes to unify all the methods in our benchmark. 
}
\blue{
We choose BERT \cite{bert} as the model to be poisoned. All the experiments are conducted on three widely-used datasets for text classification tasks, including Stanford Sentiment Treebank(SST-2) \cite{sst2}, Offensive Language Identification Dataset(OLID) \cite{olid} and AG’s
News \cite{agnews}. For all experiments, the poison rate is set to be 5\% and the default target label is 1. For LWS, the bar for ONION for each dataset is set to be the recommended value in the original implementation. All results are reported in Table ~\ref{tab:result_nlp}.
}

\begin{table*}[!h]
\centering
\renewcommand{\arraystretch}{1.8}
\caption{\blue{Full results on backdoor attack and defense methods for BERT}}
\label{tab:result_nlp}
\resizebox{0.99\linewidth}{!}{
\begin{tabular}{cc|ccc|ccc}
\toprule
        &   & \multicolumn{3}{c|}{No defense} & \multicolumn{3}{c}{Onion} \\ 
\midrule
Backdoor Attack &  Dataset &
  \multicolumn{1}{c}{C-Acc (\%)} &
  \multicolumn{1}{c}{ASR (\%)} &
  \multicolumn{1}{c|}{R-Acc (\%)} &
  \multicolumn{1}{c}{C-Acc (\%)} &
  \multicolumn{1}{c}{ASR (\%)} &
  \multicolumn{1}{c}{R-Acc (\%)}
\\ \midrule
LWS     & SST-2  & 89.01     & 94.08    & 4.28     & 86.20  & 90.42   & 9.58 \\
LWS     & OLID  & 82.67     & 97.92    & 0.83     & 79.07  & 96.77   & 3.23 \\
LWS     & AgNews  & 93.11     & 99.19    & 0.61     & 92.10  & 68.03  & 10.97 \\
HiddenKiller    & SST-2  & 90.34     & 88.93    & 11.07     & 85.67  & 88.27   & 11.73 \\
HiddenKiller    & OLID  & 82.19     & 97.42    & 2.59     & 81.37  & 96.12   & 3.88 \\
HiddenKiller & AgNews  & 93.49     & 98.67    & 1.12     & 92.05     & 95.16     & 4.21    \\
\bottomrule
\end{tabular}}
\end{table*}

\blue{
We can find that the two chosen attack methods can both achieve high attack success rate even at a low poisoning ratio. However, the defense performance of ONION against two SOTA attack methods is not quite satisfactory. The possible reason is that ONION aims to find out obvious outliers in each sentence, but both HiddenKiller and LWS are invisible methods which do not rely on special tokens as triggers.
}
\blue{
In the future, we will also keep updating latest backdoor attack and defense methods in the NLP field into our benchmark. 
}

\section{Reproducibility}
\label{appendix: sec reproducibility}

All evaluation results in BackdoorBench can be easily reproducible, just running the scripts provided in the github repository \url{https://github.com/SCLBD/BackdoorBench}, with the hyper-parameter settings presented in Tables \ref{tab:attack_param} and  \ref{tab:defense_param}. All evaluated datasets and model architectures are publicly and freely available. 
Besides, we also compress all codes into one file as a part of the supplementary materials.

\section{License}
\label{appendix: sec license}

This repository is licensed by The Chinese University of Hong Kong, Shenzhen and Shenzhen Research Institute of Big Data under Creative Commons Attribution-NonCommercial 4.0 International Public License (identified as CC BY-NC-4.0 in SPDX, see \url{https://spdx.org/licenses/}). More details about the license could be found in \url{https://github.com/SCLBD/BackdoorBench/blob/main/LICENSE}.

\begin{sidewaystable}[!ht]
\Huge
\caption{Full results on CIFAR-10 with 10\% poisoning ratio.}
\label{tab:cifar10_0.1}
\renewcommand\arraystretch{2}
\resizebox{\textwidth}{!}
{

\begin{tabular}{@{}l|l|ccc|ccc|ccc|ccc|ccc|ccc|ccc|ccc|ccc|ccc@{}}
\hline
                  & Backdoor   Defense & \multicolumn{3}{c|}{No   defense}                                                               & \multicolumn{3}{c|}{FT}                                                                         & \multicolumn{3}{c|}{FP}                                                                         & \multicolumn{3}{c|}{NAD}                                                                        & \multicolumn{3}{c|}{NC}                                                                         & \multicolumn{3}{c|}{ANP}                                                                        & \multicolumn{3}{c|}{AC}                                                                         & \multicolumn{3}{c|}{Spectral}                                                                   & \multicolumn{3}{c|}{ABL}                                                                        & \multicolumn{3}{c}{DBD}                                                                        \\
\hline
\end{tabular}

}
\end{sidewaystable}

\begin{sidewaystable}[!ht]
\Huge
\caption{Full results on CIFAR-10 with 5\% poisoning ratio.}
\label{tab:cifar10_0.05}
\renewcommand\arraystretch{2}
\resizebox{\textwidth}{!}
{

\begin{tabular}{@{}l|l|ccc|ccc|ccc|ccc|ccc|ccc|ccc|ccc|ccc|ccc@{}}
\toprule
                  & \multicolumn{1}{l|}{Backdoor   Defense  $\rightarrow$} & \multicolumn{3}{c|}{No   defense}                       & \multicolumn{3}{c|}{FT}                                 & \multicolumn{3}{c|}{FP}                                 & \multicolumn{3}{c|}{NAD}                                & \multicolumn{3}{c|}{NC}                                 & \multicolumn{3}{c|}{ANP}                                & \multicolumn{3}{c|}{AC}                                 & \multicolumn{3}{c|}{Spectral}                           & \multicolumn{3}{c|}{ABL}                                & \multicolumn{3}{c}{DBD}            \\ \midrule
Targeted Model    & \multicolumn{1}{l|}{Backdoor Attack $\downarrow$}    & C-Acc (\%) & ASR (\%) & \multicolumn{1}{l|}{R-Acc (\%)} & C-Acc (\%) & ASR (\%) & \multicolumn{1}{l|}{R-Acc (\%)} & C-Acc (\%) & ASR (\%) & \multicolumn{1}{l|}{R-Acc (\%)} & C-Acc (\%) & ASR (\%) & \multicolumn{1}{l|}{R-Acc (\%)} & C-Acc (\%) & ASR (\%) & \multicolumn{1}{l|}{R-Acc (\%)} & C-Acc (\%) & ASR (\%) & \multicolumn{1}{l|}{R-Acc (\%)} & C-Acc (\%) & ASR (\%) & \multicolumn{1}{l|}{R-Acc (\%)} & C-Acc (\%) & ASR (\%) & \multicolumn{1}{l|}{R-Acc (\%)} & C-Acc (\%) & ASR (\%) & \multicolumn{1}{l|}{R-Acc (\%)} & C-Acc (\%) & ASR (\%) & R-Acc (\%) \\ \midrule
PreAct-Resnet18   & \multicolumn{1}{l|}{BadNets}            & 92.64      & 88.74    & \multicolumn{1}{l|}{10.78}      & 91.22      & 4.49     & \multicolumn{1}{l|}{88.22}      & 92.47      & 17.47    & \multicolumn{1}{l|}{77.90}      & 91.03      & 4.73     & \multicolumn{1}{l|}{87.93}      & 89.89      & 1.17     & \multicolumn{1}{l|}{89.70}      & 87.24      & 0.34     & \multicolumn{1}{l|}{87.34}      & 90.27      & 75.61    & \multicolumn{1}{l|}{23.58}      & 91.21      & 69.40    & \multicolumn{1}{l|}{29.21}      & 84.08      & 0.00     & \multicolumn{1}{l|}{90.50}      & 87.68      & 2.11     & 87.01      \\
\bottomrule
\end{tabular}
}
\end{sidewaystable}

\begin{sidewaystable}[!ht]
\Huge
\caption{Full results on CIFAR-10 with 1\% poisoning ratio.}
\label{tab:tab:cifar10_0.01}
\renewcommand\arraystretch{2}
\resizebox{\textwidth}{!}
{

\begin{tabular}{@{}l|l|ccc|ccc|ccc|ccc|ccc|ccc|ccc|ccc|ccc|ccc@{}}
\toprule
                  & Backdoor   Defense & \multicolumn{3}{c|}{No   defense}  & \multicolumn{3}{c|}{FT}            & \multicolumn{3}{c|}{FP}            & \multicolumn{3}{c|}{NAD}           & \multicolumn{3}{c|}{NC}            & \multicolumn{3}{c|}{ANP}           & \multicolumn{3}{c|}{AC}            & \multicolumn{3}{c|}{Spectral}      & \multicolumn{3}{c|}{ABL}           & \multicolumn{3}{c}{DBD}            \\ \midrule
Targeted Model    & Backdoor Attack    & C-Acc (\%) & ASR (\%) & R-Acc (\%) & C-Acc (\%) & ASR (\%) & R-Acc (\%) & C-Acc (\%) & ASR (\%) & R-Acc (\%) & C-Acc (\%) & ASR (\%) & R-Acc (\%) & C-Acc (\%) & ASR (\%) & R-Acc (\%) & C-Acc (\%) & ASR (\%) & R-Acc (\%) & C-Acc (\%) & ASR (\%) & R-Acc (\%) & C-Acc (\%) & ASR (\%) & R-Acc (\%) & C-Acc (\%) & ASR (\%) & R-Acc (\%) & C-Acc (\%) & ASR (\%) & R-Acc (\%) \\ \midrule
PreAct-Resnet18   & BadNets            & 93.14      & 74.73    & 24.24      & 93.07      & 10.36    & 84.58      & 92.98      & 20.18    & 75.87      & 92.52      & 8.33     & 85.96      & 91.64      & 0.88     & 91.68      & 84.67      & 0.29     & 85.30      & 88.88      & 27.18    & 68.14      & 92.62      & 74.20    & 24.88      & 72.81      & 53.20    & 41.31      & 78.09      & 2.99     & 76.51      \\
PreAct-Resnet18   & Blended            & 93.76      & 94.88    & 4.79       & 93.53      & 91.96    & 7.56       & 93.48      & 94.14    & 5.41       & 93.34      & 92.07    & 7.41       & 90.98      & 65.91    & 29.24      & 86.92      & 37.01    & 40.68      & 89.77      & 86.09    & 12.67      & 93.27      & 93.32    & 6.28       & 66.26      & 0.17     & 34.91      & 70.18      & 8.04     & 55.60      \\
\bottomrule
\end{tabular}

}
\end{sidewaystable}

\begin{sidewaystable}[!ht]\Huge
\caption{Full results on CIFAR-10 with 0.5\% poisoning ratio.}
\label{tab:tab:cifar10_0.005}
\renewcommand\arraystretch{2}
\resizebox{\textwidth}{!}
{

\begin{tabular}{@{}l|l|ccc|ccc|ccc|ccc|ccc|ccc|ccc|ccc|ccc|ccc@{}}
\toprule
                  & \multicolumn{1}{l|}{Backdoor   Defense $\rightarrow$} & \multicolumn{3}{c|}{No   defense}                       & \multicolumn{3}{c|}{FT}                                 & \multicolumn{3}{c|}{FP}                                 & \multicolumn{3}{c|}{NAD}                                & \multicolumn{3}{c|}{NC}                                 & \multicolumn{3}{c|}{ANP}                                & \multicolumn{3}{c|}{AC}                                 & \multicolumn{3}{c|}{Spectral}                           & \multicolumn{3}{c|}{ABL}                                & \multicolumn{3}{c}{DBD}            \\ \midrule
Targeted Model    & \multicolumn{1}{l|}{Backdoor Attack  $\downarrow$}    & C-Acc (\%) & ASR (\%) & \multicolumn{1}{l|}{R-Acc (\%)} & C-Acc (\%) & ASR (\%) & \multicolumn{1}{l|}{R-Acc (\%)} & C-Acc (\%) & ASR (\%) & \multicolumn{1}{l|}{R-Acc (\%)} & C-Acc (\%) & ASR (\%) & \multicolumn{1}{l|}{R-Acc (\%)} & C-Acc (\%) & ASR (\%) & \multicolumn{1}{l|}{R-Acc (\%)} & C-Acc (\%) & ASR (\%) & \multicolumn{1}{l|}{R-Acc (\%)} & C-Acc (\%) & ASR (\%) & \multicolumn{1}{l|}{R-Acc (\%)} & C-Acc (\%) & ASR (\%) & \multicolumn{1}{l|}{R-Acc (\%)} & C-Acc (\%) & ASR (\%) & \multicolumn{1}{l|}{R-Acc (\%)} & C-Acc (\%) & ASR (\%) & R-Acc (\%) \\ \midrule
PreAct-Resnet18   & \multicolumn{1}{l|}{BadNets}            & 93.76      & 50.06    & \multicolumn{1}{l|}{48.14}      & 93.49      & 4.94     & \multicolumn{1}{l|}{89.83}      & 93.67      & 5.01     & \multicolumn{1}{l|}{89.60}      & 93.30      & 7.67     & \multicolumn{1}{l|}{87.31}      & 92.83      & 0.62     & \multicolumn{1}{l|}{92.67}      & 91.32      & 0.76     & \multicolumn{1}{l|}{91.31}      & 89.51      & 2.29     & \multicolumn{1}{l|}{88.72}      & 92.97      & 7.91     & \multicolumn{1}{l|}{87.31}      & 68.40      & 71.78    & \multicolumn{1}{l|}{22.14}      & 73.60      & 0.10     & 80.01      \\
\bottomrule
\end{tabular}

}
\end{sidewaystable}

\begin{sidewaystable}[!ht]\Huge
\caption{Full results on CIFAR-10 with 0.1\% poisoning ratio.}
\label{tab:cifar10_0.001}
\renewcommand\arraystretch{2}
\resizebox{\textwidth}{!}
{

\begin{tabular}{@{}l|l|ccc|ccc|ccc|ccc|ccc|ccc|ccc|ccc|ccc|ccc@{}}
\toprule
                  & \multicolumn{1}{l|}{Backdoor   Defense $\rightarrow$} & \multicolumn{3}{c|}{No   defense}                       & \multicolumn{3}{c|}{FT}                                 & \multicolumn{3}{c|}{FP}                                 & \multicolumn{3}{c|}{NAD}                                & \multicolumn{3}{c|}{NC}                                 & \multicolumn{3}{c|}{ANP}                                & \multicolumn{3}{c|}{AC}                                 & \multicolumn{3}{c|}{Spectral}                           & \multicolumn{3}{c|}{ABL}                                & \multicolumn{3}{c}{DBD}            \\ \midrule
Targeted Model    & \multicolumn{1}{l|}{Backdoor Attack  $\downarrow$}    & C-Acc (\%) & ASR (\%) & \multicolumn{1}{l|}{R-Acc (\%)} & C-Acc (\%) & ASR (\%) & \multicolumn{1}{l|}{R-Acc (\%)} & C-Acc (\%) & ASR (\%) & \multicolumn{1}{l|}{R-Acc (\%)} & C-Acc (\%) & ASR (\%) & \multicolumn{1}{l|}{R-Acc (\%)} & C-Acc (\%) & ASR (\%) & \multicolumn{1}{l|}{R-Acc (\%)} & C-Acc (\%) & ASR (\%) & \multicolumn{1}{l|}{R-Acc (\%)} & C-Acc (\%) & ASR (\%) & \multicolumn{1}{l|}{R-Acc (\%)} & C-Acc (\%) & ASR (\%) & \multicolumn{1}{l|}{R-Acc (\%)} & C-Acc (\%) & ASR (\%) & \multicolumn{1}{l|}{R-Acc (\%)} & C-Acc (\%) & ASR (\%) & R-Acc (\%) \\ \midrule
PreAct-Resnet18   & \multicolumn{1}{l|}{BadNets}            & 93.61      & 1.23     & \multicolumn{1}{l|}{93.24}      & 93.59      & 1.20     & \multicolumn{1}{l|}{93.20}      & 93.54      & 0.91     & \multicolumn{1}{l|}{93.24}      & 93.45      & 1.26     & \multicolumn{1}{l|}{93.03}      & 90.29      & 1.27     & \multicolumn{1}{l|}{90.08}      & 89.60      & 0.71     & \multicolumn{1}{l|}{89.37}      & 89.97      & 1.37     & \multicolumn{1}{l|}{89.57}      & 93.03      & 1.01     & \multicolumn{1}{l|}{92.53}      & 81.44      & 1.99     & \multicolumn{1}{l|}{79.69}      & 71.59      & 4.90     & 68.82      \\
\bottomrule
\end{tabular}
}
\end{sidewaystable}

\clearpage

\comment{
{\small
\bibliographystyle{plain}
\bibliography{backdoor}
}
}

\end{document}